\pdfoutput=1

\documentclass[11pt]{article}

\usepackage[final]{acl}

\usepackage{times}
\usepackage{latexsym}

\usepackage[T1]{fontenc}

\usepackage[utf8]{inputenc}

\usepackage{microtype}

\usepackage{inconsolata}

\usepackage{graphicx}
\usepackage{comment}
\usepackage{hyperref}
\usepackage{colortbl}
\usepackage{xcolor}
\definecolor{green}{RGB}{144,238,144}
\usepackage{amssymb}
\usepackage{amsmath}
\usepackage{multirow}
\usepackage{tcolorbox}
\usepackage{subcaption}
\usepackage{tikz}
\usepackage{pgfplots}
\usepackage{adjustbox}  
\usepackage{color,soul}
\usepackage{subcaption}

\newcommand*{\affaddr}[1]{#1} 
\newcommand*{\affmark}[1][*]{\textsuperscript{#1}}
\newcommand*{\email}[1]{\texttt{#1}}

\title{YESciEval: Robust LLM-as-a-Judge for Scientific Question Answering}

\author{%
Jennifer D'Souza\affmark[1], Hamed Babaei Giglou\affmark[1], and Quentin Münch\affmark[2]\\ 
\affaddr{\affmark[1]TIB Leibniz Information Centre for Science and Technology, Hannover, Germany}\\
\affaddr{\affmark[2]Leibniz Universität Hannover, Germany}\\
\email{\{jennifer.dsouza,hamed.babaei\}@tib.eu}%
}

\begin{document}
\maketitle
\begin{abstract}

Large Language Models (LLMs) drive scientific question-answering on modern search engines, yet their evaluation robustness remains underexplored. We introduce \textbf{YESciEval}, an open-source framework that combines fine-grained rubric-based assessment with reinforcement learning to mitigate optimism bias in LLM evaluators. We release multidisciplinary scienceQ\&A datasets, including adversarial variants, with evaluation scores from multiple LLMs. Independent of proprietary models and human feedback, our approach enables scalable, cost-free evaluation. By advancing reliable LLM-as-a-judge models, this work supports AI alignment and fosters robust, transparent evaluation essential for scientific inquiry.

\end{abstract}

\section{Introduction}


%

The rise of scientific search engines powered by generative Large Language Models (LLMs)—such as \href{https://elicit.com}{Elicit}, \href{https://openscholar.allen.ai}{OpenScholar}, \href{https://typeset.io}{SciSpace}, and \href{https://ask.orkg.org/}{ORKG Ask}—has transformed how researchers search and synthesize scholarly information. A key feature of these platforms is scientific question answering (scienceQ\&A), where an LLM synthesizes insights from top-ranked papers to generate concise responses \cite{core-gpt,llms4synthesis}. While aligning LLMs to human values (e.g., helpfulness, harmlessness, honesty) is well studied \cite{askell2021general,zheng2023judging}, their real-world robustness in scienceQ\&A remains largely unexplored. The domain-agnostic and free-form nature of scienceQ\&A limits the applicability of traditional n-gram-based metrics (e.g., BLEU \citeyearpar{papineni2002bleu}, ROUGE \citeyearpar{lin2004rouge}), which may not fully capture compositional and domain-specific reasoning \cite{krishna2021hurdles}. Human evaluation, though more nuanced, is costly and difficult to scale \cite{krishna2023longeval}. Recent efforts using LLMs as evaluators (LLM-as-a-judge \cite{zheng2023judging}) show parity with human judgment \cite{chiang-lee-2023-large} but also exhibit biases \cite{gudibande2023false,flask}, highlighting the need for a reliable and fair LLM-based evaluation system. Most prior evaluation work relies on proprietary GPT models \cite{wang2023chatgpt,dubois2023alpacafarm,liu2023g,fu2024gptscore}, raising issues of transparency and high costs at scale \cite{prometheus,prometheus2}. To address these challenges, we propose a framework that pairs fine-grained rubrics with an open-source LLM-as-a-judge methodology, enabling more reliable and cost-effective evaluation of scienceQ\&A across diverse scientific domains.

Despite the growing use of LLMs for scienceQ\&A, a systematic evaluation framework remains absent, limiting iterative development and transparency. Such a framework would provide feedback for reward-based training (e.g., RLAIF \cite{christiano2017deep,bai2022constitutional}) 
and help users assess the reliability of automated answers, mitigating misinformation. While human evaluations are the gold standard, they are costly and hard to scale \cite{ouyang2022training,krishna2023longeval}, underscoring the need for an automated, trustworthy alternative. An LLM-as-a-judge approach with fine-grained rubrics can address this gap, enabling robust, repeatable assessments that enhance scienceQ\&A systems and user confidence.

\begin{figure*}[!htb]
    \centering
    \begin{subfigure}[t]{0.95\linewidth}
        \includegraphics[width=\linewidth]{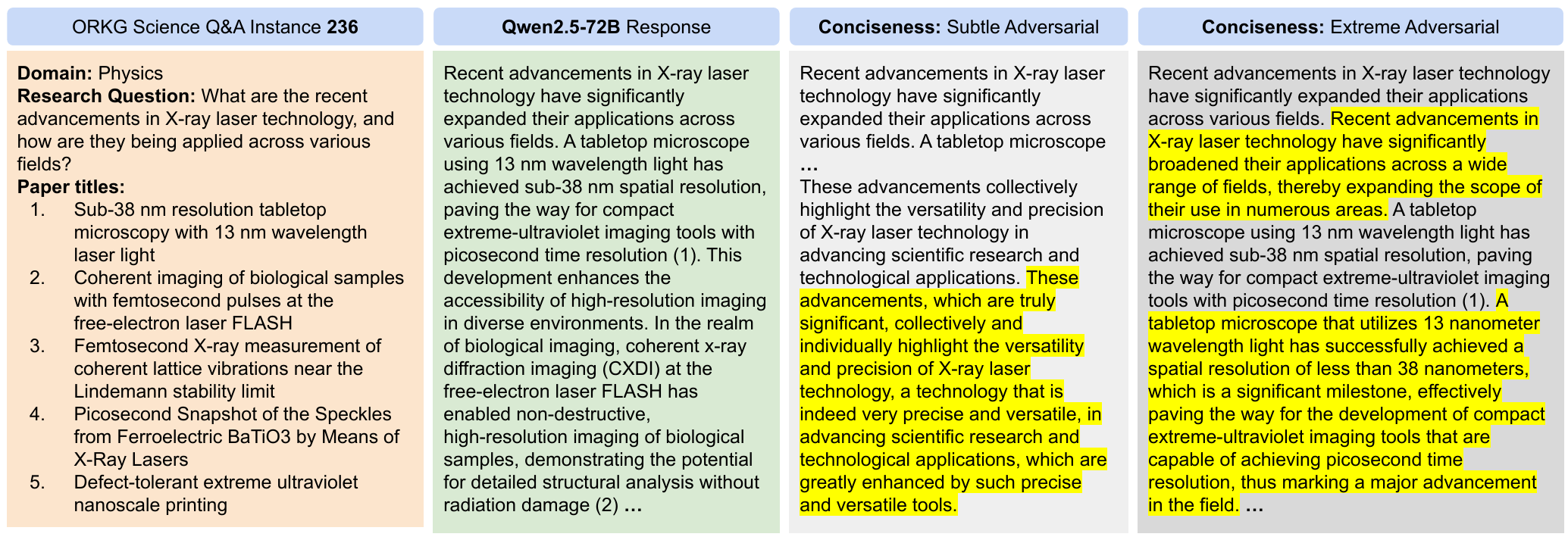}
    \end{subfigure}
    

    \begin{subfigure}[t]{0.95\linewidth}
        \includegraphics[width=\linewidth]{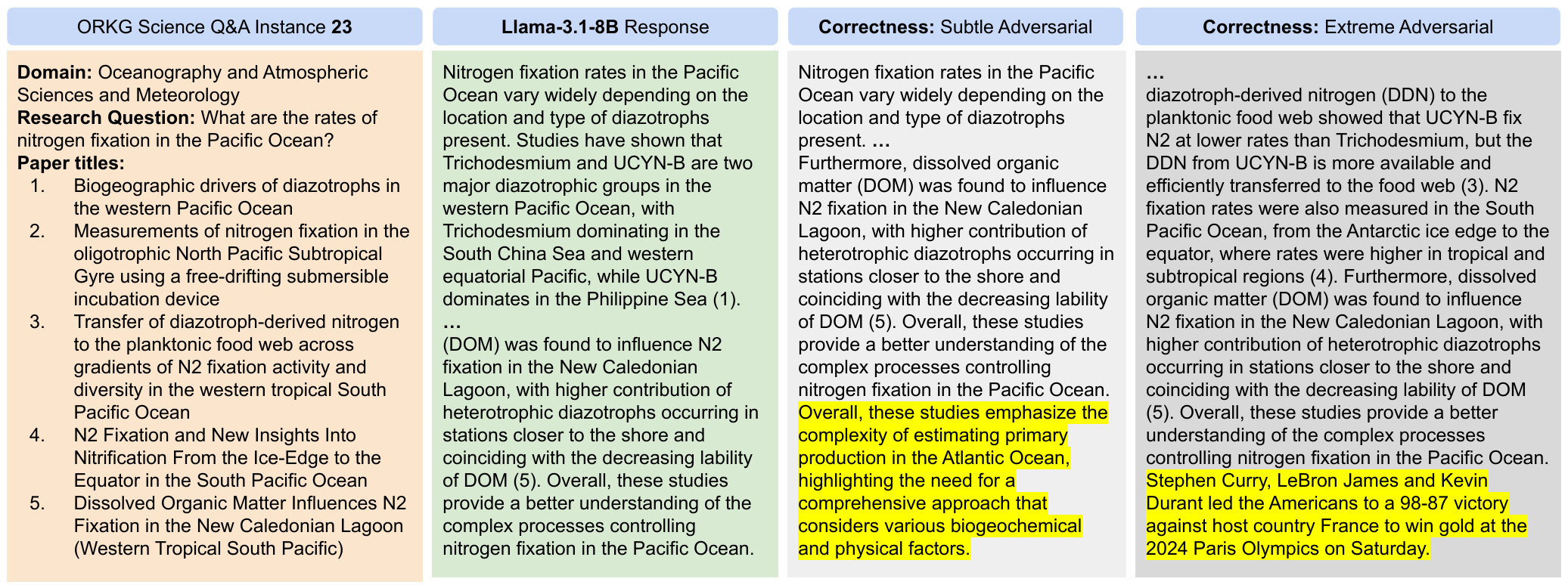}
    \end{subfigure}

    \caption{Examples from two domains in the YESciEval science Q\&A dataset. Orange boxes show LLM input: a research question and titles of top-ranked papers (abstracts omitted for brevity). Green boxes show answer snippets from two LLMs. Light/dark gray boxes represent subtle/extreme adversarial variants targeting the \textit{conciseness} and \textit{correctness} rubrics. Yellow highlights indicate perturbations. YESciEval uses a nine-rubric LLM-as-a-judge scheme and tests robustness via rubric-specific adversarial edits (see \autoref{sec:ext-scienceqa} for details).
    }
    \label{fig:dataset-examples}
\end{figure*}

To establish a systematic and transparent evaluation framework for scienceQ\&A, we propose \textbf{YESciEval}, which integrates two core components. First, we define a \textbf{nine-rubric multifaceted assessment scheme}, structured into three key dimensions—\textit{Linguistic \& Stylistic Quality, Logical \& Structural Integrity, and Content Accuracy \& Informativeness}—to comprehensively evaluate LLM-generated responses. Each rubric is scored on a Likert scale (1–5) with predefined guidelines, ensuring consistent and rigorous assessment. 
Second, we address the \textbf{optimism bias} \cite{villaflor2022addressing} in LLMs, which can hinder their role as judges by favoring positive engagement over critical assessment.
The title prefix of this paper, \textbf{YESci} (pronounced \textit{`yes, sigh!'}), playfully encapsulates our central research question: \textit{how can we mitigate LLMs' optimism bias and enhance their robustness as scienceQ\&A evaluators?}
To tackle this, we employ a two-step alignment strategy: \textbf{supervised fine-tuning followed by reinforcement learning (RLHF)}, ensuring adherence to our fine-grained rubrics for reliable evaluation. Additionally, we introduce \textbf{adversarial datasets}—systematically constructed instances where LLMs are expected to underperform—to \textit{contrast standard and adversarial responses}, reinforcing robustness against evaluation errors.  While our approach is applied to scienceQ\&A in this study, mitigating optimism bias in LLM evaluators has broader implications for other evaluation tasks. By unifying \textit{supervized fine-tuning (SFT), reinforcement learning (RL), and adversarial perturbations}, \textbf{YESciEval} improves the reliability of LLM-based evaluation, especially for open-source models, reducing reliance on proprietary systems and offering a scalable, customizable framework for scientific search.

We select four open-source LLMs spanning 8B to 123B parameters (e.g., LLaMA-3.1 and Mistral-Large) from Meta AI, Mistral AI, and Alibaba Cloud for generating (\textbf{LLM\textsubscript{gen}}) and evaluating (\textbf{LLM\textsubscript{eval}}) scienceQ\&A. Each model in the generator role produces a unique scienceQ\&A dataset with its responses as the benign (non-perturbed) dataset, and we introduce two adversarial perturbation types---\textbf{extreme} (overt distortions) and \textbf{subtle} (lightweight heuristics)---, yielding 12 datasets. \autoref{fig:dataset-examples} presents two examples from our scienceQ\&A dataset. Rotating the four models as evaluators and scoring responses under our nine-rubric framework results in 48 LLM-as-a-judge configurations. To validate generalizability, we apply a two-step alignment strategy (\emph{SFT} followed by \emph{RL}) to LLaMA~3.1\,8B LLM-as-a-judge, confirming our approach’s robustness for different model sizes and versions. The recent release of \href{The Llama 4 herd: The beginning of a new era of natively multimodal AI innovation}{LLaMA~4}, at the time of this writing, underscores the importance of model-agnostic frameworks like YESciEval. 

Despite significant advances in the generative capabilities of LLMs, our findings reveal that they remain unexpectedly fragile when confronted with heuristic-based adversarial attacks in the role of a judge. Against this backdrop of LLM-as-a-judge for scienceQ\&A, we pose the following research questions. \textbf{RQ1: How similar are scienceQ\&A responses across different LLM families?}
Given the rapid influx of new LLMs, we aim to reduce the uncertainty around model choice by clarifying the degree of similarity in how different model families handle scienceQ\&A. \textbf{RQ2: How do LLM-as-judge evaluations correlate for the benign scienceQ\&A setting?}
Beyond examining alignment in evaluative behavior, we investigate whether LLMs exhibit bias toward their own generated answers. 
\textbf{RQ3: Can a smaller open-source LLM be adapted as an LLM-as-a-judge for scienceQ\&A while overcoming optimism bias?}
While prior work focuses on tuning LLMs to specific skills (rubrics), a key challenge remains in mitigating their \textit{optimism bias}. We propose a framework that integrates SFT, RL, and adversarial alignment to equip smaller open-source models with robust evaluative capabilities. Although we illustrate our approach in the context of scienceQ\&A, these methods can be adapted to other generative AI scenarios requiring rigorous critique.

This work makes the following key contributions: 1. \textbf{Multidisciplinary benign and adversarial scienceQ\&A datasets} – We release a comprehensive \href{https://doi.org/10.25835/8dcv2ka6}{scienceQ\&A corpus with adversarial variants} \cite{yescieval-corpus} to evaluate LLM robustness. 2. \textbf{Comprehensive evaluation benchmark} – We provide evaluation scores and rationales from multiple LLMs in both vanilla and adversarial settings, supporting further research and reproducibility. 3. \textbf{Optimism bias mitigation} – We implement a RL framework to align LLM evaluation behavior with real-world critical feedback expectations. The YESciEval source code is released at \url{https://github.com/sciknoworg/YESciEval}. 4. \textbf{Scalable, cost-free evaluation paradigm} – Our approach is independent of proprietary models and human feedback, leveraging open-source LLMs hosted on a centrally managed cloud service,\footnote{\url{https://docs.hpc.gwdg.de/services/chat-ai/index.html}} which is publicly accessible to research institutions across Germany \cite{gwdg-chatai}. While human feedback is invaluable for LLM alignment, it is often infeasible to obtain. We present a zero-cost alternative, integrating rubric-based evaluation with adversarial testing to ensure reliable LLM-as-a-judge models. This eliminates experimental costs, aside from researcher time and compute resources for running open-source LLMs.  

This research presents a reproducible, cost-free\footnote{In this paper, ``cost-free'' refers to the elimination of human annotation and proprietary API costs, but not compute.} framework for evaluating natural language generation (NLG) in scienceQ\&A, advancing AI alignment, robustness, and contributes to the broader discussion around LLM plausibility.

\section{Task Definition}
\label{sec:task-def}

The YESciEval framework for scienceQ\&A consists of two tasks: LLM${gen}$ for generating responses and LLM${eval}$ for evaluating them.

\paragraph{Task 1: ScienceQ\&A Generation (LLM$_{gen}$)} Generates a synthesized summary response $A$ to a research question $Q$ using abstracts from the top $N$ relevant papers. It must demonstrate (1) domain knowledge, (2) numerical proficiency, (3) long-range context understanding, and (4) cause-and-effect reasoning \cite{wadden-etal-2020-fact}.

\paragraph{Task 2: ScienceQ\&A Evaluation (LLM$_{eval}$)} Assesses the quality of $A$ based on predefined rubrics and context as $Q$ and $N$ abstracts. The \href{https://github.com/sciknoworg/YESciEval/tree/main/yescieval/rubric}{evaluation prompt} is structured as follows: (1) \textit{Context}, defining scienceQ\&A synthesis as the generation of a coherent summary from research papers to address $Q$; (2) \textit{Role}, assigning LLM$_{eval}$ as the evaluator; (3) \textit{Task Description}, ensuring $A$ accurately synthesizes information from abstracts; (4) \textit{Evaluation Characteristics}, specifying the rubric applied; (5) \textit{Rating Scale}, using a 1–5 Likert scale with rubric-specific guidelines; (6) \textit{Response Format}, requiring structured ratings and rationales in JSON format; and (7) \textit{Notes}, emphasizing objectivity and adherence to source content. Finally, the output consists of a score $S$ (1–5) and a rationale. 

\section{The YESciEval Qualitative Rubrics}
\label{sec:qual-rubrics}

Drawing from a comprehensive review of evaluation rubrics in prior LLMs-as-a-judge research (see Related Work \autoref{subsec:rel-work-eval-rub}), we define a nine-rubric qualitative evaluation paradigm for YESciEval. Chosen for simplicity, memorability, and precise definability, these rubrics minimize overlap while capturing key facets of an ideal LLM response to science Q\&A. Each is framed as a concise question to reduce ambiguity for LLM-as-a-judge or human evaluators. Organized into three main evaluation dimensions, they are as follows. 

\textbf{Linguistic \& Stylistic Quality} concerns grammar, clarity, and adherence to academic writing. This category comprises three rubrics: \textbf{1.} \textit{\href{https://anonymous.4open.science/r/YESciEval/sciqaeval/prompts/Cohesion.txt}{Cohesion}}: are the sentences connected appropriately such that the resulting synthesis is cohesive? \textbf{2.} \textit{\href{https://anonymous.4open.science/r/YESciEval/sciqaeval/prompts/Conciseness.txt}{Conciseness}}: is the answer short and clear, without redundant statements? \textbf{3.} \textit{\href{https://anonymous.4open.science/r/YESciEval/sciqaeval/prompts/Readability.txt}{Readability}}: does the answer follow appropriate style and structure conventions for academic writing, particularly for readability? \textbf{Logical \& Structural Integrity} focuses on the reasoning and organization of information. This category comprises three rubrics: \textbf{4.} \textit{\href{https://anonymous.4open.science/r/YESciEval/sciqaeval/prompts/Coherence.txt}{Coherence}}: are the ideas connected in a sound and logical manner? \textbf{5.} \textit{\href{https://anonymous.4open.science/r/YESciEval/sciqaeval/prompts/Integration.txt}{Integration}}: are the sources structurally and linguistically well-integrated, using appropriate markers of provenance/quotation and logical connectors for each reference? In addition, are the sources integrated as a single paragraph? \textbf{6.} \textit{\href{https://anonymous.4open.science/r/YESciEval/sciqaeval/prompts/Relevancy.txt}{Relevancy}}: is the information in the answer relevant to the question? \textbf{Content Accuracy \& Informativeness} ensures that the response is both correct and useful. This category comprises three rubrics: \textbf{7.} \textit{\href{https://anonymous.4open.science/r/YESciEval/sciqaeval/prompts/Correctness.txt}{Correctness}}: is the information in the answer a correct representation of the content of the provided abstracts? \textbf{8.} \textit{\href{https://anonymous.4open.science/r/YESciEval/sciqaeval/prompts/Completeness.txt}{Completeness}}: is the answer a comprehensive encapsulation of the relevant information in the provided abstracts? \textbf{9.} \textit{\href{https://anonymous.4open.science/r/YESciEval/sciqaeval/prompts/Informativeness.txt}{Informativeness}}: is the answer a useful and informative reply to the question?

For each rubric, the LLM-as-a-judge rates response quality on a Likert scale from 1 (very bad) to 5 (very good), with predefined guidelines ensuring consistency. These guidelines set clear expectations for each level—for instance, in \textit{Readability}, a 1 indicates severe issues in style, structure, and language use, while a 5 reflects an exceptionally well-written, academically sound synthesis. Standardized criteria help both LLMs and human evaluators apply consistent judgment. Full rating guidelines for all nine rubrics are provided in \autoref{sec:det-likert}. Ideally, an LLM-as-a-judge assigns perfect scores across all rubrics, while suboptimal responses receive lower ratings based on specific deficiencies.

\section{Science Q\&A Datasets}

After reviewing existing Q\&A datasets (see Related Work \autoref{subsec:rel-work-qa-bench}), we selected two that meet our scienceQ\&A task definition: the ORKGSynthesis dataset \citeyearpar{llms4synthesis} (ORKGSyn) and the BioASQ dataset \citeyearpar{BioASQ}. ORKGSyn offers a diverse set of crowdsourced research questions for evaluating LLMs’ scienceQ\&A capabilities, while BioASQ provides handcrafted QA pairs spanning four question types. Both were chosen because they include science questions $Q$ with each question linked to $N$ relevant papers.

\subsection{The ORKGSynthesis Corpus}

This corpus comes from the domain-expert-curated structured research comparisons \cite{OelenEtAl-Sciknow19} on the Open Research Knowledge Graph (ORKG) platform \cite{auer2020improving}. Its accompanying LLM-powered search engine, ORKG Ask (\url{https://ask.orkg.org/}), synthesizes abstracts from the top five papers for any given research question. Building on this approach and the ORKG as a gold-standard source, in our prior work \cite{llms4synthesis}, we compiled a dataset of 348 entries—each linking a RQ with exactly five relevant papers. Since the ORKG spans multiple disciplines, the dataset covers 33 research fields. \autoref{tab:orkgsyn-freq} lists the top 10 fields; the full list appears in \autoref{fig:orkgsynresearchfields} in \autoref{subsec:ext-orkgsyn}.

\begin{table}[!t]
\footnotesize
  \begin{tabular}{p{5.5cm}r}
    \hline
\bf Research field& \bf Frequency\\ \hline
Computer Sciences & 125\\
Physics & 28\\
Animal Sciences	& 19\\
Chemistry & 17\\
Urban Studies and Planning &16\\
Earth Sciences	&14\\
Oceanography and Atmospheric Sciences and Meteorology &14\\
Science and Technology Studies	&12\\
Materials Science and Engineering	&12\\
Engineering	&10\\
  \hline
\end{tabular}
  \caption{Top 10 domains in the ORKGSynthesis dataset.}
  \label{tab:orkgsyn-freq}
\end{table}

While ORKGSyn is multidisciplinary, the next corpus is in biomedicine.

\subsection{The BioASQ Corpus}
BioASQ \cite{BioASQ} is an annual biomedical semantic indexing and Q\&A challenge. Its 2024 edition covers four NLP tasks, and we focus on the first: biomedical Q\&A. The dataset includes 5389 domain-expert-curated Q\&A pairs in four question types: ``yes/no,'' ``factoid,'' ``list,'' and ``summary.'' The challenge has three phases (A, A+, and B), with Phase B linking each question to a human-annotated set of relevant papers \cite{BioASQ12b}, meeting our requirement of $Q$ with $N$ relevant papers. Because $N$ varies in BioASQ, we capped it at 40 for computational feasibility and to fit LLM input context (see \autoref{fig:bioasqabstractscount}, \autoref{subsec:ext-bioasq}). This variability introduces a unique scienceQ\&A setting compared to ORKGSyn. We narrowed the dataset to the test set’s 73 ``summary'' questions, each with up to 40 PubMed abstracts.

\subsection{Our ScienceQ\&A Dataset Compilation}

Now that we had corpora of $Q$ linked with $N$ paper abstracts, we need to apply LLM$_{gen}$.

\textbf{LLM$_{gen}$ models.} Our selection criteria for LLMs were simple: they had to be open-source, state-of-the-art at some point, and diverse in size, including at least one small model. Based on this, we chose Llama 3.1 8B \& 70B \cite{meta_llama_3_1}, Qwen 2.5 72B \cite{qwen_2_5}, and Mistral Large 128B \cite{mistral_large_2407}. All feature 128K-token context windows and excel in reasoning, coding, and multilingual tasks. Llama 3.1 prioritizes efficiency and safety with Llama Guard 3, Qwen 2.5 offers robust multilingual support (29+ languages) and a specialized Coder variant, while Mistral Large 128B supports 80+ programming languages and is optimized for single-node inference. While Llama 3.1 emphasizes open-source accessibility, Qwen 2.5 specializes in multilingual and coding capabilities, and Mistral Large delivers peak computational performance despite its size.

\textbf{LLM$_{gen}$ task.} The four models were applied, in turn, to generate a synthesized summary response, $A$, for a given research question, $Q$, using abstracts from the top $N$ relevant papers. The resulting datasets (downloadable at \url{https://doi.org/10.25835/8dcv2ka6}) constitute the \textbf{``benign'' science Q\&A dataset variant} of this study.


\begin{figure*}[!htb]
    \centering
    \includegraphics[width=\linewidth]{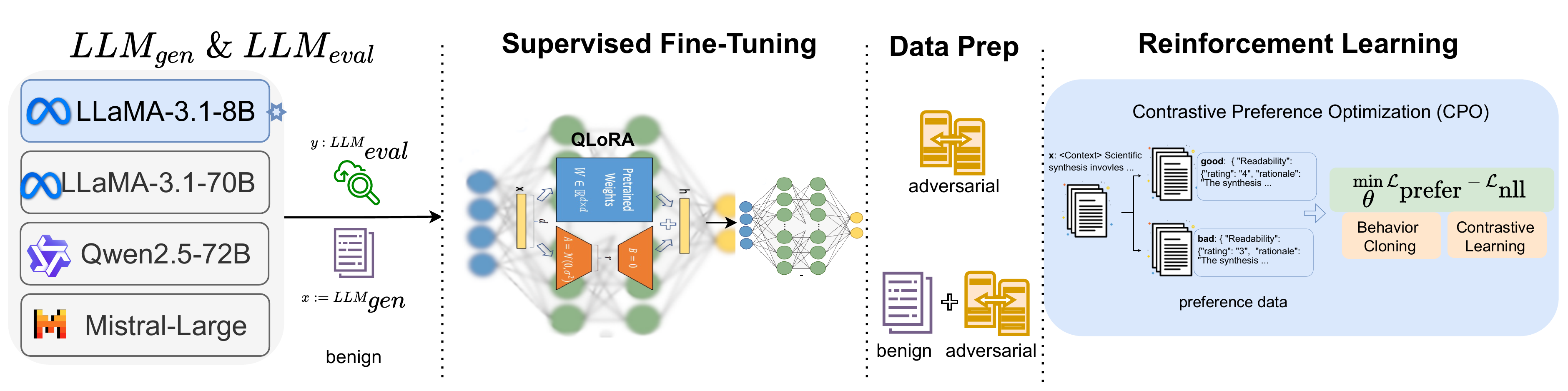}
    \caption{YESciEval LLM-as-a-Judge Alignment: Supervised fine-tuning of $LLM_{eval}$, followed by reinforcement learning via Contrastive Preference Optimization to align open-source LLMs with desired rubric-level evaluations.}
    \label{fig:RLAIF-Methodlogy}
\end{figure*}

\subsection{Our Adversarial ScienceQ\&A Corpus}

Adversarial attacks in NLP generate text samples designed to mislead models into incorrect predictions, using either heuristic-based perturbations or machine learning. Inspired by \citet{jia-liang-2017-adversarial}, who applied heuristic adversarial perturbations to assess NLP brittleness in reading comprehension \citeyearpar{rajpurkar-etal-2016-squad}, we introduce rubric-based adversarial variants of the benign scienceQ\&A dataset to evaluate the robustness of LLM-as-a-judge systems.

A key question in this study is the reliability of LLM$_{eval}$'s qualitative scores for benign synthesized answers \( A \) generated by LLM$_{gen}$. Traditionally, human evaluations would serve as a benchmark, but instead, we assess LLM-as-a-judge robustness through adversarial testing. This approach indirectly measures reliability by evaluating whether LLM$_{eval}$ appropriately differentiates between benign and perturbed responses. If the model fails to adjust its scores accordingly, it suggests an inability to critique responses effectively, thus low reliability on the LLM-as-a-judge.

To systematically evaluate this, we adopt a heuristic-based perturbation strategy, introducing deliberate errors at specific points in the benign text with the expectation that evaluation scores should reflect the resulting quality deterioration. To ensure rigor in adversarial assessment, we design two variants of adversarial attacks: (1) \textit{Subtle adversarial samples}, where minor alterations to the benign text create realistic yet difficult-to-detect errors, and (2) \textit{Extreme adversarial samples}, where substantial modifications make flaws obvious. The adversarial perturbation heuristics, for both variants, were designed at the fine-grained rubric level. Thus each of the nine rubrics have \textit{subtle} and \textit{extreme} adversarial perturbation heuristics associated with them. They are as follows. \textbf{Relevancy} assesses whether the response remains on-topic, with subtle attacks appending sentences from related synthesis paragraphs and extreme attacks injecting unrelated sports news. \textbf{Correctness} tests factual alignment with provided abstracts, using the same attack strategy. \textbf{Completeness} measures how well the response encapsulates relevant content, with subtle attacks removing the final sentence and extreme attacks also appending unrelated text. \textbf{Informativeness} evaluates the response’s utility, using the same perturbation as relevancy. \textbf{Integration} examines structural and linguistic coherence, with subtle attacks removing the first logical connector and extreme attacks eliminating all connectors. \textbf{Cohesion} ensures appropriate sentence connections, with subtle attacks swapping the last two sentences and extreme attacks randomly shuffling them. \textbf{Coherence} assesses logical idea progression, using the same attack as relevancy. \textbf{Readability} focuses on adherence to academic writing conventions, with subtle attacks adding informal blog snippets and extreme attacks inserting tweets. \textbf{Conciseness} tests redundancy, with subtle attacks appending an LLM-generated redundant version of the last sentence and extreme attacks inserting redundant text after every sentence. Detailed heuristics are provided in Appendix \autoref{subsec:adv-corpus}. These perturbations systematically degrade response quality, ensuring rigorous evaluation of LLM-as-a-judge reliability. Redundant responses were LLM-generated and manually refined for fluency. The adversarial \href{https://doi.org/10.25835/8dcv2ka6}{publicly available} dataset benchmarks LLMs' ability to mitigate optimism bias and distinguish response quality. Unlike synonym substitution attacks (SSAs) \cite{alzantot2018generating}, which are conjectured to degrade fluency and meaning \cite{chiang2023synonym}, our rubric-based heuristics introduce syntactic violations to ensure low-quality text. This controlled degradation enables precise evaluation of LLM score adjustments for scienceQ\&A.

\subsection{Our LLM-as-a-judge Evaluations}

The four models—Llama 3.1 8B \& 70B, Qwen 2.5 72B, and Mistral Large 128B—previously used for LLM$_{gen}$ were now tasked with LLM$_{eval}$. Section \ref{sec:task-def} details the task specification prompt. Each model evaluated all instances from ORKGSyn (benign, subtle-adv, extreme-adv) and BioASQ across the nine YESciEval rubrics. This resulted in 37,584 evaluation scores for ORKGSyn (\(348 \times 3 \times 9 \times 4\)) and 7,884 for BioASQ (\(73 \times 3 \times 9 \times 4\)), equating to 9,396 and 1,971 evaluations per model, respectively. Notably, despite requiring GPU compute, model access incurred zero monetary cost—whereas using proprietary models like OpenAI’s GPT would have cost at least 1,000 euros or dollars for these evaluations. This reinforces the motivation of YESciEval, aligned with prior studies \cite{prometheus,prometheus2}, to enhance open-source LLM-as-a-judge models for greater accessibility.

\begin{table}[!htb]
    \footnotesize
    \centering
    \begin{tabular}{|l|c|c|}
        \hline
         & \textbf{\scriptsize BioASQ} & \textbf{\scriptsize ORKGSynthesis} \\
        \hline
        \hline
        \textit{$LLM_{gen}$ Train} & 51 & 234 \\
        \textit{$LLM_{gen}$ Test}  & 22 & 105 \\
        \hline
        $LLM_{eval}$ Train Sets& &\\
        \;\;\;\textit{SFT} & 6,504 & 34,991 \\
        \;\;\;\textit{RL (\texttt{adversarial})} & 1,669 & 6,148 \\
        \;\;\;\textit{RL (\texttt{benign}+\texttt{adversarial})} & 2,569 & 2,290 \\
        $LLM_{eval}$ Test Set & 2,376 & 11,340 \\
        \hline
    \end{tabular}
    \caption{Dataset statistics across training and test sets for $LLM_{gen}$ and $LLM_{eval}$.}
    \label{tab:method-dataset-stats}
\end{table}

\section{The YESciEval Alignment Method}

This work aims to propose a cost-free evaluation framework without human annotators or proprietary models. Key contributions include an adversarial testing strategy for robust LLM-as-a-judge models and the YESciEval alignment method discussed in this section that can be applied to open-source LLMs to equip them as robust evaluators.

The methodology is remotely related to self-instruct \cite{wang2023self}, where LLMs are improved for instruction following self-generated instructions; except we reinforce evaluation behaviors of LLMs to desired behaviors by treating undesired behaviors as negative reward signals. \autoref{fig:RLAIF-Methodlogy} depicts our RL technique applied to learn the alignments between desirable and undesirable behaviors with the following steps:

\noindent\textbf{Supervised Fine-Tuning (SFT).} To ensure model stability after RL, as a first step, we fine-tuned the LLaMA-3.1-8B evaluator using data (see \autoref{tab:method-dataset-stats}) from $x: LLM_{gen}$ (benign scientific syntheses by four LLMs) and corresponding $y: LLM_{eval}$ rubric-wise evaluations. Each rubric score is treated as a separate training sample for SFT. To enable efficient adaptation of large models with minimal compute, we use Quantized Low-Rank Adaptation (QLoRA) \citep{dettmers2023qlora}.


\noindent\textbf{Data Preparation for RL.} Unlike traditional fine-tuning, RL does not require large datasets; instead, models learn from comparisons rather than absolute labels, reducing redundancy and cost \citep{ziegler2019fine}. To this end, the data is divided into benign and adversarial samples, ensuring that our models reinforce the distinction between `good' (desirable) and `bad' (undesirable) evaluations as it is critical for preference modeling \citep{askell2021general}.  We impose a 100-per-rubrics, per-$LLM_{eval}$ threshold (for ORKGSyn,
 this threshold is set to 500 due to the large nature of the task) to maintain a manageable dataset size while preserving diversity across criteria. 
Dataset statistics are shown in \autoref{tab:method-dataset-stats}. For the adversarial sets (``\textit{RL (\texttt{adversarial})}'' row), we define desirable scores as $1$ for extreme and $\leq3$ for subtle adversarial variants across the nine rubrics. This establishes a clear distinction between good and bad evaluations: any $LLM_{eval}$ rating above the threshold is treated as a bad sample; otherwise, it is considered good.
Given only ``\textit{RL (\texttt{adversarial})},'' RL tended to mimic poor examples and struggled to imitate from benign synthesis evaluations, where no adversarial setting is applied. To address this, we curated a separate set containing both benign and adversarial examples (statistics shown in the ``\textit{RL (\texttt{benign+adversarial})}'' row of 
\autoref{tab:method-dataset-stats}), selecting bad examples based on undesirable outcomes in the adversarial evaluations.
The final dataset of $\mathcal{D} = \{x_{LLM_{gen}}^{(i)}, y_{good}^{(i)}, y_{bad}^{(i)} \}_{i=1}^{N}$ is constructed for RL technique.


\noindent\textbf{Reinforcement Learning.} Beyond SFT, reinforcement learning (RL) is used to align LLM evaluations with desirable and undesirable behaviors, moving beyond reference-mimicking. This involves modeling preference data $\mathcal{D}$ using Imitation Learning via Contrastive Preference Optimization (CPO) \citep{xu2024cpo}. CPO extends Direct Preference Optimization (DPO) \cite{rafailov2023dpo} by incorporating hard negative examples, enabling the model to distinguish high- from low-quality outputs. This contrastive approach enhances the model’s ability to prioritize superior responses and reject suboptimal ones, resulting in more discriminative evaluations.
Considering $\pi_{\theta}$ as a parameterized policy, the CPO loss is defined as  $\min_{\theta}  \underbrace{\mathcal{L}(\pi_{\theta}, U)}_{\mathcal{L}_{\text{prefer}}}  - \underbrace{\mathbb{E}_{(x, y_{good}) \sim D} [\log \pi_{\theta}(y_{good} | x)]}_{\mathcal{L}_{\text{NLL}}}$, where $\mathcal{L}_{prefer}$ is a behavior cloning (BC)~\citep{hejna2023contrastive} regularization that encourages the policy $\pi_{\theta}$ to align with uniform prior distribution $U$, derived from reference policy $\pi_{LLM_{eval}}$ by preferring good evaluation $y_{good}$ over bad one $y_{bad}$. Moreover, the term $\mathcal{L}_{NLL}$ is the negative log-likelihood loss that penalizes the policy for making poor generation of the action $y_{good}$ from given state $x$ for high-quality judgments. 

Our resulting aligned models are released on HuggingFace for results reproducibility at \url{https://huggingface.co/SciKnowOrg/YESciEval-ASK-Llama-3.1-8B} and \url{https://huggingface.co/SciKnowOrg/YESciEval-BioASQ-Llama-3.1-8B}.

\begin{figure}[!b]
    \centering
    \includegraphics[width=\linewidth, height=9cm]{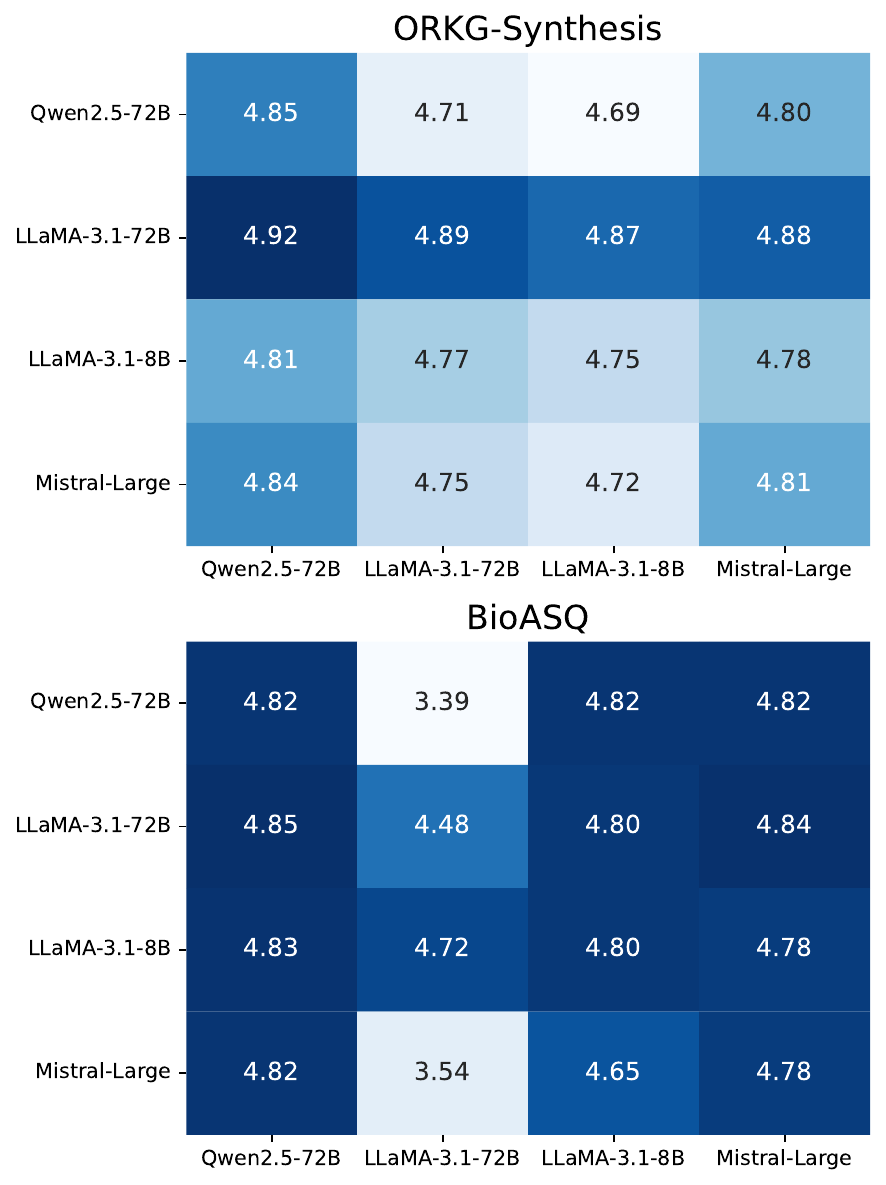}
    \caption{Heatmaps depicting agreement for synthesis evaluations on benign datasets. The x-axis represents the $LLM_{gen}$, while the y-axis denotes the $LLM_{eval}$.}
    \label{fig:enter-label}
\end{figure}

\begin{figure*}[!h]
    \centering
    \includegraphics[width=\linewidth]{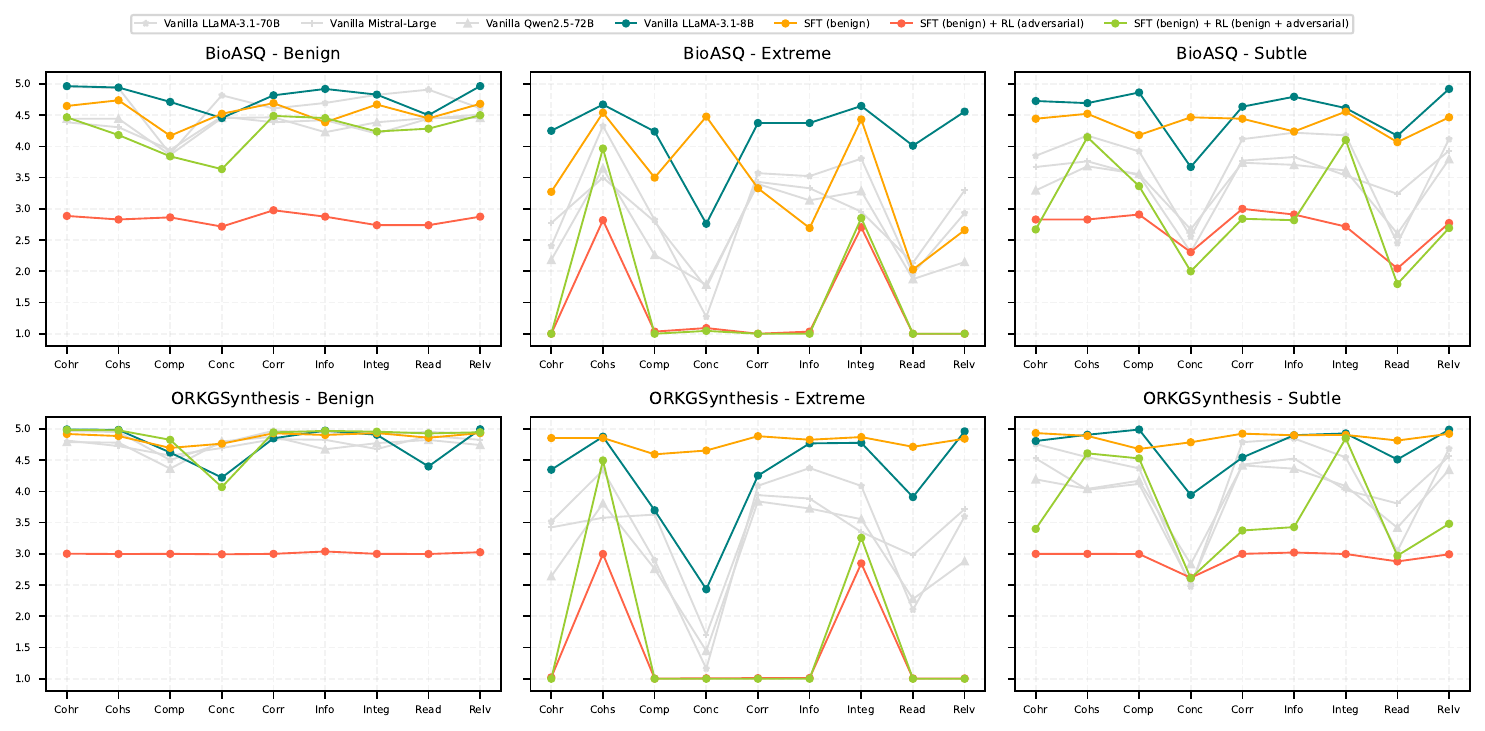}
    \caption{Evaluation of synthesis across different models and fine-tuning strategies on BioASQ and ORKGSynthesis datasets. The nine-rubrics include Coherence (Cohr), Cohesion (Cohs), Completeness (Comp), Concisenes (Conc), Correctness (Corr), Informativeness (Info), Integration (Integ), Readability (Read), and Relevancy (Relv).}
    \label{fig:results-plot}
\end{figure*}

\section{Results and Discussion}

In this section, we systematically analyze the results in relation to the three main research questions outlined in the Introduction. Specifically, we discuss in detail observations on the results obtained from the two-stage process: LLM$_{gen}$ (RQ1) and LLM$_{eval}$ (RQ2); and the application of our YESciEval LLM-as-a-judge alignment method (RQ3). For details on our experimental setup and training, we refer the reader to Appen. \ref{sec:exp-setup} and \ref{sec:train-det}.

First, we focus on: \textbf{RQ1: How similar are scienceQ\&A responses across the three different LLM families when applied as LLM$_{syn}$?} To address this RQ, we measured similarities between benign syntheses generated by the four models for ORKGSyn and BioASQ separately. Based on a comprehensive review of NLG metrics \cite{nlg-eval}, we applied eight diverse similarity metrics: four for verbatim matching, one edit-distance-based, and three embedding-based, e.g., MoverScore \citeyearpar{moverscore}, BERTScore \citeyearpar{bertscore}. To visualize LLM$_{syn}$ correlations per dataset, we computed confusion matrices with averaged similarity scores.

Overall, ORKGSyn consistently yielded higher alignment scores than BioASQ, likely due to its broader domain coverage, with Computer Science (125 questions) as the largest category. As AI and digital fields grow, general-purpose LLMs trained on large datasets, including research papers, exhibit stronger proficiency in these areas. In contrast, BioASQ's biomedical focus, a more specialized domain, led to greater uncertainty and lower correlation scores. Strong correlations emerged within model pairs: Llama 8B and 70B, likely due to their shared Meta origin and training data, differing mainly in parameter size, and Mistral and Qwen, suggesting overlapping training data. These findings highlight the role of shared training data and architecture in aligning model outputs across LLM families. Detailed results: 16 confusion matrices across eight metrics are in \autoref{sec:quant}.

\textbf{RQ2: How do LLM-as-judge evaluations correlate for the benign scienceQ\&A setting?} To address this question, we analyze the results in \autoref{fig:enter-label}, focusing on LLM$_{eval}$ outputs. Each LLM evaluated the benign synthesis dataset created by the four models in their LLM$_{syn}$ roles, with results presented as a confusion matrix where each cell represents the averaged score across all rubrics and synthesis instances, mapping evaluator LLMs (y-axis) against LLM$_{syn}$ models (x-axis). Across both datasets, evaluators assigned higher scores to BioASQ than ORKGSyn, likely due to ORKGSyn's interdisciplinary nature, where LLM$_{syn}$ models struggled with certain domains. Notably, no LLM$_{eval}$ exhibited bias toward its own generated synthesis \cite{li2025preference}. Instead, all evaluators consistently preferred synthesis outputs generated by Qwen. As the largest model, Qwen likely demonstrates superior text generation abilities, reinforcing the reliability of LLM-as-a-judge, even for smaller models like Llama 8B.

\begin{table*}[!htb]
    \centering
    \renewcommand{\arraystretch}{1.1} 
    \setlength{\tabcolsep}{4pt} 
    \small 
    \resizebox{\textwidth}{!}{
    \begin{tabular}{|p{2.5cm}|c|c|c|c|c|}
        \hline
        \textbf{Rubric} & \textbf{G-Eval \cite{liu2023g}} & \textbf{GPTScore \cite{fu2024gptscore}} & \textbf{LLM-Alt \cite{chiang-lee-2023-large}} & \textbf{FLASK \cite{flask}} & \textbf{This YESciEval Work} \\
        \hline
        Coherence & \cellcolor{green}\checkmark & \cellcolor{green}\checkmark & \cellcolor{green}\checkmark &  & \cellcolor{green}\checkmark \\
        Cohesion &  &  & \cellcolor{green}\checkmark &  & \cellcolor{green}\checkmark \\
        Completeness &  & \cellcolor{green}\checkmark &  & \cellcolor{green}\checkmark & \cellcolor{green}\checkmark \\
        Conciseness &  & \cellcolor{green}\checkmark &  & \cellcolor{green}\checkmark & \cellcolor{green}\checkmark \\
        Correctness & \cellcolor{green}\checkmark (Consistency) & \cellcolor{green}\checkmark (Factuality) &  & \cellcolor{green}\checkmark (Logical Correctness) & \cellcolor{green}\checkmark \\
        Informativeness &  & \cellcolor{green}\checkmark &  & \cellcolor{green}\checkmark & \cellcolor{green}\checkmark \\
        Integration &  &  &  &  & \cellcolor{green}\checkmark \\
        Readability & \cellcolor{green}\checkmark (Fluency) & \cellcolor{green}\checkmark & \cellcolor{green}\checkmark & \cellcolor{green}\checkmark & \cellcolor{green}\checkmark \\
        Relevancy & \cellcolor{green}\checkmark & \cellcolor{green}\checkmark & \cellcolor{green}\checkmark &  & \cellcolor{green}\checkmark \\
        Harmlessness &  &  &  & \cellcolor{green}\checkmark &  \\
        Logical Thinking &  &  &  & \cellcolor{green}\checkmark &  \\
        Insightfulness &  &  &  & \cellcolor{green}\checkmark &  \\
        Engagement &  & \cellcolor{green}\checkmark & \cellcolor{green}\checkmark &  &  \\
        Likeability &  & \cellcolor{green}\checkmark & \cellcolor{green}\checkmark &  &  \\
        \hline
    \end{tabular}}
    \caption{Comparison of evaluation rubrics across different works against ours (last column). Cells marked in green indicate rubrics that apply to a particular work.}
    \label{tab:rubric_comparison}
\end{table*}

\textbf{RQ3: Can a smaller open-source LLM be adapted as an LLM-as-a-judge for scienceQ\&A while overcoming optimism bias?} To address this question, we present a comprehensive results plot in \autoref{fig:results-plot}, where each column depicts LLM$_{eval}$ results across three dataset variants, aggregated per rubric for BioASQ (top row) and ORKGSyn (bottom row). Each LLM$_{eval}$ line in the plot corresponds to averaged scores on the synthesis output from the four LLM$_{gen}$ models. The goal is to assess the efficacy of the YESciEval alignment method on a small LLM-as-a-judge or LLM$_{eval}$ model, specifically Llama 8B, under the premise that if effective on a lower-parameter model, it should generalize to larger LLMs. The highlighted colored lines represent different Llama 8B variants: blue (vanilla model), yellow (SFT model trained on benign data), red (SFT + RL with adversarial alignment), and green (SFT + RL with a balanced subsample of benign and adversarial data). Light gray lines indicate the vanilla model performances of the other three LLMs. As hypothesized, vanilla Llama 8B (blue) exhibited excessive optimism, assigning high scores even in extreme adversarial cases—e.g., scoring above 4 on Corr (correctness rubric) despite perturbations introducing unrelated sports news sentences. Fine-tuning on benign data alone (yellow) further amplified optimism, necessitating alignment. When RL was applied only to adversarial data (red), the model became overly pessimistic. However, when RL was trained on both benign and adversarial samples (green), Llama 8B stabilized as a robust evaluator, addressing the RQ. It assigned relatively high scores for benign syntheses while distinguishing adversarial perturbations, scoring around 1 in extreme cases and around 3 in subtle cases, demonstrating rubric-specific discrimination.

\section{Related Work}
\label{sec:rel-work}

\subsection{Question \& Answering Benchmarks}
\label{subsec:rel-work-qa-bench}

Automatic Q\&A spans diverse datasets varying in domain and Q\&A type. Of 41 NLP Q\&A datasets reviewed by \citet{wang2022modern}, only BioASQ aligns with scienceQ\&A. Multiple-choice (e.g., PubMedQA \citeyearpar{jin2019pubmedqa}, MMLU \citeyearpar{hendrycks2021measuring}), Boolean (e.g., BoolQ \citeyearpar{boolq}), and numerics (e.g., Math Dataset \citeyearpar{saxton2018analysing}) fall outside our scope, as do bibliographic \cite{dblp} and knowledge graph extraction datasets \cite{sciqa,yan2024biomed}. Existing benchmarks, such as \href{https://huggingface.co/spaces/open-llm-leaderboard/open_llm_leaderboard#/}{Hugging Face leaderboard} tasks \cite{wang2024mmlupro,rein2023gpqa} and alignment-focused chat-based evaluations \cite{reddy2019coqa,zheng2023judging,kopf2024openassistant}, primarily assess multiple-choice reasoning or human preference alignment. In contrast, we introduce a generative scienceQ\&A dataset, filling a gap in current benchmarking efforts.

\subsection{LLM-as-a-judge Evaluation Rubrics}
\label{subsec:rel-work-eval-rub}

LLM-as-a-judge \cite{zheng2023judging} initially focused on correlating LLM evaluations with human judgments in open-domain NLG, primarily using pairwise preference evaluations \cite{wang2023chatgpt,chiang-lee-2023-large,dubois2023alpacafarm,liu2023g}. Some works incorporated rubrics, such as G-Eval \cite{liu2023g} for summarization and GPTScore \cite{fu2024gptscore}, which aligns closely with our criteria. Recent frameworks emphasize fine-grained rubrics; FLASK \cite{flask} assesses robustness, correctness, efficiency, factuality, and readability, of which nine align with our work. Prometheus \cite{prometheus,prometheus2} expands rubric-based evaluation but relies on human references, whereas we use adversarial data to refine LLM evaluations without annotations. ScienceQ\&A assessment evolved from three core criteria—comprehensiveness, trust, and utility \cite{core-gpt}—to the nine rubrics adopted in this work based on our prior work purely on LLMs for scientific synthesis tasks \cite{llms4synthesis}. 
Decoupling reliance on human references, 
our rubric-based adversarial approach provides a systematic, cost-free framework for scienceQ\&A evaluation.

A comparative summary of widely used LLM-as-a-judge rubrics is provided in \autoref{tab:rubric_comparison}, with an extended discussion available in Appendix~\ref{sec:ext-rel-work-llm-as-a-judge}.

\section{Conclusion}


YESciEval is a reproducible, cost-free LLM-as-a-judge framework for evaluating NLG in scienceQ\&A, advancing AI alignment, robustness, and the broader agenda of LLM plausibility—key factors toward artificial general intelligence (AGI).

\section*{Acknowledgements}

This work is jointly supported by the \href{https://scinext-project.github.io/}{SCINEXT project} (BMBF, Grant ID: 01IS22070), the KISSKI AI Service Center (BMBF, ID: 01IS22093C), and \href{https://www.nfdi4datascience.de/}{NFDI4DataScience} (DFG, ID: 460234259).

\section{Limitations}

While our approach offers a robust, cost-free framework for LLM-as-a-judge evaluation in scienceQ\&A, certain methodological choices present limitations and avenues for future enhancement. One such direction is the integration of \textbf{chain-of-thought (CoT) reasoning} \cite{wei2022chain}, which has shown effectiveness in structured reasoning tasks such as mathematics and logic (e.g., SelfCheck \cite{miao2024selfcheck}). Although CoT is not yet widely adopted in evaluation settings, it may improve judgment quality by encouraging more structured and transparent decision-making. We intend to explore CoT-based evaluators in future work.

We also opted for reinforcement learning (RL) over \textit{few-shot in-context learning} (ICL) \cite{brown2020language} to improve evaluation robustness. While ICL—where models are conditioned on exemplars, including both good and bad outputs \cite{fu2024gptscore}—is a compelling alternative, it presents practical limitations. With \textbf{nine rubrics} in our setup, ICL would necessitate long prompts potentially exceeding model context limits. Moreover, curating high-quality negative examples would require manual annotation, which conflicts with our goal of a fully automated, zero-cost alignment framework. Nonetheless, we recognize the promise of ICL and aim to investigate more efficient adaptations in future iterations.

Finally, while our results on the ORKGSynthesis and BioASQ benchmarks confirm YESciEval’s effectiveness in scienceQ\&A, broader generalization remains an open question. Notably, our dataset—available at \url{https://doi.org/10.25835/8dcv2ka6}—particularly the portion derived from the Open Research Knowledge Graph (ORKG), constitutes a strong, domain-diverse test bed \cite{yescieval-corpus}. It features research questions submitted by domain experts spanning 33 distinct scientific fields. \autoref{fig:dataset-examples} illustrates examples from the domains of Physics and Oceanography, highlighting the dataset’s richness and complexity. This makes our benchmark both novel and distinctive within the evaluation landscape. Nonetheless, future work will assess scalability by extending evaluations to substantially larger corpora and exploring the generalizability of our methodology to generative AI tasks beyond scienceQ\&A.

\bibliography{custom}

\appendix

\section{Extended Related Work}
\label{sec:ext-rel-work}

\subsection{Question \& Answering Benchmarks}

Automatic Q\&A remains a rapidly evolving field, with an expanding array of datasets supporting its development. These datasets exhibit wide variation in domains, question types, and generation methodologies. To contextualize the scienceQ\&A focus of this paper, we first provide an overview of related datasets and highlight their distinctions.

A comprehensive review of NLP Q\&A datasets prior to 2022 by \citet{wang2022modern} outlines multiple dataset categories. However, datasets using multiple-choice answer formats fall outside the scope of this work, including MCTest \citeyearpar{richardson2013mctest} for fictional stories, ARC \citeyearpar{clark2018think} for high-school science exams, OpenBookQA \citeyearpar{mihaylov2018can} for science facts, PubMedQA \citeyearpar{jin2019pubmedqa} for medical summarization, and LogiQA \citeyearpar{liu2021logiqa} for logical reasoning in exams. Similarly, datasets with Boolean answer types (e.g., BoolQ \citeyearpar{boolq}) and numeric result datasets (e.g., Mathematics Dataset \citeyearpar{saxton2018analysing}) are not within our scope. Even datasets requiring entity-based answers from structured knowledge sources, such as ComplexWebQuestions \citeyearpar{talmor2018web}, diverge from our focus on generative scienceQ\&A. Among the 41 reviewed datasets, only the 2023 release of BioASQ was found to be directly relevant to our research objective.

Additional scientific Q\&A datasets fall outside our research scope, such as DBLP-QuAD \cite{dblp} for bibliographic queries and datasets designed to extract factual knowledge from scientific literature for knowledge graph population \cite{sciqa,yan2024biomed}. One notable dataset is DBLP-QuAD \cite{dblp}, which contains 10,000 QA pairs generated via SPARQL queries over the DBLP scholarly knowledge graph (KG). While valuable for bibliographic metadata QA, its focus is restricted to bibliographic queries, limiting its applicability to more diverse or conceptual scholarly questions. Another dataset we evaluated is the SciQA benchmark \cite{sciqa}, which includes 100 handcrafted complex QA pairs alongside 2,465 automatically generated ones. These questions are derived from the Open Research Knowledge Graph (ORKG). However, the dataset's reliance on ORKG-specific entities and the need for direct KG access to produce high-quality answers posed practical challenges for our study, leading us to exclude it. We also considered the BioKGQA dataset, proposed by \citet{yan2024biomed}, which features 85,368 QA pairs generated using multi-noded triples from PrimeKG, a comprehensive KG oriented toward precision medicine. While its method of leveraging KG structures for QA generation is innovative, the dataset primarily focuses on fact-based answers, making it less suitable for exploring nuanced or multi-dimensional evaluation.

The \href{https://huggingface.co/spaces/open-llm-leaderboard/open_llm_leaderboard#/}{Hugging Face leaderboard} serves as a widely recognized benchmark space for new LLMs. Among its datasets, MMLU-PRO \cite{wang2024mmlupro} provides expert-reviewed multiple-choice questions across diverse domains, including Medicine, Law, Engineering, and Mathematics. GPQA \cite{rein2023gpqa} similarly includes multiple-choice questions authored by domain experts in Biology, Physics, and Chemistry. However, these benchmarks primarily evaluate intrinsic reasoning ability in answering fixed-choice questions rather than generative Q\&A tasks. Recent LLM evaluation trends emphasize human-aligned benchmarks for chat assistant alignment, such as MT-Bench and Chatbot Arena \cite{zheng2023judging}, which assess open-domain, multi-turn dialogue abilities. These benchmarks evaluate capabilities in writing, reasoning, extraction, and domain knowledge (e.g., STEM and humanities/social sciences), but their focus remains distinct from our objective. Instead, our work aims to develop a framework that enables open-source LLMs to robustly evaluate generative scienceQ\&A using standardized, rubrics-based multifaceted assessments, further pushing the frontier of LLM-as-a-judge capabilities.

\subsection{LLM-as-a-judge Evaluation Rubrics}
\label{sec:ext-rel-work-llm-as-a-judge}

The early notion of LLM-as-a-judge \cite{zheng2023judging} measured the correlation between an LLM judge and human evaluators in open-domain instruction following for NLG. Early works on using LLMs as evaluators of LLM-generated text emphasized pairwise evaluation along a single dimension of `preference' in determining which response was superior \cite{wang2023chatgpt,chiang-lee-2023-large,dubois2023alpacafarm,zheng2023judging,liu2023g,kocmi2023large}.

In these preference comparisons, evaluation rubrics emerged. G-Eval \cite{liu2023g} incorporated criteria such as coherence, consistency, fluency, and relevance for summarization benchmarking. The rubrics of GPTScore \cite{fu2024gptscore} for summarization tasks also align with ours. However, GPTScore also includes dialogue rubrics aligned with instruction-following chat-based evaluations such as likeability, flexibility, inquisitiveness, and engagement, which are out of scope for scienceQ\&A. Open-ended story generation rubrics \cite{chiang-lee-2023-large} share similarities with scientific QA, where likeability is juxtaposed against informativeness, relevance to answer pertinence, and grammar to cohesion, integration, and readability.

ScienceAgentBench \citep{chen2024scienceagentbench} employs four rubric-based metrics to evaluate LLMs and agentic AI for generating programmatic workflows suited to scientific data science applications. Their evaluation rubrics include Success Rate (SR) (task completion), Valid Execution Rate (VER) (program runs without error), CodeBERTScore (CBS) (similarity to reference implementations), and Cost Efficiency (measuring API expenses). They compiled data science benchmarks across Bioinformatics, Computational Chemistry, Geographical Information Science, and Psychology \& Cognitive Neuroscience. Their work recommends automating rubric-based evaluations for generated code quality, a gap we address differently in this paper.

SciCode \citep{tianscicode} sourced research-level coding problems across 16 subfields in natural science disciplines, benchmarking LLMs on pythonic code generation from docstring-format task descriptions. It relied on pure quantitative metrics such as pass@1 rate but lacked detailed rubric-based insights.

More recent works examine fine-grained rubrics. The FLASK \cite{flask} evaluation rubrics—Robustness, correctness, efficiency, factuality, commonsense, comprehension, insightfulness, completeness, metacognition, readability, conciseness, and harmlessness—are closely related to our work. Nine of their 12 rubrics align with ours, except insightfulness, harmlessness, and metacognition, which lack clear definitions for LLMs. Prometheus \citep{prometheus,prometheus2} developed 1,222 customized rubrics across open-domain benchmarks but relied on human reference answers. Instead, this work circumvents human annotation dependence by leveraging adversarial data where the desired behavior is known.

LLM-based scientific QA evaluation evolved from three criteria \cite{core-gpt,evans2024large}—comprehensiveness, trust, and utility—to nine \cite{llms4synthesis}, including relevancy, correctness, completeness, informativeness, integration, cohesion, readability, and conciseness. 

HELM \cite{liang2023holistic} set a precedent in holistic language model evaluation by considering seven quantitative metrics: accuracy, calibration, robustness, fairness, bias, toxicity, and efficiency. This multi-metric approach exposed trade-offs across different evaluation criteria, ensuring that accuracy was not the sole measure of performance. Inspired by this, our work targets a holistic evaluation of scienceQ\&A, defining a multifaceted qualitative framework of nine rubrics that comprehensively assess LLM outputs.

This work advances the field by proposing a structured evaluation methodology that moves beyond traditional ROUGE/BLEU metrics and proprietary model dependencies. By equipping open-source LLMs with robust evaluation capabilities, we eliminate reliance on expensive human annotations and proprietary LLM-generated reference answers. Our approach ensures that the evaluation of scientific QA models remains transparent, replicable, and fine-grained, aligning with broader efforts to standardize LLM assessments.

\section{Science Q\&A Datasets}
\label{sec:ext-scienceqa}

\subsection{The Multidiscplinary ORKGSyn Corpus}
\label{subsec:ext-orkgsyn}

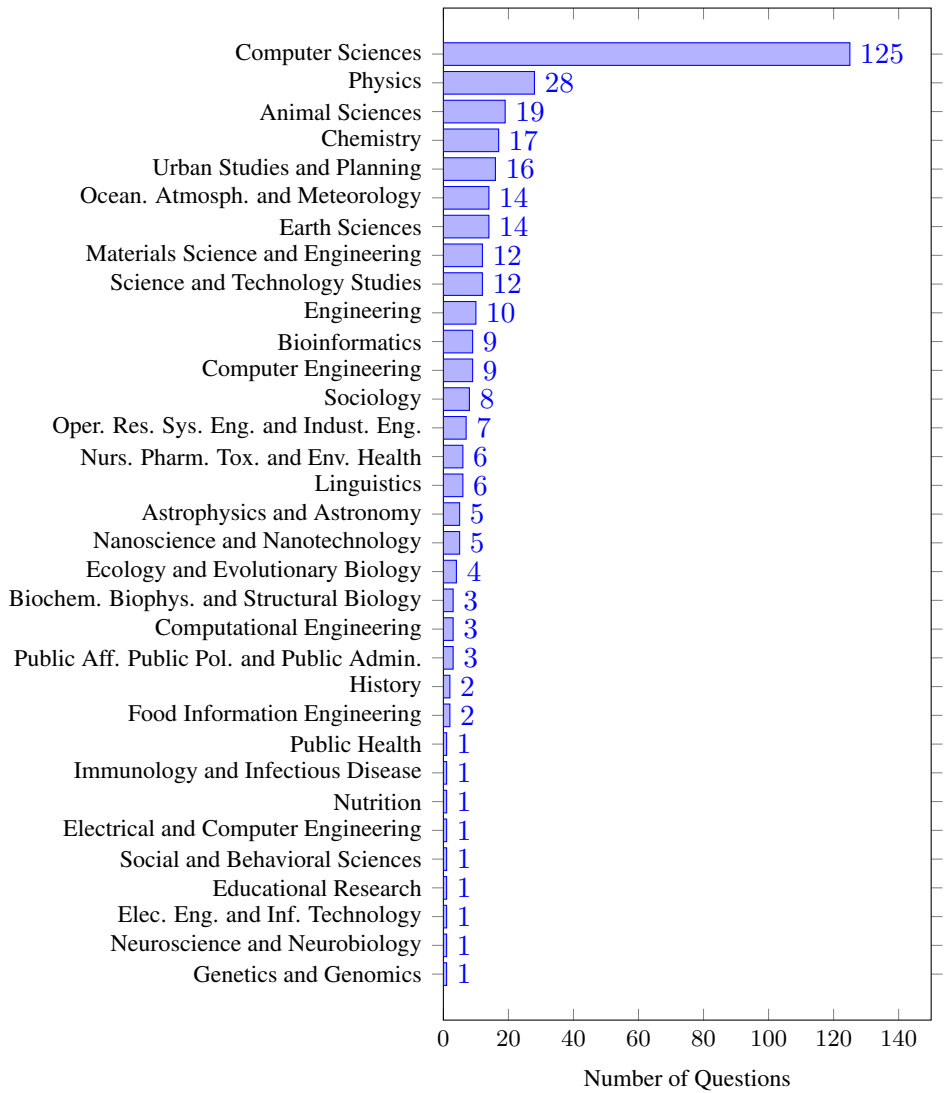
\begin{figure*}[!htb]
    \centering
    \begin{tikzpicture}
        \begin{axis}[
            xbar,
            symbolic y coords={
                {Genetics and Genomics}, 
                {Neuroscience and Neurobiology}, 
                {Elec. Eng. and Inf. Technology}, 
                {Educational Research}, 
                {Social and Behavioral Sciences}, 
                {Electrical and Computer Engineering}, 
                {Nutrition}, 
                {Immunology and Infectious Disease}, 
                {Public Health}, 
                {Food Information Engineering}, 
                {History}, 
                {Public Aff. Public Pol. and Public Admin.}, 
                {Computational Engineering}, 
                {Biochem. Biophys. and Structural Biology}, 
                {Ecology and Evolutionary Biology}, 
                {Nanoscience and Nanotechnology}, 
                {Astrophysics and Astronomy}, 
                {Linguistics}, 
                {Nurs. Pharm. Tox. and Env. Health}, 
                {Oper. Res. Sys. Eng. and Indust. Eng.}, 
                {Sociology}, 
                {Computer Engineering}, 
                {Bioinformatics}, 
                {Engineering}, 
                {Science and Technology Studies}, 
                {Materials Science and Engineering}, 
                {Earth Sciences}, 
                {Ocean. Atmosph. and Meteorology}, 
                {Urban Studies and Planning}, 
                {Chemistry}, 
                {Animal Sciences}, 
                {Physics}, 
                {Computer Sciences}
            },
            ytick=data,
            xmin=0,
            xmax=150,
            xlabel={Number of Questions},
            nodes near coords,
            bar width=0.3cm,
            enlarge y limits=0.05,
            width=8cm,
            height=15cm,
            legend pos=south east,
            tick label style={font=\footnotesize},
            label style={font=\footnotesize},
        ]
            \addplot coordinates {
                (1,{Genetics and Genomics}) 
                (1,{Neuroscience and Neurobiology}) 
                (1,{Elec. Eng. and Inf. Technology}) 
                (1,{Educational Research}) 
                (1,{Social and Behavioral Sciences}) 
                (1,{Electrical and Computer Engineering}) 
                (1,{Nutrition}) 
                (1,{Immunology and Infectious Disease}) 
                (1,{Public Health}) 
                (2,{Food Information Engineering}) 
                (2,{History}) 
                (3,{Public Aff. Public Pol. and Public Admin.}) 
                (3,{Computational Engineering}) 
                (3,{Biochem. Biophys. and Structural Biology}) 
                (4,{Ecology and Evolutionary Biology}) 
                (5,{Nanoscience and Nanotechnology}) 
                (5,{Astrophysics and Astronomy}) 
                (6,{Linguistics}) 
                (6,{Nurs. Pharm. Tox. and Env. Health}) 
                (7,{Oper. Res. Sys. Eng. and Indust. Eng.}) 
                (8,{Sociology}) 
                (9,{Computer Engineering}) 
                (9,{Bioinformatics}) 
                (10,{Engineering}) 
                (12,{Science and Technology Studies}) 
                (12,{Materials Science and Engineering}) 
                (14,{Earth Sciences}) 
                (14,{Ocean. Atmosph. and Meteorology}) 
                (16,{Urban Studies and Planning}) 
                (17,{Chemistry}) 
                (19,{Animal Sciences}) 
                (28,{Physics}) 
                (125,{Computer Sciences})
            };
        \end{axis}
    \end{tikzpicture}
    \caption{Number of Questions per Research Field on the ORKGSyn Dataset. The y-axis represents the "Research Fields".}
    \label{fig:orkgsynresearchfields}
\end{figure*}

The ORKGSyn corpus is a highly multidisciplinary scienceQ\&A dataset. In this context, \autoref{fig:orkgsynresearchfields} illustrates the distribution of instances across 34 different scientific disciplines represented in ORKGSyn.

\subsection{The BioASQ Corpus}
\label{subsec:ext-bioasq}

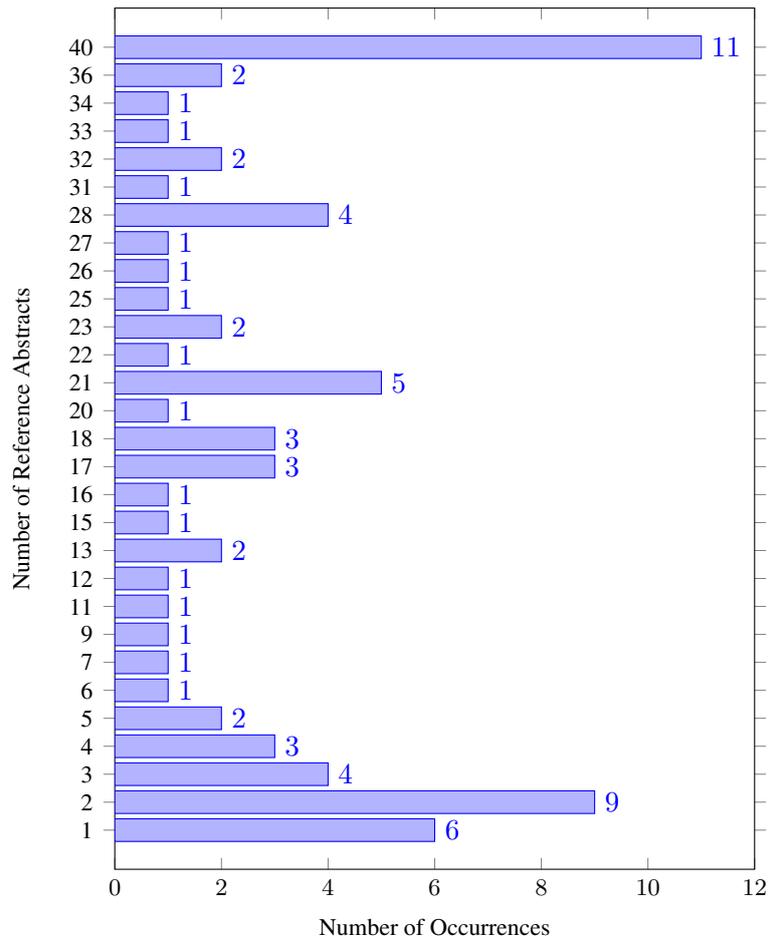
\begin{figure*}[!htb]
    \centering
    \begin{tikzpicture}
        \begin{axis}[
            xbar,
            symbolic y coords={
                1, 2, 3, 4, 5, 6, 7, 9, 11, 12, 13, 15, 16, 17, 18, 20, 21, 
                22, 23, 25, 26, 27, 28, 31, 32, 33, 34, 36, 40
            },
            ytick=data,
            xmin=0,
            xmax=12,  
            xlabel={Number of Occurrences},
            ylabel={Number of Reference Abstracts},
            nodes near coords, 
            bar width=0.3cm,  
            enlarge y limits=0.05,  
            width=10cm,  
            height=13cm,  
            tick label style={font=\footnotesize},  
            label style={font=\footnotesize},  
        ]
            \addplot coordinates {
                (6,1) (9,2) (4,3) (3,4) (2,5) (1,6) (1,7) (1,9) (1,11) (1,12)
                (2,13) (1,15) (1,16) (3,17) (3,18) (1,20) (5,21) (1,22) (2,23) 
                (1,25) (1,26) (1,27) (4,28) (1,31) (2,32) (1,33) (1,34) (2,36) (11,40)
            };
        \end{axis}
    \end{tikzpicture}
    \caption{Number of Reference Abstracts per Question on the BioASQ dataset}
    \label{fig:bioasqabstractscount}
\end{figure*}

According to the scienceQ\&A task definition, ORKGSyn consistently linked each \( Q \) to \( N = 5 \) papers with abstracts, whereas in BioASQ, \( N \) varied between 1 and 40. \autoref{fig:bioasqabstractscount} presents the distribution of instances, categorized by the number of papers associated with each \( Q \).

\subsection{Our Adversarial Corpus}
\label{subsec:adv-corpus}

By constructing an adversarial dataset, we introduce deliberate errors into the original outputs to test whether LLMs can detect and evaluate poor-quality responses. This allows for a comparative analysis of evaluation scores between original and manipulated datasets. While original syntheses may not always be flawless, adversarial datasets are designed to exhibit a marked deterioration in quality, and we expect evaluation scores to reflect this decline. 
We construct two tiers of adversarial datasets. 
\begin{enumerate}
    \item \textbf{Subtle Adversarial Dataset:} Here, reference texts are minimally altered, making it challenging for models to detect changes. These alterations mimic realistic errors that may go unnoticed in automated evaluations.
    \item \textbf{Extreme Adversarial Dataset:} This dataset involves substantial modifications to reference texts, making the adversarial setting apparent and straightforward for models to identify. The evaluations should result in significantly lower scores.
\end{enumerate}
Nine evaluation criteria are systematically targeted during adversarial dataset creation, with distinct manipulations tailored to degrade the corresponding aspect of synthesis quality. To simulate varying degrees of distortion, adversarial sentences were drawn from diverse sources: a blog post snippet from \cite{canmachinesthink}, a sentence from a sports news article provided by \cite{sportsnews}, and a tweet sourced from \cite{twitterwnut16}. Below, we detail the adversarial interventions and describe what each criterion evaluates:

\begin{enumerate}
    \item \textbf{Relevancy:} Is the information in the answer relevant to the problem?
    \begin{itemize}
        \item[--] Subtle: Append a sentence from a different synthesis paragraph within the same domain.
        \item[--] Extreme: Append a sentence from an unrelated sports news article.
    \end{itemize}
    \item \textbf{Correctness:} Is the information in the answer a correct representation of the content of the provided abstracts? 
    \begin{itemize}
        \item[--] Subtle: Append a sentence from a different synthesis paragraph within the same domain.
        \item[--] Extreme: Append a sentence from an unrelated sports news article.
    \end{itemize}
    \item \textbf{Completeness:} Is the answer a comprehensive encapsulation of the relevant information in the provided abstracts? 
    \begin{itemize}
        \item[--] Subtle: Remove the last sentence from the synthesis.
        \item[--] Extreme: Remove the last sentence and append a sentence from an unrelated sports news article.
    \end{itemize}
    \item \textbf{Informativeness:} Is the answer a useful and informative reply to the problem?
    \begin{itemize}
        \item[--] Subtle: Append a sentence from a different synthesis paragraph within the same domain.
        \item[--] Extreme: Append a sentence from an unrelated sports news article.
    \end{itemize}
    \item \textbf{Integration:} Are the sources structurally and linguistically well-integrated, using appropriate markers of provenance/quotation and logical connectors for each reference?
    \begin{itemize}
        \item[--] Subtle: Remove the first logical connector (e.g., "however", "therefore").
        \item[--] Extreme: Remove all logical connectors.
    \end{itemize}
    \item \textbf{Cohesion:} Are the sentences connected appropriately to make the resulting synthesis cohesive? 
    \begin{itemize}
        \item[--] Subtle: Swap the positions of the last two sentences.
        \item[--] Extreme: Randomly shuffle all sentences.
    \end{itemize}
    \item \textbf{Coherence:} Are the ideas connected soundly and logically?
    \begin{itemize}
        \item[--] Subtle: Append a sentence from a different synthesis paragraph within the same domain.
        \item[--] Extreme: Append a sentence from an unrelated sports news article.
    \end{itemize}
    \item \textbf{Readability:} Does the answer follow appropriate style and structure conventions for academic writing, particularly for readability?
    \begin{itemize}
        \item[--] Subtle: Append a snippet from a casual blog post.
        \item[--] Extreme: Append a sentence from an informal tweet.
    \end{itemize}
    \item \textbf{Conciseness:} Is the answer short and clear, without redundant statements?
    \begin{itemize}
        \item[--] Subtle: Use the LLM to generate a redundant version of the last sentence and append it to the response.
        \item[--] Extreme: Append a redundant version after every sentence in the original response.
    \end{itemize}
\end{enumerate}
For the conciseness criterion, redundant sentences were generated by prompting the original model to unnecessarily rephrase and extend its outputs. Despite specific instructions to avoid introductory phrases or comments about the redundancy task, the Llama models frequently failed to adhere to these guidelines, requiring extensive manual cleanup. For instance, phrases such as "Here is a redundant sentence..." were frequently appended to the generated responses. We manually revised over 100 responses generated by the Llama 8B model and 18 by the Llama 70B. This process culminated in the creation of adversarial datasets with subtle and extreme degrees of manipulation for each underlying dataset, resulting in a total of four datasets. By systematically introducing errors tailored to each quality criterion, these datasets enable a detailed evaluation of LLM performance across a range of distortions. This two-level adversarial approach provides a comprehensive method for assessing the models' sensitivity to various levels of quality deterioration and their ability to assign appropriate evaluation scores under varying conditions.

One of the early papers on LLM-as-a-judge \cite{zheng2023judging} also used the strategy of an adversarial attack to test whether the judge LLM could detect verbosity bias which is related to our attack for redundancy.

A key NLP line of research in text adversarial attacks involves synonym substitution attacks (SSAs) \cite{alzantot2018generating} generate adversarial samples by replacing words with synonyms in benign text, relying on sophisticated methods such as TextFooler \cite{jin2020bert}, PWWS \cite{ren2019generating}, and BAE \cite{garg2020bae}. While SSAs were conjectured to produce low-quality text in fluency and meaning \cite{hauser2021bert,chiang2023synonym}, there was no guarantee of this, necessitating human raters to compare their assessments against LLM scores. In contrast, our adversarial attacks, ranging from subtle to extreme, deliberately violate English syntax, ensuring inherently low-quality text and obviating the need for human raters. This design directly tests whether the LLM can overcome its optimism bias and robustly assign the expected low scores.

\section{Manual Subsample Observations of Vanilla LLM$_eval$}
\label{sec:manual-eval}

We conducted a manual analysis of evaluation scores generated by Llama 8B, using a structured annotation process to assess the model’s performance. The human adjudicator categorized issues based on six criteria from \cite{prometheus}: “rejected feedback is not consistent with its score,” “too general and abstract,” “overly optimistic,” “not relevant to the response,” “overly critical,” and “unrelated to the score rubric.” This analysis covered 20 questions from ORKGSyn and 10 from BioASQ.

Our findings reveal that Llama 8B’s evaluations of vanilla syntheses are generally aligned with human judgement. However, the feedback provided by the model tends to be overly general and abstract. For instance, while Llama 8B often identifies a lack of minor details, it frequently fails to specify what exactly is missing. In the adversarial settings, Llama 8B exhibits overly optimistic scoring across all syntheses. This pattern is consistent for both subtle and extreme adversarial datasets, though the ORKGSyn dataset receives even higher scores compared to BioASQ. Additionally, we observed that in a small fraction of cases, the scores were not relevant to the response. For example, Llama 8B occasionally hallucinated its own evaluation criteria and scored based on these fabricated metrics. Another notable issue is the inconsistency between the rationale provided and the assigned scores. In a few cases, the evaluation feedback explicitly states that there is almost no unnecessary information in the synthesis, yet the model assigns a perfect score of 5. Furthermore, for a significant portion of the scores, the rationale was very general and abstract, often merely reciting the evaluation characteristic guidelines provided in the system prompt without offering specific insights.

\section{Detailed Quantitative Experimental Results}
\label{sec:quant}
To obtain a comprehensive assessment of the LLM's performance on the scientific Q\&A task, we conducted a quantitative analysis of their outputs. This evaluation leverages \textit{summarisation-based} (such as BLUE~\citep{bleu}, ROUGE~\citep{rouge}, METEOR~\citep{meteor}, NIST~\citep{nist}, and BERTScore~\citep{bertscore}) and \textit{edit distance-based} (such as WER~\citep{wer}, WMD~\citep{wmd}, and MoverScore~\citep{moverscore}) metrics.  These metrics measure the similarity between text outputs, providing a quantitative estimate of how closely the generated responses align with one another. The evaluation process involves treating each LLM response as a reference and pairing it with the outputs of every other model as candidates, resulting in all possible pairwise combinations. This approach ensures that we capture not only the absolute performance of each model but also their relative alignment. By examining the similarity scores between models, we gain insights into how consistently they generate responses.

\subsection{Summarisation-Based Metrics}
Evaluating the quality of generated text often requires comparing it to reference summaries. To achieve this, summarisation-based evaluation metrics measure the degree of textual overlap between the generated and reference texts, typically assessing aspects such as precision, recall, and n-gram similarity. These metrics are widely used in NLP tasks such as machine translation, text summarisation, and question answering. The summarisation-based metrics (i.e. BLUE, ROUGE, METEOR, NIST, and BERTScore) are described as follows.

\noindent\textbf{Bilingual Evaluation Understudy (BLEU). } The BLEU~\citep{bleu} metric automates machine translation evaluation, offering a cost-effective alternative to human assessment. It measures n-gram overlap between a candidate and reference text, with a brevity penalty to prevent short translations. While widely used, BLEU has limitations, including insensitivity to semantic meaning and struggles with short texts.
\begin{figure}[t]  
    \centering
    \begin{subfigure}[b]{0.48\textwidth}
        \centering
        \includegraphics[width=\textwidth]{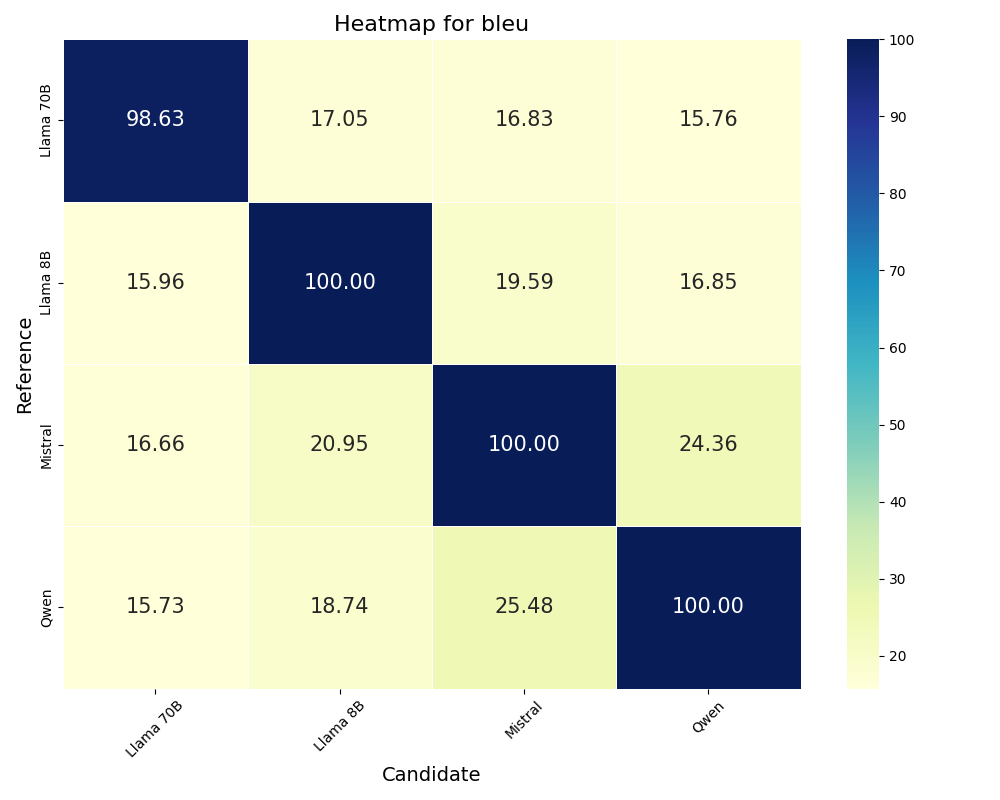}
        \caption{BioASQ}
        \label{fig:bioasq_bleu}
    \end{subfigure}
    \hfill
    \begin{subfigure}[b]{0.48\textwidth}
        \centering
        \includegraphics[width=\textwidth]{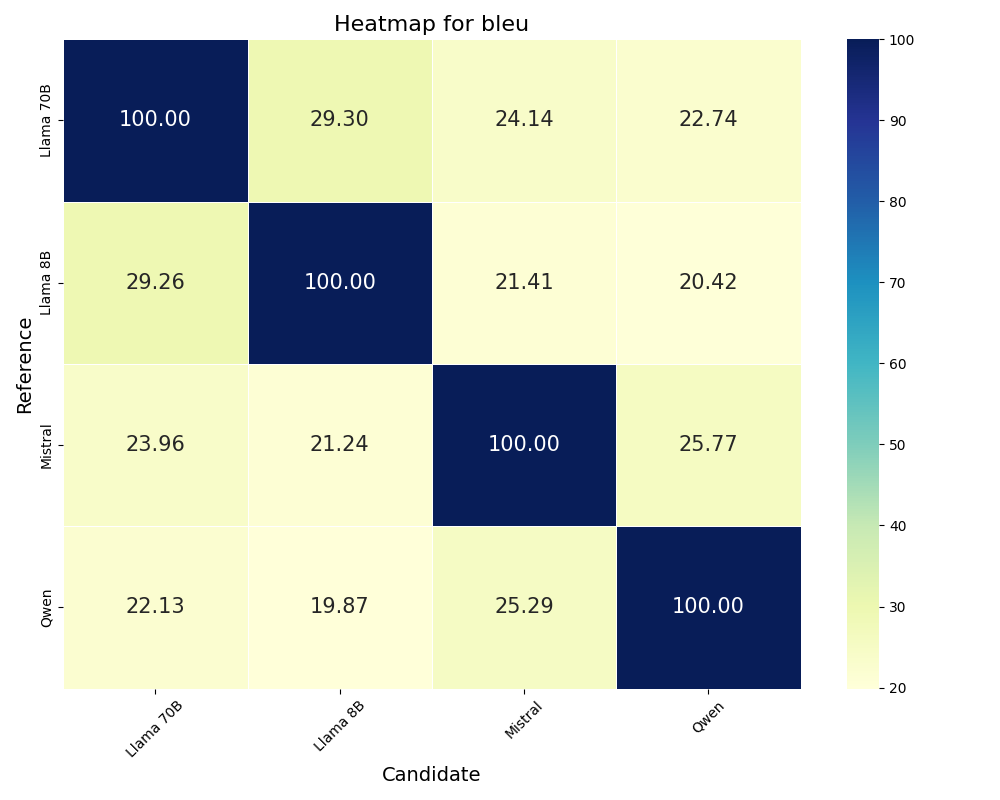}
        \caption{ORKGSynthesis}
        \label{fig:llm4syn_bleu}
    \end{subfigure}
    \caption{BLEU metric results.}
    \label{fig:blue-metric-results}
\end{figure}
According to the \autoref{fig:blue-metric-results}, an anomaly occurs for the \textit{LLaMA-3.1-70B} model on the BioASQ dataset, where the score is \textit{98.63}. This discrepancy arises because the BLUE metric averages over n-gram scores, including cases where higher-order n-grams (e.g., four-grams) are absent in shorter outputs, forcing those scores to zero. Interestingly, the results reveal patterns of correlation among specific models. For instance, \textit{Mistral-Large} and \textit{Qwen2.5-72B} exhibit moderate alignment, with scores around 25, indicating some similarity in their outputs. On the ORKGSynthesis dataset, the \textit{LLaMA-3.1-70B} and \textit{LLaMA-3.1-8B} models display a relatively high correlation, likely due to their shared Meta origin, similar training data, and primary differences in parameter count. In terms of individual performance, \textit{Mistral-Large} achieves the highest average  BLUE score on the BioASQ dataset, while on the ORKGSynthesis dataset, Meta's 70B LLM performs best.
 
\noindent\textbf{Recall-Oriented Understudy for Gisting Evaluation (ROUGE). }ROUGE~\citep{rouge} is a recall-based metric contrasting with BLEU’s precision focus. It measures n-gram, word sequence, and word pair overlap between machine-generated and human summaries. ROUGE-N emphasizes n-gram recall, while ROUGE-L captures the longest common subsequence (LCS) for better semantic similarity. Variants like ROUGE-W reward consecutive matches, and ROUGE-S use skip-bigrams for F-measure calculation. For this analysis, we used ROUGE-1 for unigram overlap, reflecting term alignment and content coverage, and ROUGE-L to assess sentence-level structure and coherence.

\begin{figure}[t]
    \centering
    \begin{subfigure}[b]{0.48\textwidth}
        \centering
        \includegraphics[width = \textwidth]{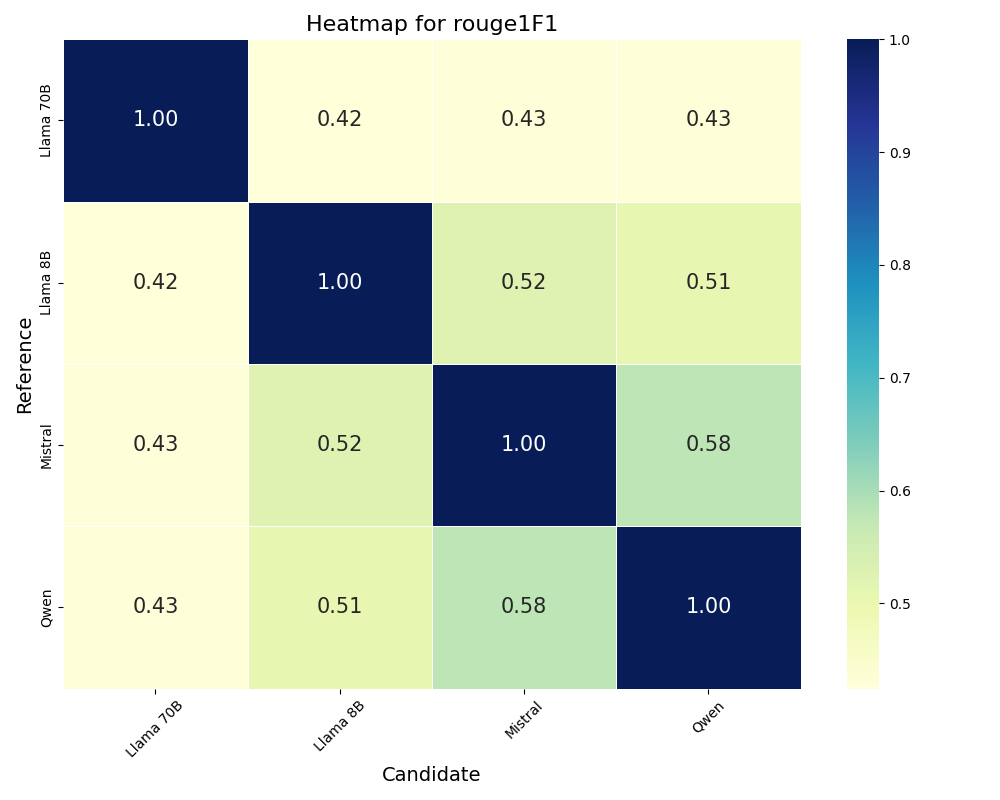}
        \caption{BioASQ}
        \label{fig:bioasq_rouge1}
    \end{subfigure}
    \begin{subfigure}[b]{0.48\textwidth}
        \centering
        \includegraphics[width = \textwidth]{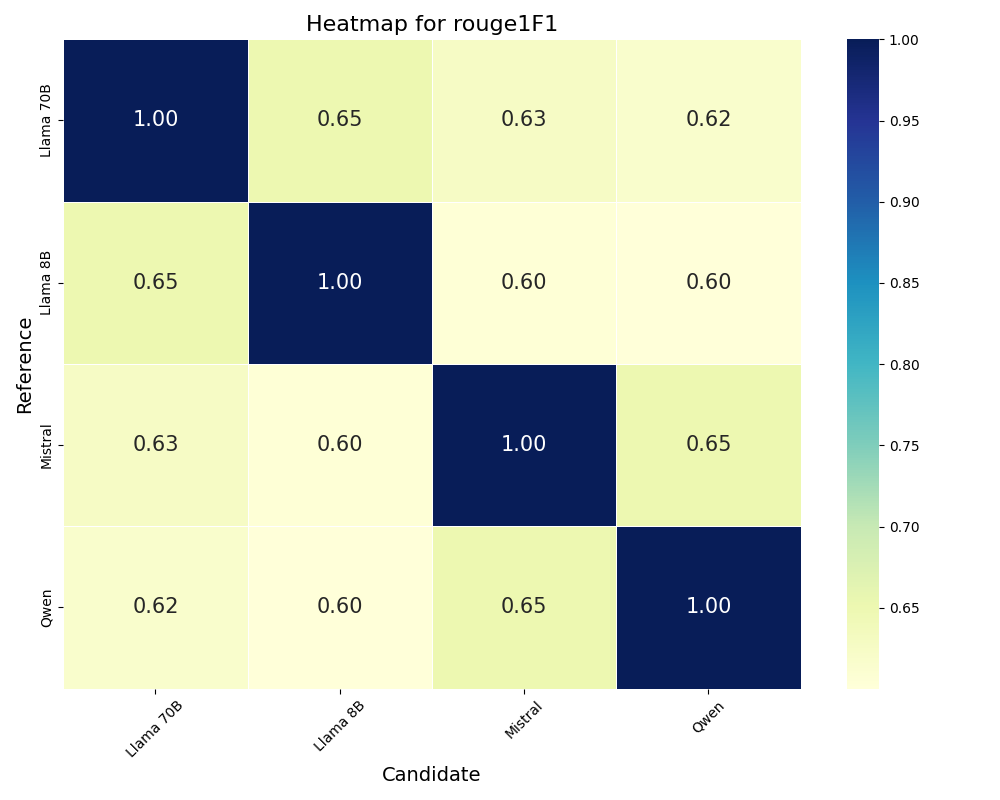}
        \caption{ORKGSynthesis}
        \label{fig:llm4syn_rouge1}
    \end{subfigure}
    \caption{ROUGE-1 metric results.}
    \label{fig:rouge1-metric-results}
\end{figure}

\begin{figure}[!htb]
    \centering
    \begin{subfigure}[b]{0.48\textwidth}
        \centering
        \includegraphics[width = \textwidth]{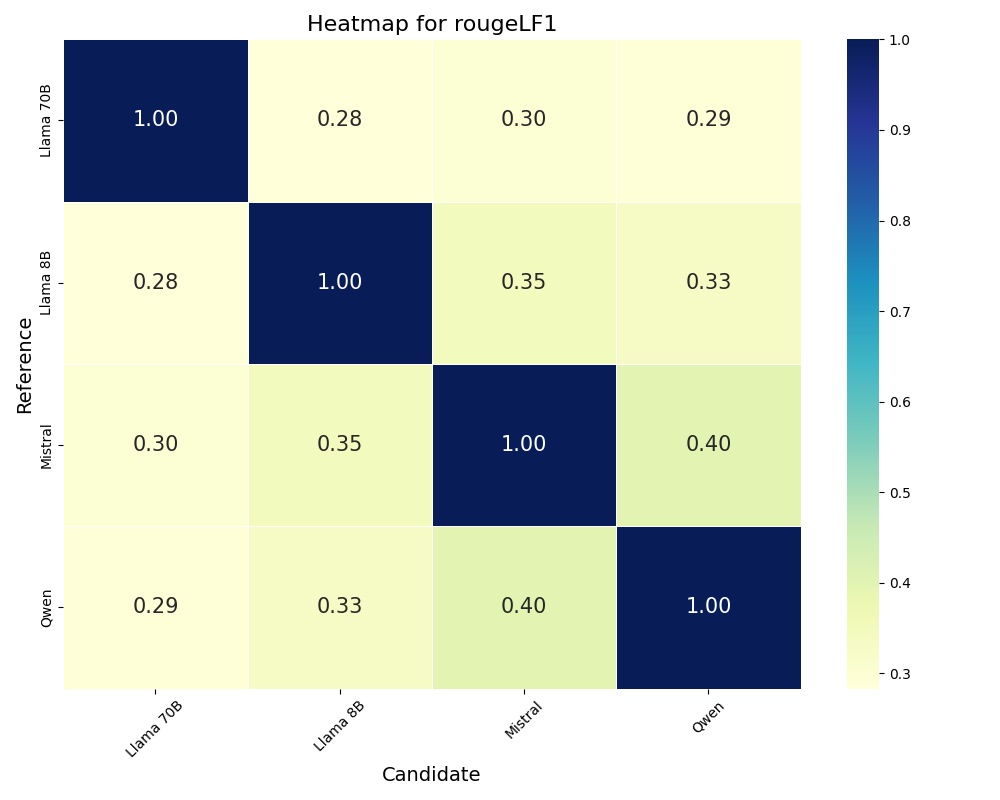}
        \caption{BioASQ}
        \label{fig:bioasq_rougel}
    \end{subfigure}
    \begin{subfigure}[b]{0.48\textwidth}
        \centering
        \includegraphics[width = \textwidth]{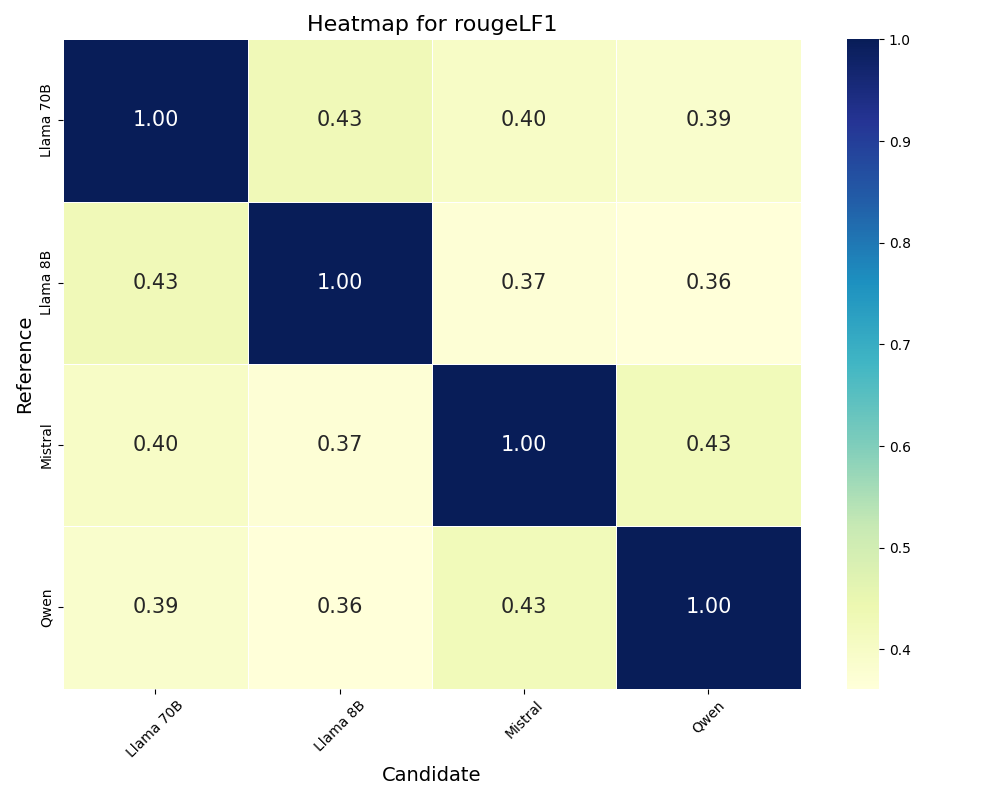}
        \caption{ORKGSynthesis}
        \label{fig:llm4syn_rougel}
    \end{subfigure}
    \caption{ROUGE-L metric results.}
    \label{fig:rougel-metric-results}
\end{figure}

Illustrated results for ROUGE-1 in \autoref{fig:rouge1-metric-results} and ROUGE-L in \autoref{fig:rougel-metric-results} exhibit distinct patterns across the two datasets. On the BioASQ dataset, \textit{Mistral-Large} and \textit{Qwen2.5-72B} demonstrate a notable correlation with a ROUGE-1 score of 0.58, suggesting shared thematic elements or vocabulary usage in their outputs. Additionally, a ROUGE-L score of 0.4 indicates a moderate similarity in sentence structure and coherence. In contrast, the \textit{LLaMA-3.1-70B} model performs relatively poorly, which may stem from inconsistencies in generating high-quality outputs for certain tasks in this domain. On the ORKGSynthesis dataset, the \textit{LLaMA-3.1-70B} model achieves the highest correlation, particularly when compared to the smaller LLaMA model. Mistral and \textit{Qwen2.5-72B} also display a strong correlation on this dataset, reinforcing their observed alignment. Exclusively focusing on unigrams in ROUGE-1 yields a higher correlation than BLUE and ROUGE-L, where overlaps are analyzed up to the four-gram level. This is because unigram-based evaluations inherently capture a broader overlap by disregarding strict positional constraints or dependencies on higher-order matches. 

\noindent\textbf{Metric for Evaluation of Translation with Explicit ORdering (METEOR). }METEOR~\citep{meteor} improves upon BLEU by prioritizing recall, which better aligns with human judgments. Unlike BLEU’s brevity penalty, METEOR explicitly integrates recall into its scoring. It also replaces BLEU’s reliance on higher-order n-grams with direct word alignment, enhancing semantic and structural accuracy. Additionally, METEOR avoids BLEU’s zero-score issue by using an alignment-based approach that captures partial matches and syntactic nuances more effectively.

\begin{figure}[!htb]
    \centering
    \begin{subfigure}[b]{0.48\textwidth}
        \centering
        \includegraphics[width = \textwidth]{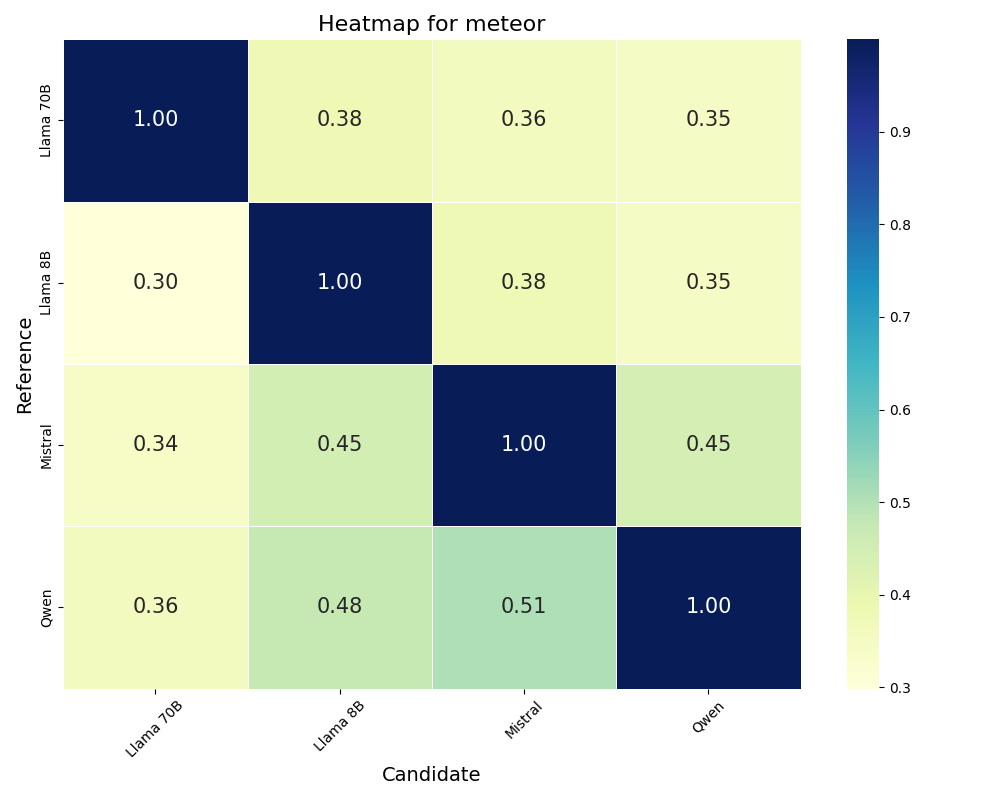}
        \caption{BioASQ}
        \label{fig:bioasq_meteor}
    \end{subfigure}
    \begin{subfigure}[b]{0.48\textwidth}
        \centering
        \includegraphics[width = \textwidth]{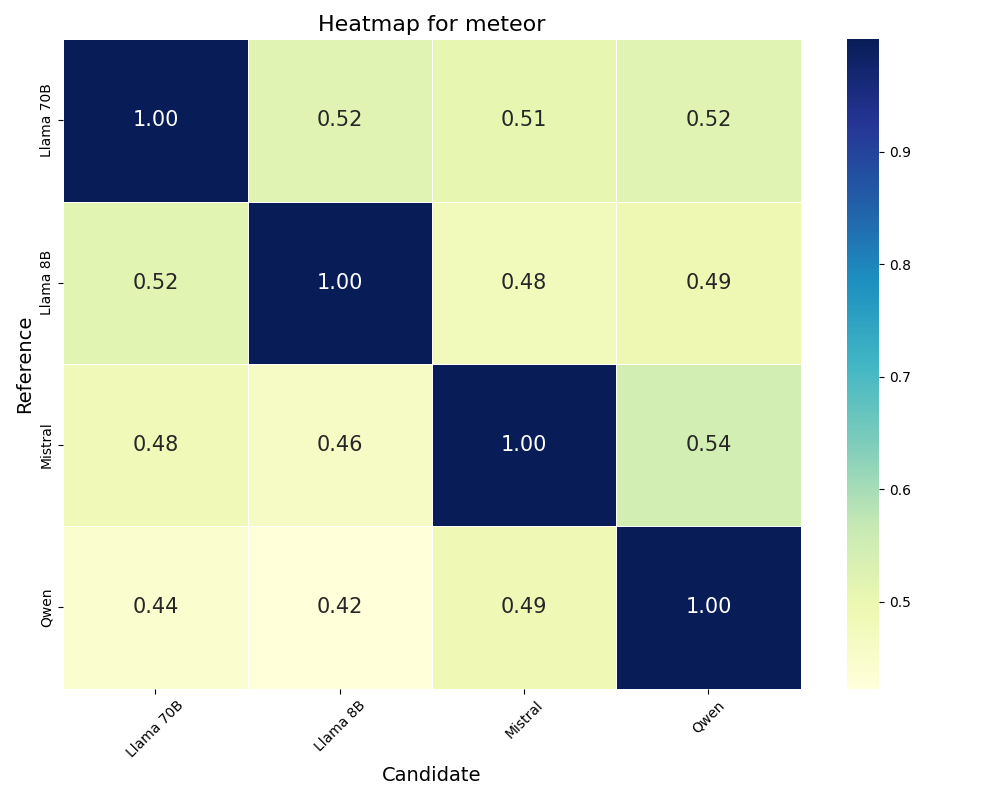}
        \caption{ORKGSynthesis}
        \label{fig:llm4syn_meteor}
    \end{subfigure}
    \caption{METEOR metric results.}
    \label{fig:meteor-metric-results}
\end{figure}
Results showed in \autoref{fig:meteor-metric-results} for METEOR metric, which produces scores ranging from 0 (no correlation) to 1 (perfect correlation). In our evaluation, METEOR highlighted differences between the models. On the BioASQ dataset, the \textit{LLaMA-3.1-70B} model performed poorly, consistent with earlier metrics. Conversely, the \textit{LLaMA-3.1-8B} model achieved comparable scores to \textit{Mistral-Large} and  \textit{Qwen2.5-72B}, suggesting it generates responses that align well with unigram matches. For the ORKGSynthesis dataset, the scores were more uniform across models, reflecting a general similarity in performance. However, an intriguing pattern emerged with \textit{Qwen2.5-72B}: when used as a reference, its scores varied substantially compared to when it was a candidate. This discrepancy may be attributable to METEOR's emphasis on recall. Higher scores when \textit{Qwen2.5-72B} is a candidate suggest it produces longer outputs, increasing the likelihood of matches with reference terms. This raises questions about the relationship between generation length and perceived quality in evaluation metrics, which warrants further investigation.

\noindent\textbf{US National Institute of Standards and Technology (NIST). }The NIST~\citep{nist} score builds on the BLUE metric but introduces a significant enhancement by focusing on the information content of n-grams. Unlike BLUE, which traditionally weights all matches equally, NIST prioritizes matches with greater informational value. This approach stems from the observation that rarer words or n-grams convey more specific and meaningful information than frequent ones. Consequently, matching infrequent n-grams contributes more to the NIST score, making it a more context-sensitive evaluation metric.

\begin{figure}[t]
    \centering
    \begin{subfigure}[b]{0.48\textwidth}
        \centering
        \includegraphics[width = \textwidth]{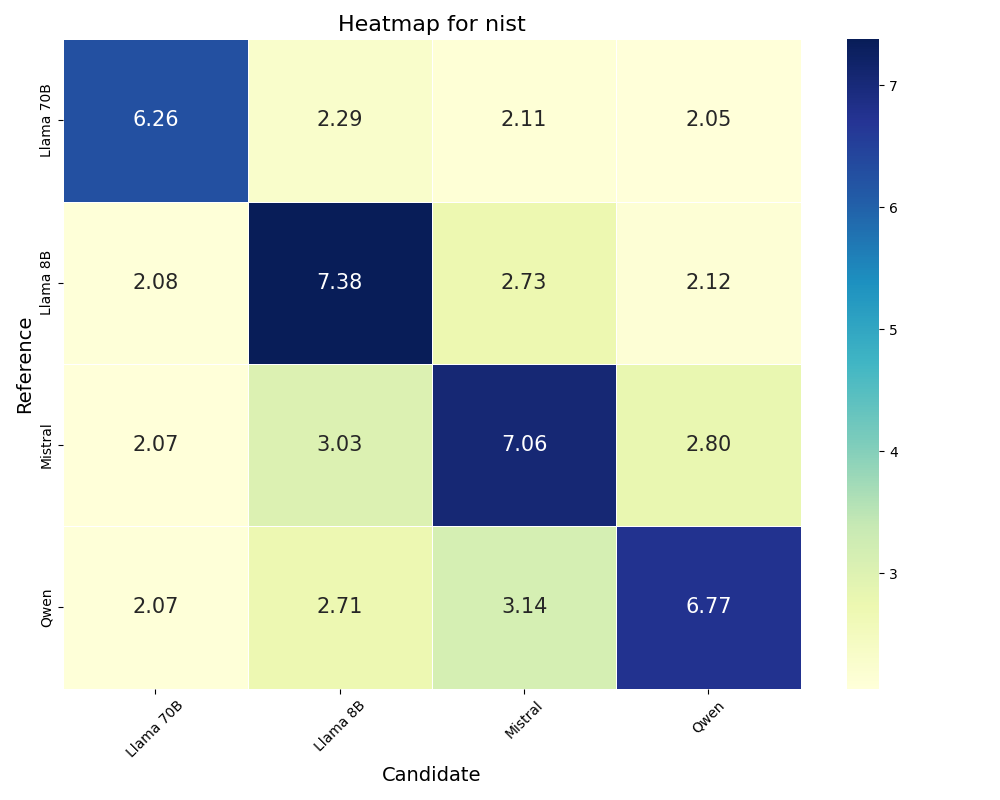}
        \caption{BioASQ}
        \label{fig:bioasq_nist}
    \end{subfigure}
    \begin{subfigure}[b]{0.48\textwidth}
        \centering
        \includegraphics[width = \textwidth]{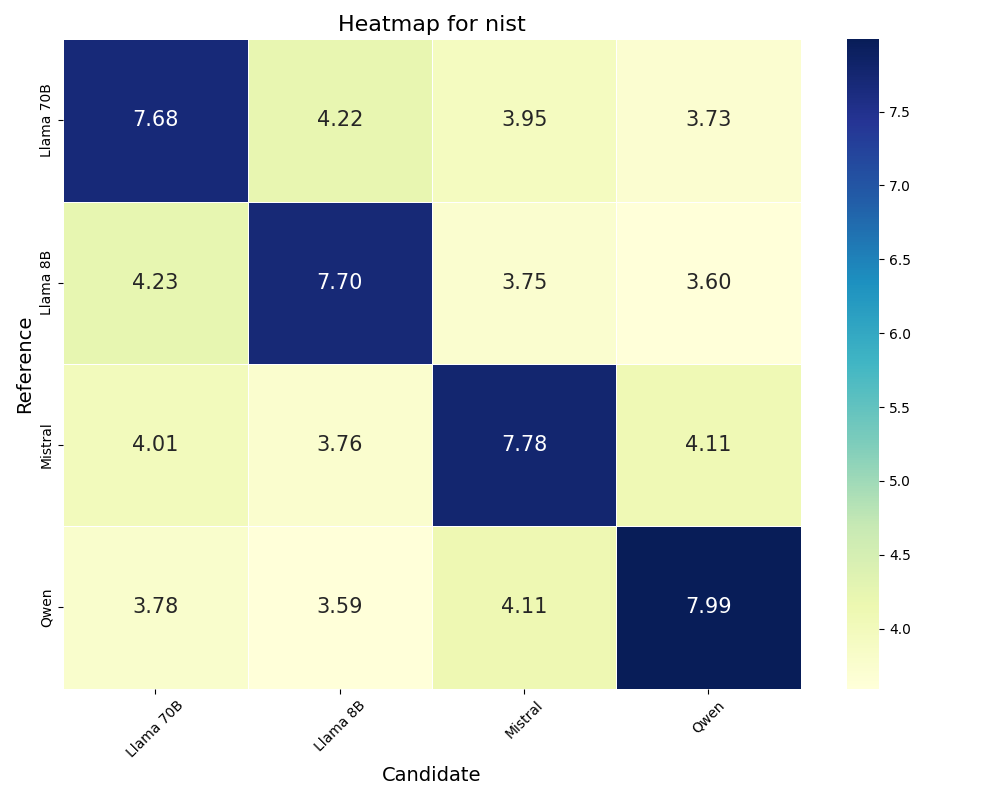}
        \caption{ORKGSynthesis}
        \label{fig:llm4syn_nist}
    \end{subfigure}
    \caption{NIST metric results}
    \label{fig:nist-metric-results}
\end{figure}
The NIST scores are unbounded and range from 0 (indicating poor quality) to higher positive values, with larger scores reflecting better-quality matches. Unlike BLUE, the NIST score's sensitivity to informational content allows it to provide more nuanced assessments. As shown in \autoref{fig:nist-metric-results} for the BioASQ dataset, the \textit{LLaMA-3.1-70B} model consistently underperforms, even when compared against its output. This poor performance can be attributed to the model's tendency to generate shorter responses and repeat words within a single response, leading to lower overall information gain. In contrast, other models demonstrate relatively strong correlations, suggesting more balanced and information-rich outputs. In the ORKGSynthesis dataset, correlations are notably higher across all models. The \textit{LLaMA-3.1-70B} model achieves strong alignment with its smaller counterpart, consistent with observations in other metrics. Similarly, \textit{Mistral-Large} and \textit{Qwen2.5-72B} continue to display strong correlations.

\noindent\textbf{BERTScore.} BERTScore~\citep{bertscore} is designed to evaluate text similarity by leveraging contextualized embeddings from pre-trained transformer models like BERT~\citep{devlin-etal-2019-bert}. Unlike traditional metrics that rely on exact token matches or n-gram overlap, BERTScore assesses semantic similarity at a token level, enabling it to recognize paraphrases, capture long-range dependencies, and account for nuanced semantic ordering. This approach allows it to surpass earlier metrics in evaluating complex and varied outputs, as it does not depend solely on surface-level text similarity. The strength of BERTScore lies in its ability to align tokens in a candidate sentence with those in a reference sentence using their embeddings, which encode rich contextual information. Studies by~\citet{bertscore} demonstrate that BERTScore highly correlates with human judgment, making it a valuable tool for evaluating machine-generated text.

\begin{figure}[t]
    \centering
    \begin{subfigure}[b]{0.48\textwidth}
        \centering
        \includegraphics[width = \textwidth]{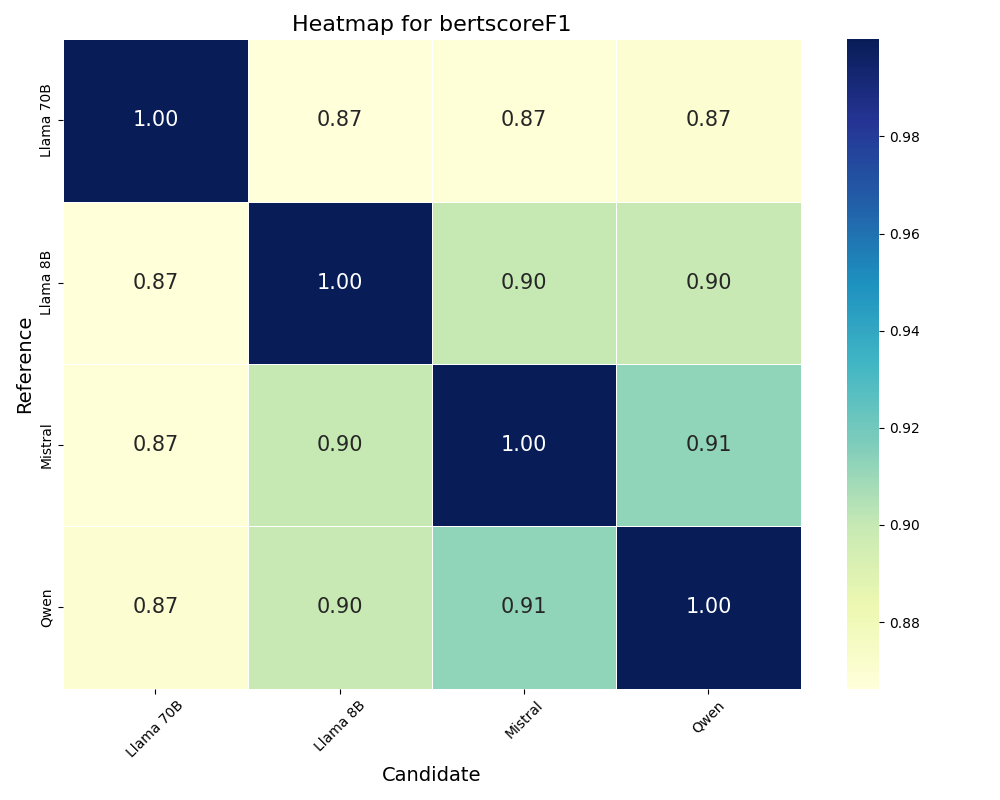}
        \caption{BioASQ}
        \label{fig:bioasq_bertscore}
    \end{subfigure}
    \begin{subfigure}[b]{0.48\textwidth}
        \centering
        \includegraphics[width = \textwidth]{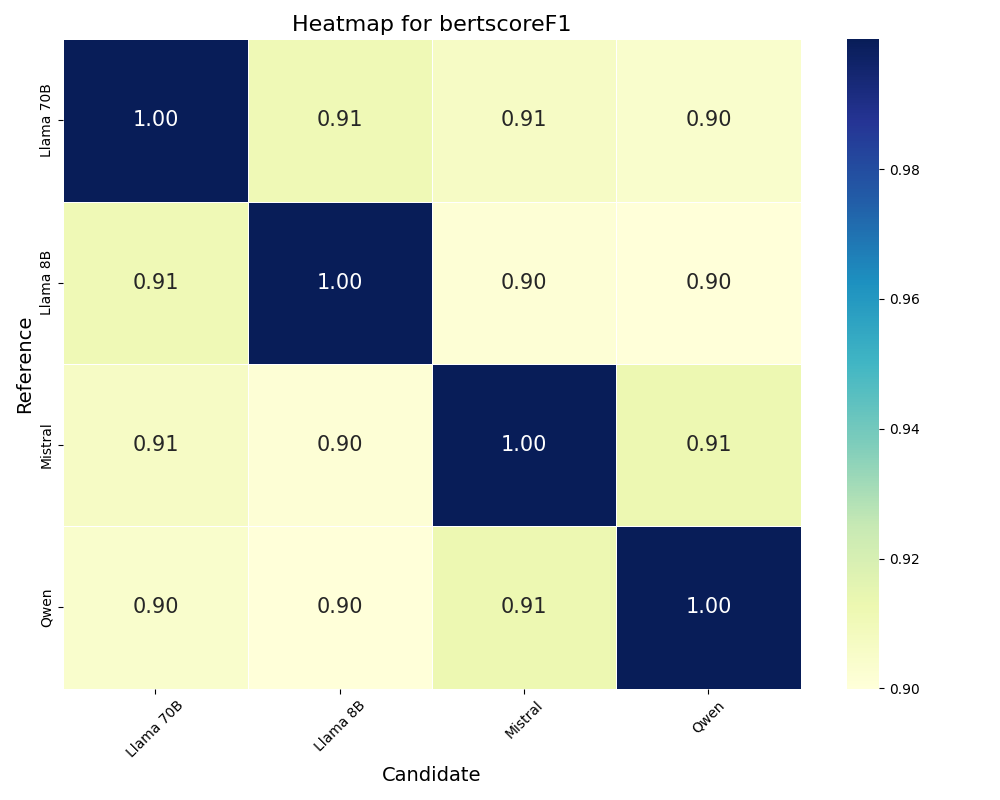}
        \caption{ORKGSynthesis}
        \label{fig:llm4syn_bertscore}
    \end{subfigure}
    \caption{BERTScore metric results.}
    \label{fig:bertscore-metric-results}
\end{figure}
BERTScore generates values between 0 and 1, where a score closer to 1 indicates stronger semantic alignment. In our analysis w.r.t \autoref{fig:bertscore-metric-results}, BERTScore highlights the limitations of the \textit{LLaMA-3.1-70B} model on the BioASQ dataset, as its tendency to produce incomplete or incoherent answers results in lower scores. The other models, including those evaluated on the ORKGSynthesis dataset, achieve scores near 0.9, which reflects a high degree of semantic similarity between their outputs.  Unlike metrics such as BLUE, ROUGE, or METEOR, which primarily evaluate surface-level similarity based on token overlap or n-gram matches, BERTScore incorporates the contextual meaning of tokens. This capability allows it to capture deeper semantic relationships, even when lexical or syntactic differences exist between candidate and reference sentences. Consequently, while earlier metrics may indicate lower correlation, particularly for models generating paraphrased or stylistically varied outputs, BERTScore reveals that the underlying semantic content remains closely aligned. This observation underscores the potential of embedding-based metrics to provide a more nuanced evaluation of language model outputs, particularly in tasks where paraphrasing and creative rewording are common. Furthermore, the high BERTScore results across most models suggest that their outputs are semantically coherent, even if traditional metrics fail to capture this aspect. This finding highlights the value of embedding-based metrics in complementing traditional approaches, providing a broad evaluation framework. Future work could explore fine-tuning the embeddings used in BERTScore to align even more closely with domain-specific human judgments, particularly in specialized tasks such as medical or scientific text generation.

\subsection{Edit Distance-Based Metrics}
To complement summary-based evaluation, we also incorporated edit distance-based metrics, which assess text similarity by measuring the number of modifications required to convert one string into another. This approach originates from Levenshtein distance~\cite{levenshtein}, a fundamental concept in text processing commonly applied in spell-checking and auto-correction. Edit distance evaluation quantifies text dissimilarity through insertions, deletions, and substitutions. For instance, converting the word \textit{mug} to \textit{hut} requires substituting two characters (\textit{m} to \textit{h} and \textit{g} to \textit{t}), which, if we assign a cost of 1 per insertion/deletion and 2 per substitution, yields an edit distance of 4. These costs can be adjusted depending on the task. 

\noindent\textbf{Word Error Rate (WER). }The WER~\cite{wer} is a similarity metric grounded in the concept of minimum edit distance, measuring the number of edits (substitutions, insertions, and deletions) required to transform a candidate text into reference text. Unlike character-level edit distance, WER operates at the word level, treating entire words as the basic transformation units. It is particularly suited for speech recognition and machine translation, where word-level alignment is essential. However, the metric has inherent limitations, particularly its sensitivity to word order. Sentences with semantically identical meanings but different word arrangements are heavily penalised, potentially leading to overly pessimistic assessments of similarity.

\begin{figure}[!htb]
    \centering
    \begin{subfigure}[b]{0.48\textwidth}
        \centering
        \includegraphics[width = \textwidth]{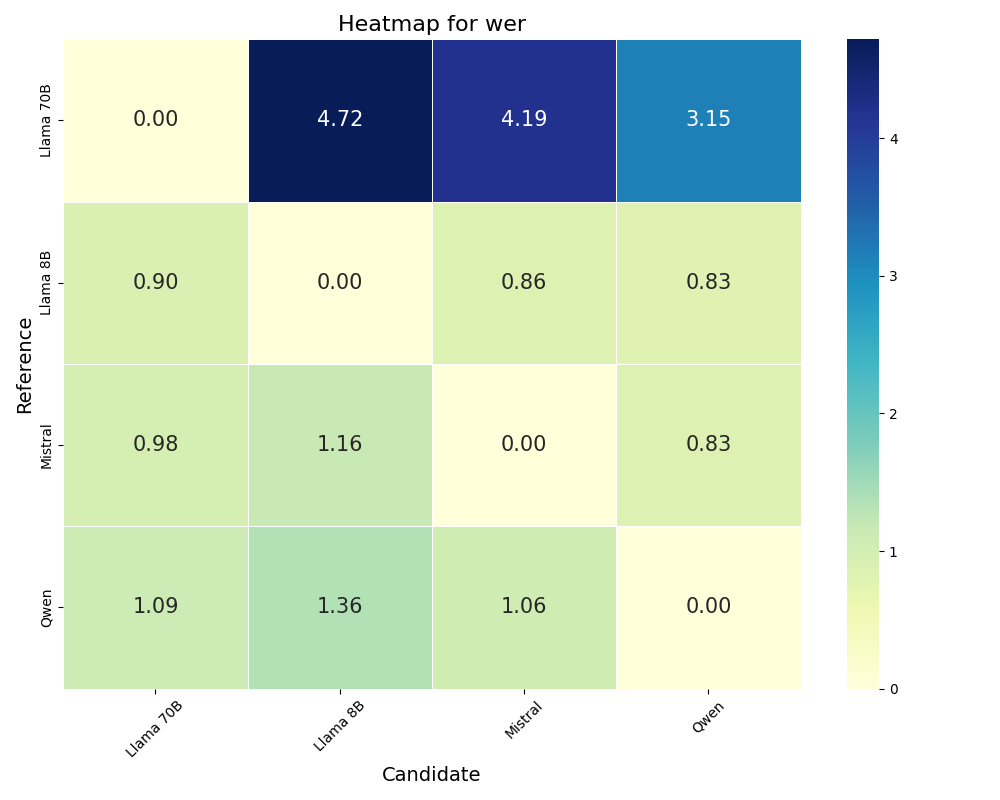}
        \caption{BioASQ}
        \label{fig:bioasq_wer}
    \end{subfigure}
    \begin{subfigure}[b]{0.48\textwidth}
        \centering
        \includegraphics[width = \textwidth]{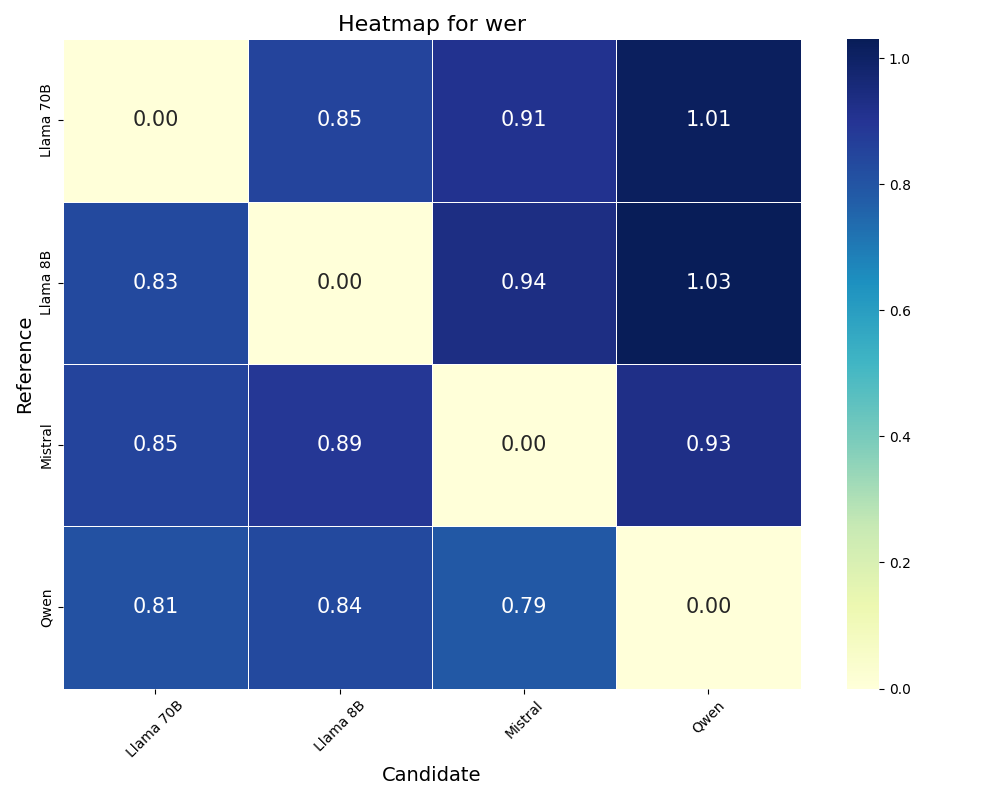}
        \caption{ORKGSynthesis}
        \label{fig:llm4syn_wer}
    \end{subfigure}
    \caption{WER metric results}
    \label{fig:wer-metric-results}
\end{figure}
The analysis of WER according to \autoref{fig:wer-metric-results} scores across different datasets and models reveals several notable patterns. For the BioASQ dataset, the \textit{LLaMA-3.1-70B} model exhibits markedly poor performance as a reference. This behavior can be attributed to the model's tendency to produce shorter outputs. Since WER normalizes the number of edits by reference length, shorter reference texts amplify the impact of any discrepancies, leading to inflated WER values. This observation aligns with earlier findings highlighting \textit{LLaMA-3.1-70B}'s challenges in generating comprehensive responses for the BioASQ dataset. In contrast, for the ORKGSynthesis dataset, \textit{Qwen2.5-72B} demonstrates superior performance when serving as a reference. However, as a candidate, \textit{Qwen2.5-72B} achieves lower scores. This discrepancy likely arises from \textit{Qwen2.5-72B}'s tendency to generate longer sequences. In such cases, the normalization by reference length in the WER formula leads to more significant deviations when \textit{Qwen2.5-72B}'s outputs are compared against shorter references. These findings underscore the metric's dependence on the relative length of candidate and reference texts, which can introduce biases when evaluating models with different generation strategies. While WER provides a straightforward measure of surface-level similarity, its inability to account for semantic equivalence or tolerate variations in word order limits its applicability in evaluating generative models. For instance, outputs with paraphrased structures or stylistic differences might receive high WER scores despite being semantically aligned with the reference.

\noindent\textbf{Word Mover's Distance (WMD). }The WMD \cite{wmd} quantifies the dissimilarity between two text documents by calculating the minimum cumulative "distance" that the embedded words in one document must travel to align with the words in another document. The metric is inspired by the Earth Movers Distance, a concept in optimal transport theory, which measures the minimum work required to transform one probability distribution into another. Unlike greedy matching approaches like BERTScore, WMD leverages an optimal matching strategy, ensuring a more precise alignment of semantically relevant terms. In its original formulation, WMD used Word2Vec embeddings to represent words as vectors in a continuous space. However, we employed SPECTER2~\cite{specter} and SciBERT~\cite{scibert} embeddings, designed for scientific texts and have shown superior performance in domain-specific applications.

\begin{figure}[!htb]
    \centering
    \begin{subfigure}[b]{0.48\textwidth}
        \centering
        \includegraphics[width = \textwidth]{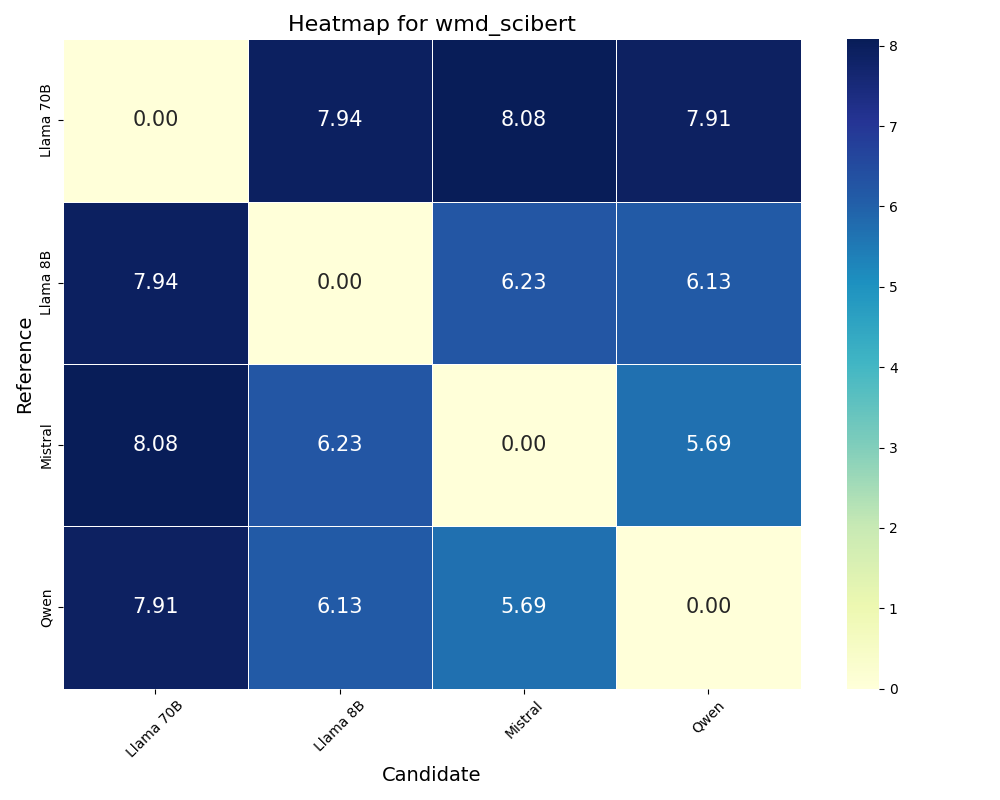}
        \caption{BioASQ}
        \label{fig:bioasq_wmdscibert}
    \end{subfigure}
    \begin{subfigure}[b]{0.48\textwidth}
        \centering
        \includegraphics[width = \textwidth]{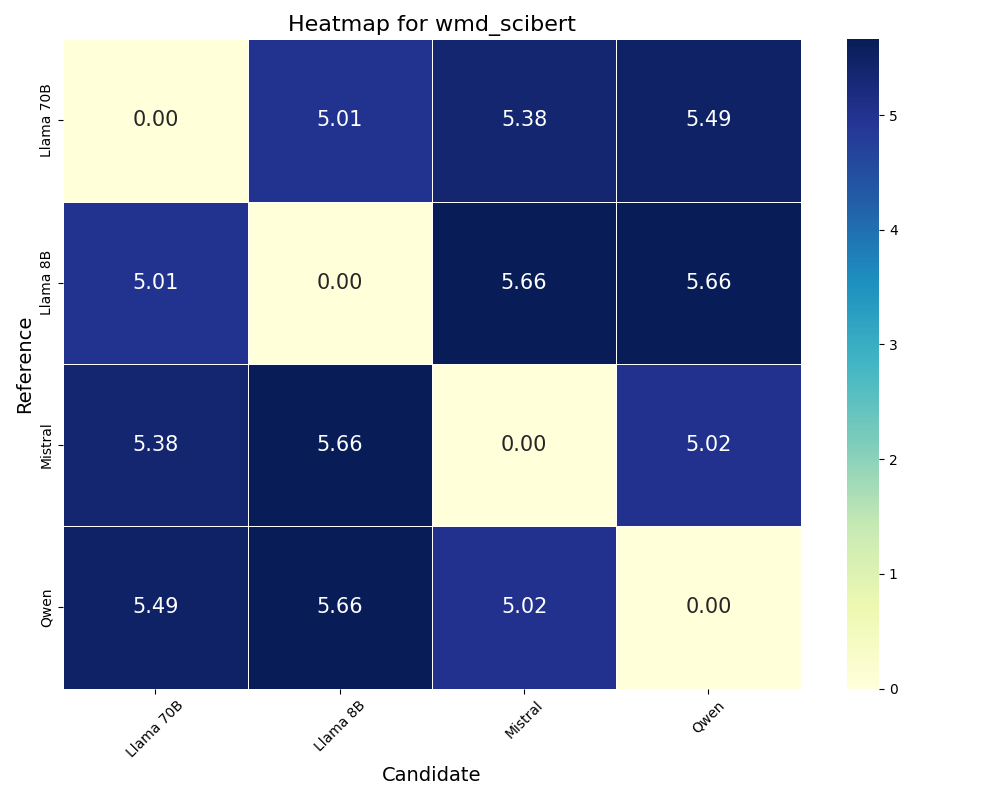}
        \caption{ORKGSynthesis}
        \label{fig:llm4syn_wmdscibert}
    \end{subfigure}
    \caption{WMD metric results using SciBERT as embeddings.}
    \label{fig:wmdscibert-metric-results}
\end{figure}

\begin{figure}[!htb]
    \centering
    \begin{subfigure}[b]{0.48\textwidth}
        \centering
        \includegraphics[width = \textwidth]{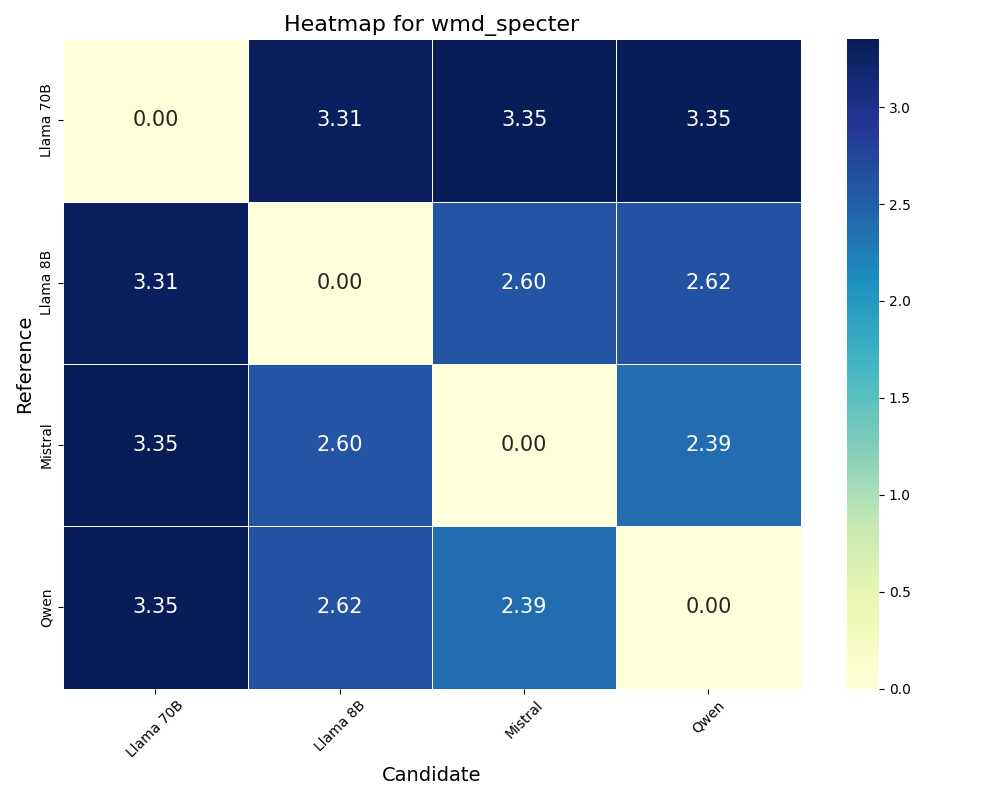}
        \caption{BioASQ}
        \label{fig:bioasq_wmdspecter}
    \end{subfigure}
    \begin{subfigure}[b]{0.48\textwidth}
        \centering
        \includegraphics[width = \textwidth]{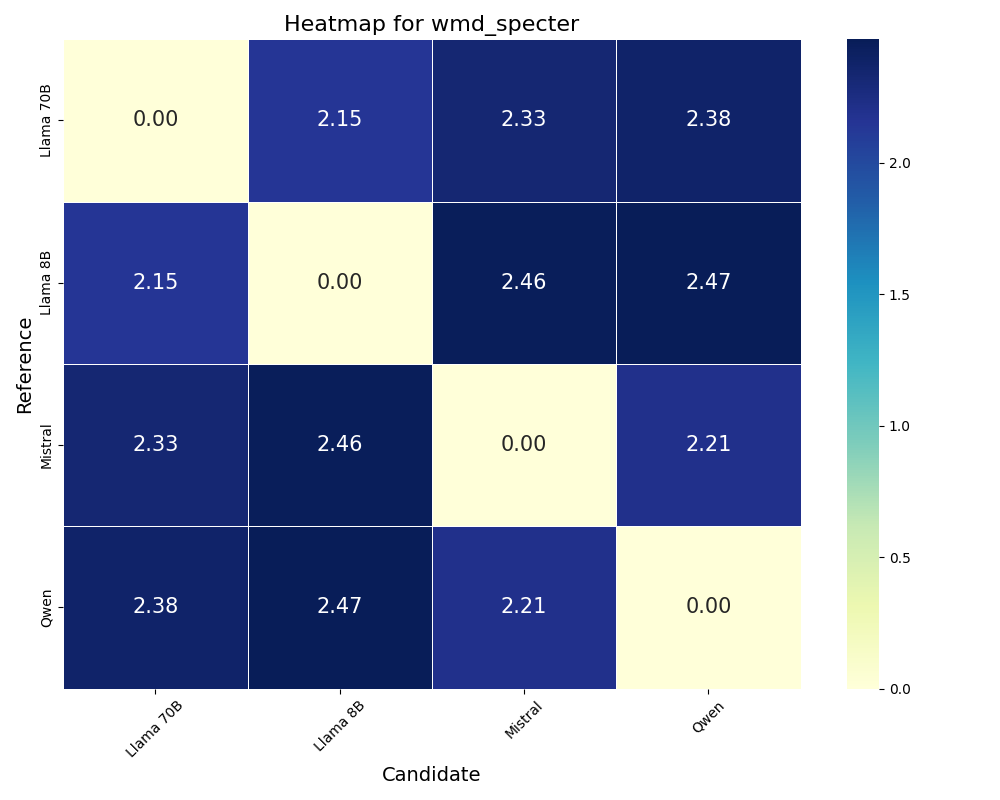}
        \caption{ORKGSynthesis}
        \label{fig:llm4syn_wmdspecter}
    \end{subfigure}
    \caption{WMD metric results using SPECTER2 as embeddings.}
    \label{fig:wmdspecter-metric-results}
\end{figure}

The results of WMD using SPECTER2 embeddings are represented in \autoref{fig:wmdspecter-metric-results} and WMD with SciBERT is represented in \autoref{fig:wmdscibert-metric-results}. A WMD score of 0 indicates perfect alignment between the candidate and reference texts, while higher scores denote greater dissimilarity. Analysis of the WMD scores across datasets highlights several trends. As expected, \textit{LLaMA-3.1-70B} exhibits poor performance on the BioASQ dataset, consistent with its tendency to generate incoherent responses. In contrast on the ORKGSynthesis dataset, the two LLaMA models show strong alignment with one another, as do \textit{Mistral-Large} and \textit{Qwen2.5-72B}. An intriguing finding is the significant difference in WMD scores when using SPECTER embeddings compared to SciBERT embeddings. Specifically, SPECTER embeddings yield substantially lower WMD scores, suggesting they provide better semantic representations for this task. 

\noindent\textbf{MoverScore.} The MoverScore~\cite{moverscore} is an advanced metric that extends the principles of WMD to evaluate the dissimilarity of text documents by comparing both words and n-grams. One of its primary advantages lies in its use of contextual embeddings, such as those generated by BERT, instead of static embeddings. This enables MoverScore to capture nuanced meanings, including word sense disambiguation and contextual relationships. Another key improvement is its allowance for many-to-one soft alignments, enabling more flexible matching between text elements. Furthermore, MoverScore incorporates inverse document frequency (IDF) weighting, emphasising rare and meaningful words, ensuring that these words contribute more significantly to the similarity score. The combination of BERT's contextual embeddings and IDF weighting has been shown by \citet{moverscore} to correlate highly with human judgment.

\begin{figure}[!htb]
    \centering
    \begin{subfigure}[b]{0.48\textwidth}
        \centering
        \includegraphics[width = \textwidth]{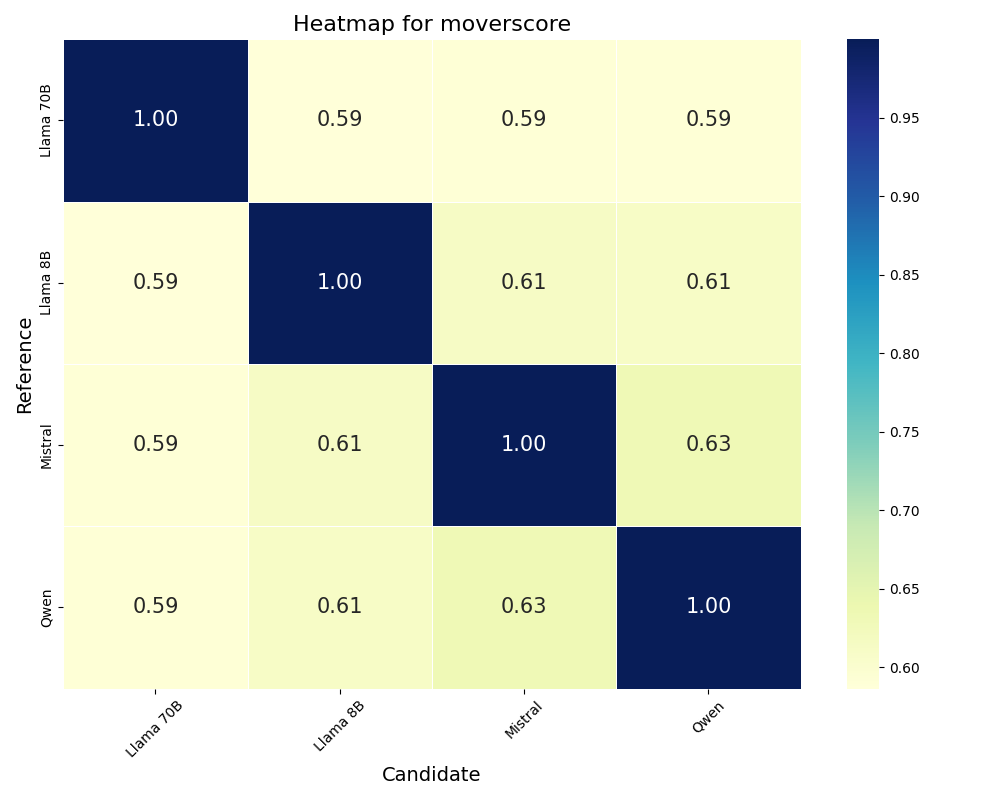}
        \caption{BioASQ}
        \label{fig:bioasq_moverscore}
    \end{subfigure}
    \begin{subfigure}[b]{0.48\textwidth}
        \centering
        \includegraphics[width = \textwidth]{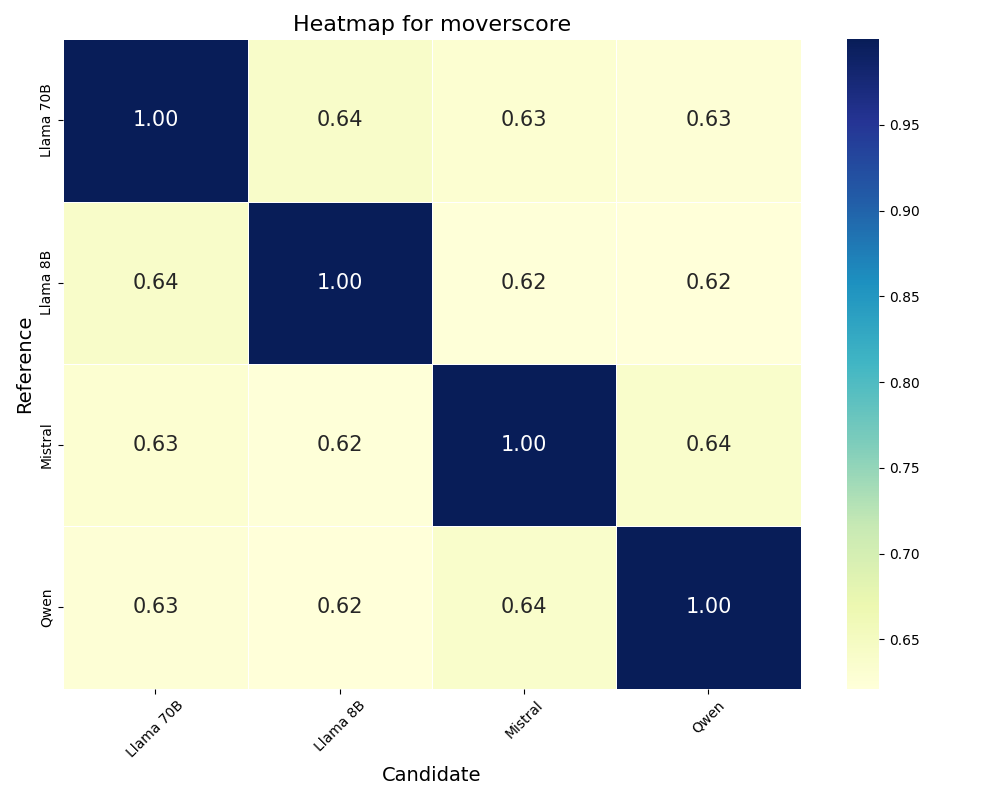}
        \caption{ORKGSynthesis}
        \label{fig:llm4syn_moverscore}
    \end{subfigure}
    \caption{MoverScore metric results}
    \label{fig:moverscore-metric-results}
\end{figure}
MoverScore produces a value between 0 and 1, where higher scores correspond to greater similarity between the candidate and reference texts. Across both datasets as shown in \autoref{fig:moverscore-metric-results}, MoverScore results hover around 0.62, reflecting moderate alignment. As observed with other metrics, the \textit{LLaMA-3.1-70B} model performs slightly worse on the BioASQ dataset, likely attributable to the model's struggles with generating comprehensive and coherent responses. In contrast, the ORKGSynthesis dataset reveals strong correlations between the outputs of the \textit{Mistral-Large} and \textit{Qwen2.5-72B} models, and between the LLaMA models. The relatively consistent scores across models suggest that, according to MoverScore, the quality of responses is comparable. This uniformity can be interpreted as evidence that the models generate outputs with similar semantic content and structure, despite potential stylistic differences.

\section{Evaluation Scoring Rubrics and the Five-point Likert Scale}
\label{sec:det-likert}
This appendix presents the quality rubrics and their corresponding 5-point Likert scale descriptions used in our evaluation. \autoref{tab:linguistic-and-stylistic-rubrics-scale-description} outlines the Linguistic and Stylistic quality rubrics, assessing aspects such as cohesion, conciseness, and readability. \autoref{tab:logical-and-structural-integrity-rubrics-scale-description} details the Logical and Structural Integrity rubrics, evaluating factors like argument coherence, integration, and relevancy. Lastly, \autoref{tab:content-and-informativeness-rubrics-scale-description} describes the Content Accuracy and Informativeness rubrics, which measure correctness, completeness, and informativeness.



\begin{table*}[]
    \centering
    \small
    \begin{tabular}{|p{1.5cm}|p{14cm}|}
        \hline
        \textbf{Rubric} & \textbf{5-point Likert scale description} \\
        \hline
        \hline
         \multirow{5}{*}{\textbf{Cohesion}} & \textit{Rating 1. Very bad}: The sentences within the synthesis are disconnected, resulting in a disjointed and fragmented narrative. \\
         & \textit{Rating 2. Bad}: There are attempts at connecting sentences, but the synthesis often feels disjointed.\\
         & \textit{Rating 3. Moderate}: The sentences are connected in a way that the synthesis is mostly cohesive, with some areas of improvement.\\
         & \textit{Rating 4. Good}: The synthesis is cohesive, with sentences well-connected to form a unified narrative.\\
         & \textit{Rating 5. Very good}: The synthesis is highly cohesive, with all sentences and paragraphs logically connected, facilitating a clear and coherent narrative flow.\\
         \hline
         \multirow{5}{*}{\textbf{Conciseness}} & \textit{Rating 1. Very Bad}: The synthesis is verbose and cluttered with redundant or irrelevant information, significantly detracting from its clarity and focus. \\
         & \textit{Rating 2. Bad}: The synthesis includes some redundant or irrelevant statements, detracting from its clarity.\\
         & \textit{Rating 3. Moderate}: The synthesis is relatively clear and to the point, but could be more concise by eliminating a few redundant elements.\\
         & \textit{Rating 4. Good}: The synthesis is concise and to the point, with virtually no redundant statements or unnecessary information.\\
         & \textit{Rating 5. Very Good}: The synthesis is precisely concise, delivering information clearly and directly without any superfluous details or redundancy, enhancing its clarity and impact.\\
         \hline
         \multirow{5}{*}{\textbf{Readability}}& \textit{Rating 1. Very bad}: The synthesis is poorly written, with pervasive issues in style, structure, and language use, making it difficult to understand. \\ 
        & \textit{Rating 2. Bad}: The text has noticeable issues with style, structure, or language use, affecting clarity.\\ 
        & \textit{Rating 3. Moderate}: The synthesis follows appropriate conventions and uses language correctly, with minor issues in style or structure.\\ 
        & \textit{Rating 4. Good}: The text is well-structured and easy to read, with language that is appropriately used and only minor stylistic improvements needed.\\ 
        & \textit{Rating 5. Very good}: The synthesis is exceptionally well-written, following stylistic and structural conventions with precise language use, making it accessible and easy to read.\\ 
        \hline
    \end{tabular}
    \caption{\textbf{Linguistic and Stylistic} Quality Rubrics and their 5-point Likert Scale Description}
    \label{tab:linguistic-and-stylistic-rubrics-scale-description}
\end{table*}

\begin{table*}[]
    \centering
    \small
    \begin{tabular}{|p{1.5cm}|p{14cm}|}
        \hline
        \textbf{Rubric} & \textbf{5-point Likert scale description} \\
        \hline
        \hline
         \multirow{5}{*}{\textbf{Coherence}}& \textit{Rating 1. Very bad}: The synthesis lacks logical connection between ideas, leading to a narrative that is confusing and difficult to follow. \\
        &\textit{Rating 2. Bad}: The ideas are not always logically connected, leading to a somewhat confusing narrative.\\
        &\textit{Rating 3. Moderate}: The ideas are logically connected for the most part, but the narrative could be strengthened for better clarity.\\
        &\textit{Rating 4. Good}: The ideas are logically and soundly connected, offering a clear and understandable narrative.\\
        &\textit{Rating 5. Very good}: The ideas within the synthesis are connected in a logical and sound manner, forming a coherent and compelling narrative that is easy to follow.\\
        \hline
         \multirow{5}{*}{\textbf{Integration}}& \textit{Rating 1. Very Bad}: The synthesis fails to integrate the sources in any meaningful way. It lacks appropriate markers, connectors, or transitions between ideas and fails to combine the information into a single, cohesive paragraph.\\
        &\textit{Rating 2. Bad}: The sources are somewhat integrated but inconsistently. The use of markers and connectors is sporadic or inappropriately applied, and the information is presented in multiple paragraphs without a clear unifying structure.\\
        &\textit{Rating 3. Moderate}: The sources are integrated into a coherent manner within one or multiple paragraphs. The transitions or connections could be smoother, and the text would benefit from better paragraph structure to enhance clarity and unity.\\
        &\textit{Rating 4. Good}: The sources are well-integrated, using appropriate markers and connectors to create a seamless narrative. The information is effectively organized into a single paragraph, showing a clear, unified approach.\\
        &\textit{Rating 5. Very Good}: The synthesis seamlessly integrates information from the various sources, using appropriate markers and connectors to create a smooth and unified narrative. All information is skillfully condensed into a single, well-structured paragraph, exemplifying excellent integration.\\
        \hline
         \multirow{5}{*}{\textbf{Relevancy}}& \textit{Rating 1. Very bad}: The information provided does not relate to the research question, showing a lack of understanding or connection to the topic. \\
            &\textit{Rating 2. Bad}: The information occasionally relates to the research question but lacks direct and consistent relevance.\\
            &\textit{Rating 3. Moderate}: The information is generally related to the research question, with occasional lapses in direct relevance.\\
            &\textit{Rating 4. Good}: The information is consistently relevant to the research question, with only minor exceptions.\\
            &\textit{Rating 5. Very good}: The synthesis is directly and consistently relevant to the research question, demonstrating a deep understanding of the topic and its nuances.\\
        \hline
    \end{tabular}
    \caption{\textbf{Logical and Structural Integrity} Quality Rubrics and their 5-point Likert Scale Description}
    \label{tab:logical-and-structural-integrity-rubrics-scale-description}
\end{table*}

\begin{table*}[]
    \centering
    \small
    \begin{tabular}{|p{2cm}|p{13.5cm}|}
        \hline
        \textbf{Rubric} & \textbf{5-point Likert scale description} \\
        \hline
        \hline
         \multirow{5}{*}{\textbf{Correctness}}&  \textit{Rating 1. Very bad}: The synthesis consistently misrepresents or inaccurately portrays the content of the provided abstracts, showing a significant deviation from the original sources.\\
        &\textit{Rating 2. Bad}: The synthesis contains several inaccuracies or misinterpretations of the source abstracts.\\
        &\textit{Rating 3. Moderate}: The synthesis accurately represents most of the content from the provided abstracts but may contain minor errors.\\
        &\textit{Rating 4. Good}: The synthesis provides an accurate representation of the content from the provided abstracts with minor exceptions.\\
        &\textit{Rating 5. Very good}: The information in the synthesis is an accurate and faithful representation of the content from the provided abstracts, without any factual errors or misinterpretations.\\
        \hline
         \multirow{5}{*}{\textbf{Completeness}}&\textit{ Rating 1. Very bad}: The synthesis omits most of the relevant information, failing to capture the essential points or details from the provided abstracts. \\
         &\textit{Rating 2. Bad}: Significant portions of relevant information from the provided abstracts are missing. \\ 
         &\textit{Rating 3. Moderate}: The synthesis captures a fair amount of the relevant information, though it may overlook some details. \\
         &\textit{Rating 4. Good}: The synthesis includes almost all relevant information, missing only minor details.\\
         &\textit{Rating 5. Very good}: The synthesis comprehensively encapsulates all relevant information from the provided abstracts, leaving no pertinent details or points unaddressed.\\
         \hline
         \multirow{5}{*}{\textbf{Informativeness}}& \textit{Rating 1. Very bad}: The synthesis offers no valuable insights or useful information in response to the research question, lacking depth and utility. \\
            &\textit{Rating 2. Bad}: The answer provides limited new insights or useful information in response to the research question.\\
            &\textit{Rating 3. Moderate}: The answer is somewhat informative, offering insights or useful information but not in a comprehensive or detailed manner.\\
            &\textit{Rating 4. Good}: The answer is informative and insightful, providing comprehensive information in response to the research question.\\
            &\textit{Rating 5. Very good}: The synthesis is highly informative, providing valuable insights and detailed information that thoroughly addresses the research question. \\
    \hline
    \end{tabular}
    \caption{\textbf{Content Accuracy and Informativeness} Quality Rubrics and their 5-point Likert Scale Description}
    \label{tab:content-and-informativeness-rubrics-scale-description}
\end{table*}

\section{Experimental Setup}
\label{sec:exp-setup}

\noindent\textbf{Vanilla $LLM_{eval}$ models. }The Vanilla $LLM_{eval}$ models employ various LLMs, including \textit{Mistral-Large-Instruct}, \textit{LLaMA-3.1-70B-Instruct}, \textit{Qwen2.5-72B-Instruct}, and \textit{LLaMA-3.1-8B-Instruct}, as $LLM_{gen}$ to generate response to the questions $Q$ based on provided relevant papers. The evaluations for these models are conducted using $LLM_{eval}$, where each LLM acts as a generator and evaluator in a pairwise format. For example, \textit{Mistral-Large-Instruct} serves as the evaluator for all four $LLM_{gen}$ models, and the same process is applied to the other LLMs. The evaluation results are rated on a 5-point Likert scale, with each rating accompanied by a rationale that explains the model's reasoning. This configuration serves as a baseline for comparing the performance of different $LLM_{eval}$.

\noindent\textbf{SFT (benign). }The \textit{SFT (benign)} experiment involves finetuning the \textit{LLaMA-3.1-8B-Instruct} model using structured response $A$'s from all four models from $x=LLM_{gen}$ as inputs and quality assessments from $y=LLM_{eval}$ as outputs. The aim is to fine-tune $LLM_{eval}$ and this process is conducted using benign datasets, which contain no adversarial examples, to ensure that the model is trained in a controlled, non-hostile environment. The goal is to refine the model's performance in a straightforward, non-challenging scenario.

\noindent\textbf{SFT (benign) + RL (adversarial). }In this experiment, the finetuned \textit{LLaMA-3.1-8B-Instruct} model is further finetuned using the RLAIF technique, focusing solely on adversarial datasets. As seen in the results from \autoref{fig:results-plot}, while \textit{SFT (benign)} learns how to be the evaluator, but fails in recognizing the bad examples in adversarial sets. The \textit{SFT (benign) + RL (adversarial)} helps $LLM_{eval}$ to imitate the bad examples as well.

\noindent\textbf{SFT (benign) + RL (benign + adversarial). }Considering, the \textit{SFT (benign) + RL (adversarial)} experiments, while the model achieves high performance in identifying adversarial samples, it struggles to assign higher Likert scale ratings to benign samples. This is because the model's threshold for defining a "good" sample has been lowered to a rating of 3. The reason for such a drop in the maximum Likert scale in testing is due to poor penalization in RL that considered the desirable threshold of $\leq 3$ during the sampling of the data preparation for RLAIF for the subtle adversarial set. Essentially, training the model exclusively on adversarial behavior leads to confusion when it encounters benign examples. To solve this, the \textit{SFT (benign) + RL (benign + adversarial)} is ideal for imitating both benign and adversarial behaviors, which ideally leads to success.

\section{Training Details}
\label{sec:train-det}

\begin{figure}[t]
\centering
\begin{tcolorbox}[colframe=blue!40, colback=blue!5, coltitle=black, title=Conciseness Evaluation]
\{"Readability": \{ "rating": 1, "rationale": "The synthesis is poorly written, with pervasive issues in style, structure, and language use, making it difficult to understand due to the inclusion of an unrelated joke about \#jets fans and \#marksanchez at the end."\}\}
\end{tcolorbox}
\caption{The $y$ format for fine-tuning models.}
\label{example-output-format}
\end{figure}

\textbf{Supervised Fine-Tuning.} The fine-tuning threats each $LLM_{gen}$ as an input $x$ and outputs of all four $LLM_{eval}$ as a $y$. The fine-tuning $SFT$ dataset stats are represented in \autoref{tab:method-dataset-stats}. It includes research questions, synthesized responses, source papers, and evaluation rubrics scale to ensure the model learns both synthesis assessment and evaluation reasoning. The \autoref{example-output-format} shows the $y=LLM_{eval}$ format during the fine-tuning process for the "Readability" rubric (the example is chosen from the Adversarial set). The tokenized sequences were processed with a maximum sequence length specific to each dataset (ORKG-Synthesis: $4,645$ tokens, BioASQ: $8,874$ tokens). We finetuned the model per dataset for 5 epochs using the Paged AdamW $8$-bit optimizer, ensuring memory efficiency. Key hyperparameters included a batch size of $1$ per GPU (we used two GPUs), gradient accumulation steps equal to batch size, a learning rate of $2e-4$ with a warmup ratio of $0.03$, weight decay of $0.001$, and a max gradient norm of $0.3$.  The fine-tuned model checkpoints were saved for further analysis and RL-based fine-tuning.

\noindent\textbf{Reinforcement Learning.} Several hyperparameters and configurations for training an LLM with RL are used. Key hyperparameters include a learning rate of $2e-4$, a batch size of $1$ (per GPU), and a total of $2$ training epochs. The model's training is configured with a maximum prompt length of $4500$ and a maximum completion length of $150$. The CPO uses a per-device batch size of $1$, gradient accumulation steps of $1$, and mixed precision (fp16). Additionally, the model undergoes fine-tuning with a learning rate of $2e-4$. 

\noindent\textbf{Hardware and Resource Allocation.} For the system setup, two H100 GPUs were utilized for RL fine-tunings, each with 80 GB of GPU memory, while for SFT models, only one H100 GPU with the same memory capacity was used. The CPU configuration for SFT involved 60 GB of memory with 8 cores of CPU, while RL took 60 GB of memory and 16 CPU cores.

\section{Detailed Qualitative Experimental Results}
\label{sec:qual}
This section represents the detailed qualitative experimental results for seven models. In the tables, $LLM_{eval}$ models are defined as follows in the table columns: 
\begin{itemize}
    \item \textbf{M1}: Qwen2.5-72B
    \item \textbf{M2}: LLaMA-3.1-70B
    \item \textbf{M3}: Mistral-Large
    \item \textbf{M4}: LLaMA-3.1-8B
    \item \textbf{M5}: SFT (\texttt{benign})
    \item \textbf{M6}: SFT (\texttt{benign}) + RL (\texttt{adversarial})
    \item  \textbf{M7}: SFT (\texttt{benign}) + RL (\texttt{benign} + \texttt{adversarial})
\end{itemize}

The results for the BioASQ dataset are presented in \autoref{tab:bioasq-LLaMA-3.1-8B-detailed-results} for LLaMA-3.1-8B $LLM_{gen}$, \autoref{tab:bioasq-LLaMA-3.1-70B-detailed-results} for LLaMA-3.1-70B $LLM_{gen}$, \autoref{tab:bioasq-Qwen2.5-72B-detailed-results} for Qwen2.5-72B $LLM_{gen}$, and \autoref{tab:bioasq-Mistral-Large-detailed-results} for Mistral-Large $LLM_{gen}$. While, for the ORKGSynthesis dataset, results are presented in \autoref{tab:orkgsynthesis-LLaMA-3.1-8B-detailed-results} for LLaMA-3.1-8B $LLM_{gen}$, \autoref{tab:orkgsynthesis-LLaMA-3.1-70B-detailed-results} for LLaMA-3.1-70B $LLM_{gen}$, \autoref{tab:orkgsynthesis-Qwen2.5-72B-detailed-results} for Qwen2.5-72B $LLM_{gen}$, and \autoref{tab:orkgsynthesis-Mistral-Large-detailed-results} for Mistral-Large $LLM_{gen}$.

\begin{table*}[t]
        \centering
        \small
        \begin{tabular}{|p{2cm}|l|l|l|l|l|l|l|l|}
            \hline
             \textbf{Set}&\textbf{Rubrics} &  \textbf{M1} &  \textbf{M2} & \textbf{M3}& \textbf{M4}& \textbf{M5}  & \textbf{M6} & \textbf{M7}\\
             \hline 
             \hline 
                \multirow{9}{*}{\textbf{benign}}& 1. Coherence & 4.95 & 4.91 & 4.73 & 5.00 & 4.95 & 3.00 & 4.91\\ 
        & 2. Cohesion  & 4.95 & 4.91 & 4.68 & 4.95 & 4.95 & 3.00 & 4.68\\ 
        & 3. Completeness  & 4.41 & 4.33 & 4.27 & 4.91 & 4.23 & 3.00 & 4.32\\ 
        & 4. Conciseness  & 4.95 & 4.91 & 4.68 & 4.45 & 4.77 & 3.00 & 3.82\\ 
        & 5. Correctness  & 4.95 & 4.92 & 4.64 & 4.73 & 5.00 & 3.00 & 5.00\\ 
        & 6. Informativeness  & 4.82 & 4.82 & 4.86 & 5.00 & 4.68 & 3.00 & 5.00\\ 
        & 7. Integration  & 4.91 & 4.91 & 4.55 & 4.77 & 4.95 & 3.00 & 4.59\\ 
        & 8. Readability  & 4.95 & 4.91 & 4.73 & 4.64 & 4.95 & 3.00 & 4.55\\ 
        & 9. Relevancy  & 4.95 & 4.83 & 4.86 & 5.00 & 5.00 & 3.00 & 5.00\\ 

        \hline
        \multirow{9}{*}{\textbf{extreme}}& 1. Coherence & 2.50 & 3.00 & 3.00 & 4.64 & 3.55 & 1.00 & 1.00\\ 
        & 2. Cohesion  & 4.05 & 4.55 & 3.82 & 4.77 & 4.93 & 3.00 & 4.41\\ 
        & 3. Completeness  & 2.68 & 3.31 & 3.27 & 4.64 & 3.82 & 1.05 & 1.00\\ 
        & 4. Conciseness  & 1.41 & 1.00 & 1.50 & 2.27 & 4.77 & 1.00 & 1.00\\ 
        & 5. Correctness  & 3.91 & 3.90 & 3.73 & 4.68 & 3.59 & 1.00 & 1.00\\ 
        & 6. Informativeness  & 3.59 & 4.20 & 3.68 & 4.68 & 3.41 & 1.09 & 1.00\\ 
        & 7. Integration  & 3.73 & 4.08 & 3.18 & 4.77 & 4.82 & 3.00 & 3.14\\ 
        & 8. Readability  & 2.05 & 2.15 & 2.55 & 4.14 & 1.89 & 1.00 & 1.00\\ 
        & 9. Relevancy  & 2.64 & 3.40 & 3.55 & 4.86 & 2.77 & 1.00 & 1.00\\ 

        \hline
        \multirow{9}{*}{\textbf{subtle}}& 1. Coherence & 3.82 & 4.10 & 4.14 & 4.82 & 4.86 & 2.91 & 2.73\\ 
        & 2. Cohesion  & 4.05 & 4.25 & 4.23 & 4.68 & 4.82 & 3.00 & 4.64\\ 
        & 3. Completeness  & 4.14 & 4.09 & 3.82 & 4.82 & 4.36 & 3.00 & 3.91\\ 
        & 4. Conciseness  & 3.09 & 2.60 & 2.41 & 3.68 & 4.55 & 2.41 & 2.27\\ 
        & 5. Correctness  & 4.23 & 4.50 & 4.14 & 4.64 & 4.86 & 3.00 & 2.82\\ 
        & 6. Informativeness  & 4.18 & 4.40 & 4.23 & 4.95 & 4.59 & 3.00 & 2.77\\ 
        & 7. Integration  & 4.09 & 4.27 & 3.86 & 4.64 & 4.91 & 3.00 & 4.64\\ 
        & 8. Readability  & 3.14 & 2.92 & 3.45 & 4.18 & 4.18 & 2.18 & 1.82\\ 
        & 9. Relevancy  & 4.32 & 4.20 & 4.36 & 5.00 & 4.77 & 2.77 & 2.73\\ 

        \hline

    \end{tabular}
        \caption{BioASQ dataset detailed evaluation results of various $LLM_{eval}$ over $LLM_{gen}=$ LLaMA-3.1-8B.}
        \label{tab:bioasq-LLaMA-3.1-8B-detailed-results}    
    \end{table*}

\begin{table*}[t]
        \centering
        \small
        \begin{tabular}{|p{2cm}|l|l|l|l|l|l|l|l|}
            \hline
             \textbf{Set}&\textbf{Rubrics} &  \textbf{M1} &  \textbf{M2} & \textbf{M3}& \textbf{M4}& \textbf{M5}  & \textbf{M6} & \textbf{M7}\\
             \hline 
             \hline 
                \multirow{9}{*}{\textbf{benign}}& 1. Coherence & 2.91 & 4.90 & 2.91 & 4.91 & 3.64 & 2.55 & 3.09\\ 
        & 2. Cohesion  & 2.91 & 4.90 & 2.86 & 4.86 & 4.05 & 2.32 & 2.86\\ 
        & 3. Completeness  & 2.50 & 3.29 & 2.73 & 4.64 & 3.86 & 2.45 & 2.64\\ 
        & 4. Conciseness  & 2.91 & 4.80 & 3.36 & 4.23 & 3.73 & 2.00 & 2.50\\ 
        & 5. Correctness  & 3.00 & 3.73 & 3.18 & 4.86 & 3.91 & 2.91 & 3.09\\ 
        & 6. Informativeness  & 2.77 & 4.36 & 3.00 & 4.95 & 3.27 & 2.50 & 2.95\\ 
        & 7. Integration  & 2.86 & 4.55 & 2.77 & 4.77 & 3.86 & 1.95 & 2.82\\ 
        & 8. Readability  & 2.91 & 4.90 & 3.05 & 4.41 & 3.23 & 2.05 & 2.86\\ 
        & 9. Relevancy  & 2.95 & 3.80 & 3.23 & 4.95 & 3.73 & 2.50 & 3.09\\ 

        \hline
        \multirow{9}{*}{\textbf{extreme}}& 1. Coherence & 1.50 & 2.09 & 2.14 & 3.59 & 2.73 & 1.00 & 1.00\\ 
        & 2. Cohesion  & 2.50 & 3.92 & 2.64 & 4.50 & 3.45 & 2.27 & 2.68\\ 
        & 3. Completeness  & 1.73 & 2.36 & 2.32 & 4.18 & 2.59 & 1.09 & 1.00\\ 
        & 4. Conciseness  & 1.50 & 1.17 & 1.55 & 2.41 & 3.91 & 1.18 & 1.00\\ 
        & 5. Correctness  & 2.32 & 3.27 & 2.50 & 3.91 & 2.95 & 1.00 & 1.00\\ 
        & 6. Informativeness  & 2.09 & 3.18 & 2.41 & 3.95 & 2.05 & 1.00 & 1.00\\ 
        & 7. Integration  & 2.23 & 3.38 & 2.32 & 4.45 & 3.14 & 1.91 & 2.14\\ 
        & 8. Readability  & 1.50 & 1.75 & 1.50 & 3.68 & 2.05 & 1.00 & 1.00\\ 
        & 9. Relevancy  & 1.45 & 2.73 & 2.41 & 3.95 & 2.00 & 1.00 & 1.00\\ 

        \hline
        \multirow{9}{*}{\textbf{subtle}}& 1. Coherence & 1.95 & 3.40 & 2.32 & 4.41 & 3.09 & 2.45 & 1.91\\ 
        & 2. Cohesion  & 2.50 & 3.33 & 2.45 & 4.64 & 3.55 & 2.32 & 2.95\\ 
        & 3. Completeness  & 2.59 & 3.36 & 2.95 & 4.77 & 3.59 & 2.64 & 2.27\\ 
        & 4. Conciseness  & 1.86 & 1.86 & 1.64 & 3.23 & 3.82 & 1.64 & 1.32\\ 
        & 5. Correctness  & 2.27 & 3.50 & 2.82 & 4.55 & 3.50 & 3.00 & 2.18\\ 
        & 6. Informativeness  & 2.32 & 3.82 & 2.59 & 4.55 & 3.14 & 2.68 & 2.18\\ 
        & 7. Integration  & 2.45 & 3.83 & 2.50 & 4.36 & 3.41 & 1.86 & 2.86\\ 
        & 8. Readability  & 1.91 & 1.92 & 2.41 & 4.05 & 3.27 & 1.68 & 1.50\\ 
        & 9. Relevancy  & 2.36 & 3.67 & 2.68 & 4.82 & 3.14 & 2.41 & 2.09\\ 

        \hline

    \end{tabular}
        \caption{BioASQ dataset detailed evaluation results of various $LLM_{eval}$ over $LLM_{gen}=$ LLaMA-3.1-70B.}
        \label{tab:bioasq-LLaMA-3.1-70B-detailed-results}    
    \end{table*}

\begin{table*}[t]
        \centering
        \small
        \begin{tabular}{|p{2cm}|l|l|l|l|l|l|l|l|}
            \hline
             \textbf{Set}&\textbf{Rubrics} &  \textbf{M1} &  \textbf{M2} & \textbf{M3}& \textbf{M4}& \textbf{M5}  & \textbf{M6} & \textbf{M7}\\
             \hline 
             \hline 
                \multirow{9}{*}{\textbf{benign}}& 1. Coherence & 5.00 & 5.00 & 4.95 & 4.95 & 5.00 & 3.00 & 4.95\\ 
        & 2. Cohesion  & 5.00 & 5.00 & 4.91 & 4.95 & 5.00 & 3.00 & 4.45\\ 
        & 3. Completeness  & 4.27 & 4.00 & 4.36 & 4.73 & 4.27 & 3.00 & 4.05\\ 
        & 4. Conciseness  & 5.00 & 4.91 & 4.91 & 4.55 & 4.86 & 3.00 & 4.45\\ 
        & 5. Correctness  & 4.95 & 5.00 & 4.91 & 4.82 & 4.91 & 3.00 & 4.91\\ 
        & 6. Informativeness  & 4.68 & 4.83 & 4.91 & 4.82 & 4.86 & 3.00 & 4.95\\ 
        & 7. Integration  & 4.91 & 5.00 & 4.86 & 4.77 & 4.91 & 3.00 & 4.86\\ 
        & 8. Readability  & 5.00 & 5.00 & 5.00 & 4.45 & 4.86 & 3.00 & 4.86\\ 
        & 9. Relevancy  & 5.00 & 5.00 & 5.00 & 4.91 & 5.00 & 3.00 & 4.91\\ 

        \hline
        \multirow{9}{*}{\textbf{extreme}}& 1. Coherence & 2.45 & 2.50 & 2.82 & 4.36 & 3.32 & 1.00 & 1.00\\ 
        & 2. Cohesion  & 4.09 & 4.46 & 3.77 & 4.73 & 4.86 & 3.00 & 4.59\\ 
        & 3. Completeness  & 2.32 & 3.10 & 2.68 & 3.95 & 3.91 & 1.00 & 1.00\\ 
        & 4. Conciseness  & 2.27 & 1.60 & 2.09 & 3.27 & 4.77 & 1.09 & 1.09\\ 
        & 5. Correctness  & 3.68 & 3.70 & 3.77 & 4.36 & 3.36 & 1.00 & 1.00\\ 
        & 6. Informativeness  & 3.41 & 3.60 & 3.50 & 4.41 & 2.32 & 1.00 & 1.00\\ 
        & 7. Integration  & 3.59 & 4.00 & 3.23 & 4.68 & 4.95 & 3.00 & 3.14\\ 
        & 8. Readability  & 1.91 & 1.75 & 2.23 & 4.23 & 2.05 & 1.00 & 1.00\\ 
        & 9. Relevancy  & 2.18 & 2.70 & 3.59 & 4.68 & 2.55 & 1.00 & 1.00\\ 

        \hline
        \multirow{9}{*}{\textbf{subtle}}& 1. Coherence & 3.64 & 4.10 & 4.09 & 4.77 & 4.91 & 3.00 & 3.00\\ 
        & 2. Cohesion  & 4.18 & 4.45 & 4.09 & 4.68 & 4.86 & 3.00 & 4.59\\ 
        & 3. Completeness  & 3.64 & 4.10 & 3.86 & 4.95 & 4.27 & 3.00 & 3.59\\ 
        & 4. Conciseness  & 2.82 & 2.70 & 2.36 & 3.86 & 4.59 & 2.59 & 2.14\\ 
        & 5. Correctness  & 4.23 & 4.30 & 4.00 & 4.59 & 4.73 & 3.00 & 3.14\\ 
        & 6. Informativeness  & 4.14 & 4.50 & 4.18 & 4.82 & 4.64 & 3.00 & 3.18\\ 
        & 7. Integration  & 4.00 & 4.55 & 4.00 & 4.86 & 4.95 & 3.00 & 4.45\\ 
        & 8. Readability  & 2.55 & 2.45 & 3.36 & 4.09 & 4.23 & 2.05 & 1.91\\ 
        & 9. Relevancy  & 4.23 & 4.50 & 4.27 & 4.86 & 4.95 & 2.95 & 3.00\\ 

        \hline

    \end{tabular}
        \caption{BioASQ dataset detailed evaluation results of various $LLM_{eval}$ over $LLM_{gen}=$ Qwen2.5-72B.}
        \label{tab:bioasq-Qwen2.5-72B-detailed-results}    
    \end{table*}

\begin{table*}[t]
        \centering
        \small
        \begin{tabular}{|p{2cm}|l|l|l|l|l|l|l|l|}
            \hline
             \textbf{Set}&\textbf{Rubrics} &  \textbf{M1} &  \textbf{M2} & \textbf{M3}& \textbf{M4}& \textbf{M5}  & \textbf{M6} & \textbf{M7}\\
             \hline 
             \hline 
                \multirow{9}{*}{\textbf{benign}}& 1. Coherence & 4.91 & 5.00 & 4.95 & 5.00 & 5.00 & 3.00 & 4.91\\ 
        & 2. Cohesion  & 4.91 & 4.92 & 4.77 & 5.00 & 4.95 & 3.00 & 4.73\\ 
        & 3. Completeness  & 4.27 & 4.00 & 4.41 & 4.57 & 4.32 & 3.00 & 4.36\\ 
        & 4. Conciseness  & 4.95 & 4.67 & 4.95 & 4.59 & 4.73 & 2.86 & 3.77\\ 
        & 5. Correctness  & 4.95 & 5.00 & 4.82 & 4.86 & 4.95 & 3.00 & 4.95\\ 
        & 6. Informativeness  & 4.64 & 4.75 & 4.91 & 4.91 & 4.73 & 3.00 & 4.91\\ 
        & 7. Integration  & 4.86 & 4.83 & 4.64 & 5.00 & 4.95 & 3.00 & 4.68\\ 
        & 8. Readability  & 4.95 & 4.83 & 5.00 & 4.50 & 4.75 & 2.91 & 4.86\\ 
        & 9. Relevancy  & 4.91 & 4.93 & 4.95 & 5.00 & 5.00 & 3.00 & 5.00\\ 

        \hline
        \multirow{9}{*}{\textbf{extreme}}& 1. Coherence & 2.27 & 2.09 & 3.14 & 4.41 & 3.50 & 1.00 & 1.00\\ 
        & 2. Cohesion  & 3.95 & 4.42 & 3.77 & 4.68 & 4.91 & 3.00 & 4.18\\ 
        & 3. Completeness  & 2.32 & 2.45 & 2.91 & 4.18 & 3.68 & 1.00 & 1.00\\ 
        & 4. Conciseness  & 1.95 & 1.38 & 1.82 & 3.09 & 4.45 & 1.09 & 1.09\\ 
        & 5. Correctness  & 3.68 & 3.45 & 3.73 & 4.55 & 3.41 & 1.00 & 1.00\\ 
        & 6. Informativeness  & 3.45 & 3.18 & 3.73 & 4.45 & 3.00 & 1.05 & 1.00\\ 
        & 7. Integration  & 3.59 & 3.77 & 3.09 & 4.68 & 4.82 & 2.91 & 3.00\\ 
        & 8. Readability  & 2.05 & 2.15 & 2.27 & 4.00 & 2.14 & 1.00 & 1.00\\ 
        & 9. Relevancy  & 2.32 & 2.92 & 3.64 & 4.73 & 3.32 & 1.00 & 1.00\\ 

        \hline
        \multirow{9}{*}{\textbf{subtle}}& 1. Coherence & 3.77 & 3.80 & 4.14 & 4.91 & 4.91 & 2.95 & 3.05\\ 
        & 2. Cohesion  & 4.00 & 4.79 & 4.27 & 4.77 & 4.86 & 3.00 & 4.41\\ 
        & 3. Completeness  & 3.86 & 4.20 & 3.50 & 4.91 & 4.50 & 3.00 & 3.68\\ 
        & 4. Conciseness  & 2.91 & 3.15 & 2.82 & 3.91 & 4.91 & 2.59 & 2.27\\ 
        & 5. Correctness  & 4.23 & 4.27 & 4.14 & 4.77 & 4.68 & 3.00 & 3.23\\ 
        & 6. Informativeness  & 4.18 & 4.20 & 4.32 & 4.86 & 4.59 & 2.95 & 3.14\\ 
        & 7. Integration  & 3.91 & 4.09 & 3.77 & 4.59 & 4.95 & 3.00 & 4.45\\ 
        & 8. Readability  & 2.82 & 2.54 & 3.73 & 4.36 & 4.59 & 2.27 & 1.95\\ 
        & 9. Relevancy  & 4.27 & 4.18 & 4.41 & 5.00 & 5.00 & 2.95 & 2.95\\ 

        \hline

    \end{tabular}
        \caption{BioASQ dataset detailed evaluation results of various $LLM_{eval}$ over $LLM_{gen}=$ Mistral-Large.}
        \label{tab:bioasq-Mistral-Large-detailed-results}    
    \end{table*}

\begin{table*}[t]
        \centering
        \small
        \begin{tabular}{|p{2cm}|l|l|l|l|l|l|l|l|}
            \hline
             \textbf{Set}&\textbf{Rubrics} &  \textbf{M1} &  \textbf{M2} & \textbf{M3}& \textbf{M4}& \textbf{M5}  & \textbf{M6} & \textbf{M7}\\
             \hline 
             \hline 
                \multirow{9}{*}{\textbf{benign}}& 1. Coherence & 4.74 & 4.95 & 4.78 & 4.97 & 4.95 & 3.00 & 4.98\\ 
        & 2. Cohesion  & 4.72 & 4.93 & 4.71 & 4.97 & 4.88 & 3.00 & 4.95\\ 
        & 3. Completeness  & 4.28 & 4.40 & 4.52 & 4.50 & 4.59 & 3.00 & 4.76\\ 
        & 4. Conciseness  & 4.74 & 4.71 & 4.70 & 4.18 & 4.77 & 3.00 & 4.03\\ 
        & 5. Correctness  & 4.86 & 4.97 & 4.79 & 4.83 & 4.91 & 3.00 & 4.94\\ 
        & 6. Informativeness  & 4.57 & 4.93 & 4.77 & 4.97 & 4.89 & 3.06 & 4.97\\ 
        & 7. Integration  & 4.70 & 4.90 & 4.68 & 4.89 & 4.93 & 3.00 & 4.91\\ 
        & 8. Readability  & 4.77 & 4.94 & 4.86 & 4.41 & 4.85 & 3.00 & 4.86\\ 
        & 9. Relevancy  & 4.70 & 4.92 & 4.81 & 4.99 & 4.93 & 3.02 & 4.96\\ 

        \hline
        \multirow{9}{*}{\textbf{extreme}}& 1. Coherence & 2.58 & 3.42 & 3.35 & 4.27 & 4.81 & 1.04 & 1.00\\ 
        & 2. Cohesion  & 3.76 & 4.33 & 3.44 & 4.85 & 4.78 & 3.00 & 4.41\\ 
        & 3. Completeness  & 2.63 & 2.94 & 3.56 & 3.69 & 4.59 & 1.00 & 1.00\\ 
        & 4. Conciseness  & 1.11 & 1.05 & 1.57 & 1.70 & 4.64 & 1.00 & 1.00\\ 
        & 5. Correctness  & 3.81 & 4.09 & 3.98 & 4.24 & 4.82 & 1.01 & 1.00\\ 
        & 6. Informativeness  & 3.62 & 4.26 & 3.80 & 4.78 & 4.82 & 1.03 & 1.00\\ 
        & 7. Integration  & 3.37 & 3.90 & 3.16 & 4.72 & 4.86 & 2.73 & 3.10\\ 
        & 8. Readability  & 2.20 & 2.13 & 2.93 & 3.77 & 4.76 & 1.01 & 1.00\\ 
        & 9. Relevancy  & 2.78 & 3.51 & 3.58 & 4.99 & 4.81 & 1.02 & 1.00\\ 

        \hline
        \multirow{9}{*}{\textbf{subtle}}& 1. Coherence & 4.21 & 4.70 & 4.51 & 4.81 & 4.95 & 3.00 & 3.44\\ 
        & 2. Cohesion  & 3.98 & 4.49 & 3.96 & 4.90 & 4.84 & 3.00 & 4.52\\ 
        & 3. Completeness  & 4.10 & 4.23 & 3.92 & 4.99 & 4.67 & 3.00 & 4.20\\ 
        & 4. Conciseness  & 2.71 & 2.36 & 2.50 & 3.62 & 4.70 & 2.47 & 2.38\\ 
        & 5. Correctness  & 4.48 & 4.78 & 4.44 & 4.57 & 4.93 & 3.00 & 3.37\\ 
        & 6. Informativeness  & 4.35 & 4.79 & 4.50 & 4.90 & 4.87 & 3.03 & 3.38\\ 
        & 7. Integration  & 4.01 & 4.52 & 4.02 & 4.92 & 4.91 & 2.99 & 4.72\\ 
        & 8. Readability  & 3.22 & 2.88 & 3.79 & 4.49 & 4.82 & 2.85 & 2.98\\ 
        & 9. Relevancy  & 4.35 & 4.70 & 4.56 & 4.99 & 4.96 & 2.99 & 3.54\\ 

        \hline

    \end{tabular}
        \caption{ORKGSynthesis dataset detailed evaluation results of various $LLM_{eval}$ over $LLM_{gen}=$ LLaMA-3.1-8B.}
        \label{tab:orkgsynthesis-LLaMA-3.1-8B-detailed-results}    
    \end{table*}

\begin{table*}[t]
        \centering
        \small
        \begin{tabular}{|p{2cm}|l|l|l|l|l|l|l|l|}
            \hline
             \textbf{Set}&\textbf{Rubrics} &  \textbf{M1} &  \textbf{M2} & \textbf{M3}& \textbf{M4}& \textbf{M5}  & \textbf{M6} & \textbf{M7}\\
             \hline 
             \hline 
                \multirow{9}{*}{\textbf{benign}}& 1. Coherence & 4.77 & 4.97 & 4.78 & 5.00 & 4.88 & 3.00 & 4.98\\ 
        & 2. Cohesion  & 4.76 & 4.95 & 4.65 & 4.99 & 4.87 & 3.00 & 4.98\\ 
        & 3. Completeness  & 4.32 & 4.46 & 4.54 & 4.61 & 4.69 & 3.00 & 4.75\\ 
        & 4. Conciseness  & 4.77 & 4.75 & 4.65 & 4.15 & 4.73 & 2.99 & 4.05\\ 
        & 5. Correctness  & 4.89 & 4.96 & 4.83 & 4.80 & 4.92 & 3.00 & 4.92\\ 
        & 6. Informativeness  & 4.65 & 4.92 & 4.77 & 4.98 & 4.87 & 3.01 & 4.95\\ 
        & 7. Integration  & 4.76 & 4.90 & 4.56 & 4.92 & 4.92 & 3.00 & 4.96\\ 
        & 8. Readability  & 4.80 & 4.98 & 4.90 & 4.41 & 4.89 & 3.00 & 4.94\\ 
        & 9. Relevancy  & 4.73 & 4.90 & 4.75 & 4.99 & 4.93 & 3.03 & 4.94\\ 

        \hline
        \multirow{9}{*}{\textbf{extreme}}& 1. Coherence & 2.66 & 3.39 & 3.33 & 4.31 & 4.82 & 1.01 & 1.00\\ 
        & 2. Cohesion  & 3.80 & 4.30 & 3.56 & 4.88 & 4.87 & 3.00 & 4.46\\ 
        & 3. Completeness  & 2.74 & 2.97 & 3.56 & 3.63 & 4.56 & 1.00 & 1.00\\ 
        & 4. Conciseness  & 1.09 & 1.02 & 1.62 & 2.29 & 4.56 & 1.00 & 1.00\\ 
        & 5. Correctness  & 3.90 & 4.07 & 3.85 & 4.25 & 4.89 & 1.02 & 1.00\\ 
        & 6. Informativeness  & 3.76 & 4.30 & 3.87 & 4.69 & 4.81 & 1.02 & 1.00\\ 
        & 7. Integration  & 3.50 & 4.03 & 3.27 & 4.74 & 4.88 & 2.83 & 3.07\\ 
        & 8. Readability  & 2.16 & 2.03 & 2.95 & 3.85 & 4.71 & 1.00 & 1.00\\ 
        & 9. Relevancy  & 2.80 & 3.67 & 3.70 & 4.92 & 4.86 & 1.00 & 1.00\\ 

        \hline
        \multirow{9}{*}{\textbf{subtle}}& 1. Coherence & 4.16 & 4.72 & 4.52 & 4.80 & 4.89 & 3.00 & 3.42\\ 
        & 2. Cohesion  & 4.03 & 4.56 & 4.05 & 4.85 & 4.88 & 3.00 & 4.60\\ 
        & 3. Completeness  & 4.13 & 4.37 & 4.13 & 4.98 & 4.66 & 2.99 & 4.49\\ 
        & 4. Conciseness  & 2.51 & 2.07 & 2.29 & 3.44 & 4.77 & 2.22 & 2.20\\ 
        & 5. Correctness  & 4.35 & 4.76 & 4.31 & 4.50 & 4.92 & 3.00 & 3.38\\ 
        & 6. Informativeness  & 4.32 & 4.85 & 4.41 & 4.87 & 4.91 & 3.00 & 3.46\\ 
        & 7. Integration  & 4.08 & 4.49 & 3.98 & 4.93 & 4.90 & 3.00 & 4.87\\ 
        & 8. Readability  & 3.31 & 3.09 & 3.76 & 4.42 & 4.77 & 2.90 & 2.95\\ 
        & 9. Relevancy  & 4.31 & 4.63 & 4.50 & 5.00 & 4.85 & 2.98 & 3.48\\ 

        \hline

    \end{tabular}
        \caption{ORKGSynthesis dataset detailed evaluation results of various $LLM_{eval}$ over $LLM_{gen}=$ LLaMA-3.1-70B.}
        \label{tab:orkgsynthesis-LLaMA-3.1-70B-detailed-results}    
    \end{table*}

\begin{table*}[t]
        \centering
        \small
        \begin{tabular}{|p{2cm}|l|l|l|l|l|l|l|l|}
            \hline
             \textbf{Set}&\textbf{Rubrics} &  \textbf{M1} &  \textbf{M2} & \textbf{M3}& \textbf{M4}& \textbf{M5}  & \textbf{M6} & \textbf{M7}\\
             \hline 
             \hline 
                \multirow{9}{*}{\textbf{benign}}& 1. Coherence & 4.86 & 4.99 & 4.87 & 5.00 & 4.93 & 3.01 & 5.00\\ 
        & 2. Cohesion  & 4.86 & 4.98 & 4.78 & 4.99 & 4.90 & 3.00 & 5.00\\ 
        & 3. Completeness  & 4.47 & 4.65 & 4.62 & 4.76 & 4.70 & 3.00 & 4.93\\ 
        & 4. Conciseness  & 4.87 & 4.81 & 4.72 & 4.30 & 4.81 & 3.00 & 4.13\\ 
        & 5. Correctness  & 4.90 & 4.98 & 4.86 & 4.89 & 4.94 & 3.00 & 4.98\\ 
        & 6. Informativeness  & 4.78 & 4.98 & 4.91 & 4.97 & 4.93 & 3.06 & 5.00\\ 
        & 7. Integration  & 4.86 & 4.97 & 4.72 & 4.91 & 4.94 & 3.00 & 4.98\\ 
        & 8. Readability  & 4.89 & 4.97 & 4.90 & 4.44 & 4.87 & 3.00 & 4.98\\ 
        & 9. Relevancy  & 4.83 & 4.96 & 4.91 & 5.00 & 4.94 & 3.05 & 4.98\\ 

        \hline
        \multirow{9}{*}{\textbf{extreme}}& 1. Coherence & 2.74 & 3.75 & 3.54 & 4.47 & 4.91 & 1.04 & 1.00\\ 
        & 2. Cohesion  & 3.94 & 4.44 & 3.70 & 4.91 & 4.88 & 3.00 & 4.51\\ 
        & 3. Completeness  & 2.90 & 2.91 & 3.78 & 3.81 & 4.65 & 1.01 & 1.00\\ 
        & 4. Conciseness  & 1.81 & 1.31 & 1.90 & 2.87 & 4.69 & 1.01 & 1.00\\ 
        & 5. Correctness  & 3.84 & 4.15 & 4.01 & 4.31 & 4.92 & 1.01 & 1.00\\ 
        & 6. Informativeness  & 3.84 & 4.51 & 3.98 & 4.78 & 4.81 & 1.00 & 1.00\\ 
        & 7. Integration  & 3.82 & 4.34 & 3.60 & 4.86 & 4.89 & 2.89 & 3.52\\ 
        & 8. Readability  & 2.38 & 2.16 & 3.07 & 4.18 & 4.76 & 1.00 & 1.00\\ 
        & 9. Relevancy  & 3.06 & 3.70 & 3.83 & 4.98 & 4.88 & 1.00 & 1.00\\ 

        \hline
        \multirow{9}{*}{\textbf{subtle}}& 1. Coherence & 4.23 & 4.85 & 4.63 & 4.84 & 4.93 & 3.00 & 3.41\\ 
        & 2. Cohesion  & 4.16 & 4.61 & 4.11 & 4.96 & 4.92 & 3.00 & 4.83\\ 
        & 3. Completeness  & 4.40 & 4.50 & 4.30 & 4.99 & 4.73 & 3.00 & 4.79\\ 
        & 4. Conciseness  & 3.13 & 2.82 & 2.68 & 4.50 & 4.83 & 2.89 & 2.97\\ 
        & 5. Correctness  & 4.48 & 4.83 & 4.51 & 4.60 & 4.93 & 3.00 & 3.39\\ 
        & 6. Informativeness  & 4.46 & 4.90 & 4.70 & 4.91 & 4.87 & 3.04 & 3.49\\ 
        & 7. Integration  & 4.21 & 4.68 & 4.06 & 4.96 & 4.92 & 3.00 & 4.90\\ 
        & 8. Readability  & 3.71 & 3.21 & 3.86 & 4.53 & 4.82 & 2.91 & 2.98\\ 
        & 9. Relevancy  & 4.42 & 4.73 & 4.66 & 4.97 & 4.92 & 3.00 & 3.49\\ 

        \hline

    \end{tabular}
        \caption{ORKGSynthesis dataset detailed evaluation results of various $LLM_{eval}$ over $LLM_{gen}=$ Qwen2.5-72B.}
        \label{tab:orkgsynthesis-Qwen2.5-72B-detailed-results}    
    \end{table*}

\begin{table*}[t]
        \centering
        \small
        \begin{tabular}{|p{2cm}|l|l|l|l|l|l|l|l|}
            \hline
             \textbf{Set}&\textbf{Rubrics} &  \textbf{M1} &  \textbf{M2} & \textbf{M3}& \textbf{M4}& \textbf{M5}  & \textbf{M6} & \textbf{M7}\\
             \hline 
             \hline 
                \multirow{9}{*}{\textbf{benign}}& 1. Coherence & 4.79 & 4.95 & 4.83 & 5.00 & 4.92 & 3.00 & 4.98\\ 
        & 2. Cohesion  & 4.78 & 4.93 & 4.76 & 4.99 & 4.91 & 2.99 & 4.98\\ 
        & 3. Completeness  & 4.39 & 4.50 & 4.61 & 4.62 & 4.81 & 3.00 & 4.86\\ 
        & 4. Conciseness  & 4.82 & 4.80 & 4.71 & 4.26 & 4.75 & 2.98 & 4.07\\ 
        & 5. Correctness  & 4.87 & 4.97 & 4.87 & 4.89 & 4.95 & 3.00 & 4.93\\ 
        & 6. Informativeness  & 4.70 & 4.94 & 4.85 & 4.97 & 4.92 & 3.03 & 4.95\\ 
        & 7. Integration  & 4.77 & 4.92 & 4.75 & 4.90 & 4.95 & 3.00 & 4.97\\ 
        & 8. Readability  & 4.84 & 4.91 & 4.90 & 4.34 & 4.83 & 2.99 & 4.93\\ 
        & 9. Relevancy  & 4.70 & 4.92 & 4.83 & 5.00 & 4.94 & 3.01 & 4.90\\ 

        \hline
        \multirow{9}{*}{\textbf{extreme}}& 1. Coherence & 2.58 & 3.50 & 3.47 & 4.34 & 4.88 & 1.00 & 1.00\\ 
        & 2. Cohesion  & 3.72 & 4.29 & 3.62 & 4.87 & 4.90 & 2.99 & 4.60\\ 
        & 3. Completeness  & 2.76 & 2.76 & 3.61 & 3.67 & 4.58 & 1.00 & 1.00\\ 
        & 4. Conciseness  & 1.77 & 1.27 & 1.71 & 2.88 & 4.73 & 1.02 & 1.00\\ 
        & 5. Correctness  & 3.82 & 4.06 & 3.92 & 4.21 & 4.91 & 1.00 & 1.00\\ 
        & 6. Informativeness  & 3.69 & 4.43 & 3.90 & 4.83 & 4.87 & 1.00 & 1.00\\ 
        & 7. Integration  & 3.52 & 4.09 & 3.37 & 4.79 & 4.85 & 2.94 & 3.34\\ 
        & 8. Readability  & 2.36 & 2.10 & 2.97 & 3.84 & 4.62 & 1.01 & 1.00\\ 
        & 9. Relevancy  & 2.89 & 3.50 & 3.76 & 4.95 & 4.83 & 1.00 & 1.00\\ 

        \hline
        \multirow{9}{*}{\textbf{subtle}}& 1. Coherence & 4.17 & 4.76 & 4.45 & 4.78 & 4.97 & 3.00 & 3.33\\ 
        & 2. Cohesion  & 3.99 & 4.53 & 3.98 & 4.92 & 4.92 & 3.00 & 4.49\\ 
        & 3. Completeness  & 4.05 & 4.38 & 4.11 & 5.00 & 4.66 & 3.00 & 4.64\\ 
        & 4. Conciseness  & 2.98 & 2.66 & 2.60 & 4.22 & 4.85 & 2.90 & 2.89\\ 
        & 5. Correctness  & 4.36 & 4.78 & 4.45 & 4.50 & 4.92 & 3.00 & 3.35\\ 
        & 6. Informativeness  & 4.32 & 4.85 & 4.50 & 4.91 & 4.95 & 3.02 & 3.39\\ 
        & 7. Integration  & 4.05 & 4.50 & 4.03 & 4.90 & 4.89 & 3.00 & 4.91\\ 
        & 8. Readability  & 3.43 & 3.02 & 3.82 & 4.61 & 4.85 & 2.86 & 2.97\\ 
        & 9. Relevancy  & 4.30 & 4.68 & 4.52 & 4.99 & 4.96 & 3.00 & 3.42\\ 

        \hline

    \end{tabular}
        \caption{ORKGSynthesis dataset detailed evaluation results of various $LLM_{eval}$ over $LLM_{gen}=$ Mistral-Large.}
        \label{tab:orkgsynthesis-Mistral-Large-detailed-results}    
    \end{table*}

\section{Example Evaluations}
\label{sec:example-outputs}
This section presents example evaluations of LLM-generated responses based on different rubrics. The examples were selected randomly from experimental models $LLM_{eval}$ outputs in test sets. The examples showcase both well-performing and problematic cases, emphasizing the strengths and weaknesses of different models and fine-tuning strategies. The examples are shown in \autoref{tab:bioasq-example-outputs} and \autoref{tab:orkgsynthesis-example-outputs}, which illustrate the challenges in vanilla and SFT models in distinguishing the desired behaviors. However, the \textit{SFT (benign) + RL (benign+adversarial)} model demonstrates superior performance, even in terms of its rationales. 

Examples from BioASQ in \autoref{tab:bioasq-example-outputs} show that \textit{SFT (benign) + RL (adversarial)} often struggle with coherence and factual accuracy, as evidenced in examples 3 and 5. In contrast, models combining \textit{SFT (benign) + RL (benign + adversarial)} tend to deliver more relevant and informative outputs, as seen in example 4. In \autoref{tab:orkgsynthesis-example-outputs} for the ORKGSynthesis dataset, examples in extreme settings, such as the \textit{SFT (benign) + RL (adversarial)} rationales in example 8, demonstrate that $LLM_{eval}$ specifically identifies the inclusion of off-topic sentences, resulting in lower relevance and completeness ratings, as anticipated.


\begin{table*}[h]
    \centering
        \renewcommand{\arraystretch}{1} 
        \setlength{\tabcolsep}{4pt} 
        \small 
        \resizebox{\textwidth}{!}{
        \begin{tabular}{|l|c|c|p{3cm}|c|c|p{8cm}|}
            \hline
             & \textbf{Set} & \textbf{LLM\textsubscript{gen}} & \textbf{LLM\textsubscript{eval}} & \textbf{Rubric} & \textbf{Rating} & \textbf{Rationale} \\ 
            \hline
            \hline
        1 & subtle & LLaMA-3.1-8B & LLaMA-3.1-8B & Readability & 4 & The synthesis follows academic writing conventions almost perfectly and displays appropriate style. However, the last sentence is a bit unrelated to the research question and seems to be a non-sequitur, which slightly affects the overall rating. \\ 
        \hline
        2 &benign & LLaMA-3.1-70B & SFT (benign) & Coherence & 1 & The synthesis provided does not form a coherent narrative, making it difficult to follow. \\ 
        \hline
        3 &benign & Mistral-Large & SFT (benign) + RL (adversarial) & Integration & 3 & The synthesis integrates the sources into a coherent manner within a single paragraph. However, the transitions or connections between ideas could be smoother. For example, the transition from the information about Denmark to the information about England could be more fluid. Additionally, the synthesis could benefit from better markers of provenance/quotation to clearly indicate which study provided each piece of information. \\ 
        \hline
        4 &benign & Qwen2.5-72B & SFT (benign) + RL (benign+adversarial) & Relevancy & 5 & The synthesis directly and consistently addresses the research question of how circRNAs are derived, providing a clear and concise overview of the back-splicing process and its key aspects. \\ 
        \hline
        5 &adversarial & Mistral-Large & SFT (adversarial) + RL (benign+adversarial) & Accuracy & 2 & The synthesis contains factual inaccuracies regarding the regulatory mechanisms of circRNAs. While the overall structure is well-organized, some key claims about exon-skipping and alternative splicing are incorrect. \\ 
        \hline
        6 &subtle & Qwen2.5-72B & LLaMA-3.1-8B & Coherence & 4 & The ideas are logically connected for the most part, but the narrative could be strengthened for better clarity. Some minor issues with sentence structure and wording were noted, but they do not significantly impact the overall coherence of the synthesis. \\ 
        \hline
        7 &benign & LLaMA-3.1-70B & SFT (benign) & Informativeness & 5 & The synthesis is highly informative, providing valuable insights and detailed information that thoroughly addresses the research question. \\ 
        \hline
        8 &subtle & LLaMA-3.1-70B & SFT (benign) + RL (adversarial) & Coherence & 3 & The ideas are logically connected for the most part, but the narrative could be strengthened for better clarity. The synthesis jumps abruptly from discussing the biogenesis of circRNAs to mentioning a mortality rate without adequately transitioning or providing context. \\ 
        \hline
        9 &extreme & LLaMA-3.1-70B & SFT (benign) + RL (benign+adversarial) & Readability & 1 & The synthesis is poorly written, with pervasive issues in style, structure, and language use, making it difficult to understand. The sentence '\#jets fans must be really confused with \#marksanchez performance as a 2nd string QB for phily!! Lmao' is entirely unrelated to the topic of eRNAs and disrupts the flow of the text.\\ 
        \hline
        \end{tabular}}
    \caption{Example evaluation of LLM-generated responses based on different rubrics for the BioASQ dataset.}
    \label{tab:bioasq-example-outputs}
\end{table*}

\begin{table*}[h]
    \centering
    \renewcommand{\arraystretch}{1} 
    \setlength{\tabcolsep}{4pt} 
    \small 
    \resizebox{\textwidth}{!}{
    \begin{tabular}{|l|c|c|p{3cm}|c|c|p{8cm}|}
        \hline
        &\textbf{Set} & \textbf{LLM\textsubscript{gen}} &\textbf{LLM\textsubscript{eval}} & \textbf{Rubric} & \textbf{Rating} & \textbf{Rationale} \\ 
        \hline
        \hline
        1&benign & LLaMA-3.1-70B & LLaMA-3.1-8B & Correctness & 5 & The synthesis accurately represents the content from the provided abstracts, without any factual errors or misinterpretations. The information is a faithful representation of the original sources, demonstrating a high level of correctness. \\ 
        \hline
        2&benign & Mistral-Large & SFT (benign) & Integration & 5 & The synthesis seamlessly integrates information from the various sources, using appropriate markers and connectors to create a smooth and unified narrative. All information is skillfully condensed into a single, well-structured paragraph. \\ 
        \hline
        3&extreme & Qwen2.5-72B & SFT (benign) + RL (adversarial) & Relevancy & 1 & The synthesis includes a relevant discussion about fake news detection methods, but it deviates significantly from the topic of detecting fake accounts. The last sentence about Stephen Curry, LeBron James, and Kevin Durant winning the 2024 Paris Olympics is completely out of scope and irrelevant to the research question. \\ 
        \hline
        4&extreme & Qwen2.5-72B & SFT (benign) + RL (benign+adversarial) & Completeness & 1 & The synthesis omits most of the relevant information, failing to capture the essential points or details from the provided abstracts. The last sentence of the synthesis is unrelated to the research question and appears to be a random statement about a sporting event. \\ 
        \hline
        5&subtle & LLaMA-3.1-70B & LLaMA-3.1-8B & Readability & 5 & The synthesis follows academic writing conventions perfectly and displays appropriate style, structure, and language use, making it exceptionally accessible and easy to read. \\ 
        \hline
        6&extreme & Qwen2.5-72B & SFT (benign) & Readability & 5 & The synthesis is exceptionally well-written, following stylistic and structural conventions with precise language use, making it accessible and easy to read. The text is well-structured and free of any noticeable issues with style or structure. \\ 
        \hline
        7&subtle & LLaMA-3.1-8B & SFT (benign) + RL (adversarial) & Conciseness & 2 & The synthesis includes some redundant or irrelevant statements, detracting from its clarity. For example, the synthesis repeats the phrase 'ultimately, finally, and in the end' and uses phrases like 'undoubtedly revealed, made evident, and clearly demonstrated' and 'highlighting, underscoring, and emphasizing the pressing, urgent, and critical need.' These repetitions make the synthesis less concise and more verbose than necessary. \\ 
        \hline
        8&extreme & Mistral-Large & SFT (benign) + RL (benign+adversarial) & Relevancy & 1 & The synthesis starts off well by addressing the research question, but it includes an unrelated sentence about Stephen Curry, LeBron James, and Kevin Durant, which is completely out of scope and lacks any connection to the topic of optimizing drug delivery methods to effectively cross the blood-brain barrier. \\ 
        \hline
    \end{tabular}}
    \caption{Example evaluation of LLM-generated responses based on different rubrics for the ORKGSynthesis dataset.}
    \label{tab:orkgsynthesis-example-outputs}
\end{table*}

\end{document}